\documentclass{article}

\usepackage[final,nonatbib]{neurips_data_2023}

\usepackage[T1]{fontenc}    
\usepackage{url}            
\usepackage{booktabs}       
\usepackage{amsfonts}       
\usepackage{nicefrac}       
\usepackage{microtype}      
\usepackage{xcolor}         
\usepackage[colorlinks,citecolor=green]{hyperref}

\definecolor{green}{RGB}{47,110,200}
\usepackage{titlesec}
\usepackage{textcomp,booktabs}

\usepackage{colortbl}
\definecolor{pink}{rgb}{.99,.91,.95}
\usepackage{graphics}
\usepackage{multirow}
\usepackage{pdfpages}
\usepackage{amsmath, amssymb, amsthm, xspace, color}

\usepackage{graphicx} 
\usepackage{float} 
\usepackage{bm}
\usepackage{subfigure}
\usepackage{enumitem}
\usepackage{wrapfig}

\usepackage{macros}

\usepackage{xspace}
\newcommand{\method}{AttrPrompt\xspace}
\newcommand{\baseline}{SimPrompt\xspace}

\newcommand{\hl}[1]{\textcolor{magenta}{\small $\{$\texttt{#1}$\}$}}

\newcommand{\red}[1]{\textcolor{red}{#1}}
\newcommand{\green}[1]{\textcolor{teal}{#1}}

\title{Large Language Model as Attributed Training Data Generator: A Tale of Diversity and Bias}

\author{%
Yue Yu$^{1}$\footnotemark[1], Yuchen Zhuang$^{1}$\footnotemark[1], Jieyu Zhang$^{2}$\thanks{These authors contributed equally to this work.},  Yu Meng$^3$, \\ 
\bf Alexander Ratner$^2$, Ranjay Krishna$^2$, Jiaming Shen$^4$, Chao Zhang$^{1}$\\
$^1$ Georgia Institute of Technology \quad $^2$ University of Washington \\ 
$^3$ University of Illinois at Urbana-Champaign \quad 
$^4$ Google Research\\
\texttt{\small \{yueyu, yczhuang, chaozhang\}@gatech.edu}, \texttt{\small yumeng5@illinois.edu} \\
\texttt{\small \{jieyuz2, ajratner, ranjay\}@cs.washington.edu}, 
\texttt{\small jmshen@google.com}
}

\begin{document}

\maketitle

\begin{abstract}

Large language models (LLMs) have been recently leveraged as training data generators for various natural language processing (NLP) tasks. While previous research has explored different approaches to training models using generated data, they generally rely on simple class-conditional prompts, which may limit the diversity of the generated data and inherit systematic biases of LLM. Thus, we investigate training data generation with diversely attributed prompts (e.g., specifying attributes like length and style), which have the potential to yield diverse and attributed generated data. Our investigation focuses on datasets with high cardinality and diverse domains, wherein we demonstrate that attributed prompts outperform simple class-conditional prompts in terms of the resulting model's performance. Additionally, we present a comprehensive empirical study on data generation encompassing vital aspects like bias, diversity, and efficiency, and highlight three key observations: firstly, synthetic datasets generated by simple prompts exhibit significant biases, such as regional bias; secondly, attribute diversity plays a pivotal role in enhancing model performance; lastly, attributed prompts achieve the performance of simple class-conditional prompts while utilizing only 5\% of the querying cost of ChatGPT associated with the latter. We release the generated dataset and used prompts to facilitate future research\footnote {The data and code is available on {\url{https://github.com/yueyu1030/AttrPrompt}}.}.
\end{abstract}

\section{Introduction}
Large language models (LLMs) have demonstrated exceptional performance across a broad range of NLP tasks~\cite{brown2020language,ouyang2022training,liang2022holistic,gpt4,chatgpt,zhang2022opt}. In recent research, LLMs have been proposed as \emph{task-specific} training data generators, particularly for text classification, aiming to alleviate the need for task-specific data and annotations~\cite{ye-etal-2022-zerogen,gao2023selfguided,ye-etal-2022-progen,supergen,yu2023zero,Chen2023MixtureOS}.
While these efforts have showcased the effectiveness of LLMs as data generators, the focus has primarily been on advancing the training stage, where the generated data are utilized to train task-specific models, leaving the upstream data generation process relatively unexplored. Notably, the prevailing approach employs a simple class-conditional prompt for querying LLMs during data generation, potentially limiting the diversity of the generated data~\cite{Chen2023MixtureOS,tevet-berant-2021-evaluating,zhang2023trade} and inheriting systematic biases inherent in LLMs~\cite{Zhuo2023ExploringAE,Kirk2021BiasOA}. We refer to this simple class-conditional prompt as \baseline, providing an example in Table~\ref{tbl:prompt}.

In this work, we ground the LLM to ChatGPT~\cite{chatgpt}\footnote{We use \texttt{gpt-3.5-turbo} in our main experiments.} for its ability to generate high-quality, human-like text~\cite{ling2023beyond}, and consider four challenging topic classification tasks with high cardinality from various domains. 
Our investigation primarily revolves around assessing the bias and diversity present within the generated training set through the lens of \emph{data attributes}.
In particular, data attributes encompass multiple attribute dimensions and their corresponding attribute values, where the latter represent possible instantiations of the former. For example, an attribute value such as \mquote{shorter than 200 words} could serve as an instantiation of the attribute dimension \mquote{length}.

On one hand, we employ a trained attribute classifier to examine the \emph{attribute bias} present in the dataset generated using \baseline. When analyzing the \mquote{location} attribute in the NYT news dataset, we observe a striking bias towards \mquote{North America} in the predicted values of the generated data, accounting for a significant majority (68.01\%). In contrast, instances associated with \mquote{Africa} are remarkably rare, comprising only 0.69\% of the dataset (100 times less prevalent than \mquote{North America}). This regional bias exhibited in the generated dataset can pose substantial challenges when constructing reliable machine learning models~\cite{Li2022HERBMH,cao-etal-2023-assessing}.

On the other hand, we explore the influence of \emph{attribute diversity} on the downstream model performance. Specifically, we leverage ChatGPT to generate attributed data by incorporating desired attributes as constraints in the prompts. By comparing the performance of models trained on datasets generated using prompts with random attributes against those with fixed attributes, we observe a substantial underperformance of the latter, uncovering the importance of attribute diversity of the generated dataset.

\begin{table}[!t]
 \caption{Prompt template for the NYT news dataset.}
 \centering
\scalebox{0.75}{
\begin{tabular}{c|l}
\toprule 
{\textbf{Method}} &
  \multicolumn{1}{c}{\textbf{Prompt}} \\  \midrule
\baseline & {Suppose you are a news writer. Please generate a  \texttt{\{topic-class\}} news in NYT. } \\\midrule
\multirow{5}{*}{\method} & {Suppose you are a news writer. Please generate a  \texttt{\{topic-class\}} news in NYT following the requirements below:} \\
&1. Should focus on \texttt{\{subtopic\}};\\ 
&2. Should be in length between \texttt{\{length:min-words\}} and \texttt{\{length:max-words\}} words;\\
&3. The writing style of the news should be \texttt{\{style\}};\\
&4. The location of the news should be in \texttt{\{location\}}. 
  \\\bottomrule
\end{tabular}
}
\vspace{-2ex}
\label{tbl:prompt}
\end{table}

To alleviate attribute biases and enhance the attribute diversity of the generated data, we propose to generate data with diversely attributed prompts.
For a given classification task, we start by identifying attribute dimensions and their corresponding attribute values in an interactive, semi-automated process facilitated by the LLM. Subsequently, we generate diverse prompts by combining attributes randomly, replacing the simple class-conditional prompt typically used for querying data from the LLM. We refer to these diversely attributed prompts as \method. An example of such prompts can be found in Table~\ref{tbl:prompt}, where the LLM is instructed to generate training data based on attributes such as \texttt{location} and \texttt{style}.

On the four classification tasks, we empirically evaluate the generated datasets by measuring the performance of models trained using two scenarios: 1) solely on the generated dataset, and 2) on a merged dataset comprising the real training set and the generated set. In both scenarios, the dataset generated with \method significantly outperforms its counterpart generated with \baseline. 
Furthermore, we demonstrate the superiority of \method over \baseline in terms of data/budget efficiency and compatibility with different model sizes/various LLM-as-training-data-generator approaches. Notably, \method achieves the performance of \baseline while utilizing only 5\% of the querying cost of ChatGPT associated with \baseline.
Lastly, we extend the LLM-as-training-data-generator paradigm to the more challenging \emph{multi-label classification} tasks for the first time, and \method outperforms \baseline across all evaluation metrics.

\section{Related Work} 
\paragraph{LLMs as Training Data Generators.} With the remarkable success of large language models (LLMs), researchers have recently attempted to leverage them as training data generators. Such applications include generating tabular data~\cite{borisov2023language}, relation triplets~\cite{chia2022relationprompt}, sentence pairs~\cite{schick2021generating}, instruction data~\cite{Peng2023InstructionTW, sun2023principle,wang2022self,shao2023synthetic}, \etc.
Among these applications, we anchor on training data generation for topic classification in a zero-shot setting where no labeled data is available. In this direction, existing approaches typically use simple class-conditional prompts while focusing on mitigating low-quality issues after generation.
Initial explorations in this domain include  SuperGen~\cite{supergen} and ZeroGen~\cite{ye-etal-2022-zerogen}, which use LLMs for text classification and noise robust learning techniques~\cite{muller2019does,wang2019symmetric} to handle data quality issues. 
SunGen~\cite{gao2023selfguided} reweights the generated data during training with learned data quality weight,  and ProGen~\cite{ye-etal-2022-progen} selects highly influential generated data via model feedback.
In this work, we instead explore attributed prompts to reduce the issue of low informativeness and redundancy, which can be readily incorporated into the existing systems mentioned above. 
Notably, Chen et al.~\cite{Chen2023MixtureOS} also explore prompts to advance the data generation process, yet it adopts soft prompts and requires a white-box LLM and seed examples to tune them. In contrast, our method is applicable to black-box LLMs and even LLM APIs (\eg, ChatGPT) and does not rely on any labeled examples. 
A recent work WANLI \cite{liu-etal-2022-wanli} also considers human-AI collaboration for creating more challenging training data, but requires an initial dataset and a strong task model. Instead, we aim to generate training data without any initial dataset or a pre-existing task model, which allows us to effectively handle resource-limited scenarios.

\paragraph{Attribute-aware Text Generation.}  There are also several existing works \cite{logeswaran2018content,russo2020control,yu2021attribute} that incorporate attributes for controlled text generation, but these are concentrated on very different tasks like style transfer. Typically, these methods necessitate explicit provision of attributes. Differently, we introduce a semi-automated strategy that allows LLMs to propose attribute values autonomously.

\paragraph{Discrete Prompt Optimization.} Several works attempt to optimize discrete prompts for querying LLMs with large language models \cite{reynolds2021prompt,zhou2023large,mishra-nouri-2023-help}. 
More related to us, \cite{mishra-etal-2022-reframing,khot2023decomposed} reframe prompts by decomposing a complex task instruction into multiple simple ones. 
However, these approaches mainly focus on the \emph{inference} stage for directly predicting the answer and may rely on additional labeled examples for validation. 
Our focus is on an orthogonal setting, optimizing prompts for LLMs with attributes to diversify the generated training data. This approach improves the model's overall performance without the need for additional labeled examples.

\section{Large Language Model as Attributed Training Data Generator}
\begin{figure}[!t]
  \centering
  \includegraphics[width=\linewidth]{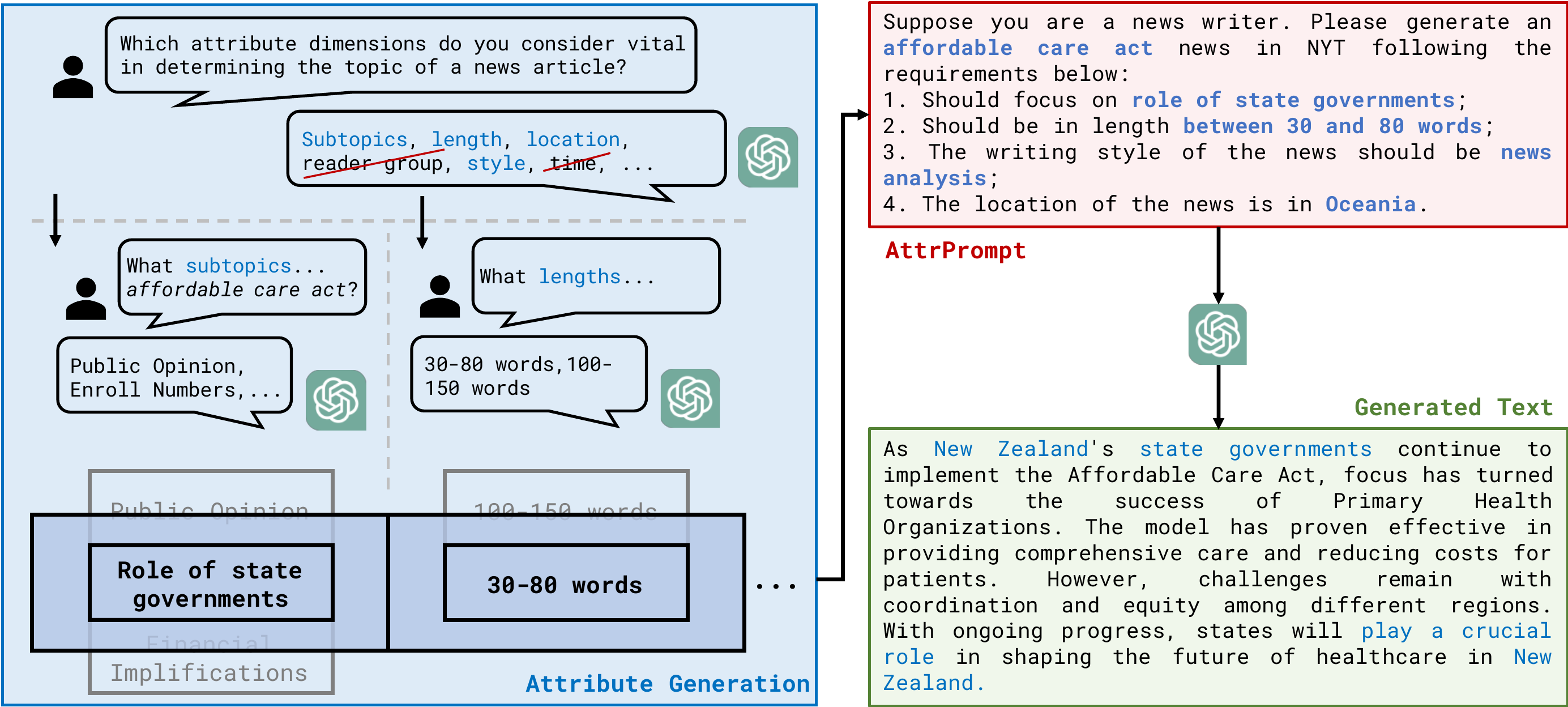}
  \caption{The overall workflow of \method.}
  \vspace{-2ex}
  \label{fig:workflow}
\end{figure}
In this section, we present the design  of our proposed method, \method. This technique employs class-conditional attributes as an enhancement to the query prompts employed in Large Language Models (LLMs). These augmented prompts enable more effective data generation for training purposes. A detailed workflow of the \method can be referenced in Figure \ref{fig:workflow}. 

\subsection{Datasets}
\label{sec:datasets}
While previous research has primarily focused on binary classification datasets~\cite{ye-etal-2022-zerogen,supergen,ye-etal-2022-progen} or datasets containing a maximum of 14 classes~\cite{gao2023selfguided,yu2023zero}, the performance of LLM as a data generator for topic classification with high cardinality (\ie, many topic classes) remains unclear. Thus, we consider the following datasets from various domains with the number of topics ranging from 23 to 50\footnote{We also conduct experiments on those datasets widely used in prior works~\cite{ye-etal-2022-zerogen,ye-etal-2022-progen,gao2023selfguided,yu2023zero}, the details can be found in Appendix \ref{sec:other_datasets}.}:
\vspace{-1ex}
\begin{itemize}[leftmargin=0.5cm]
    \item \textbf{NYT}~\cite{meng2019weakly}: The NYT dataset comprises news articles that were authored and published by \emph{The New York Times}. These articles are categorized into   26 fine-grained categories.
    \item \textbf{Amazon}~\cite{blitzer2007biographies}: The Amazon dataset contains customer reviews on products from Amazon's online store. It covers products from 23 different categories.
    \item \textbf{Reddit}~\cite{geigle2021tweac}: The Reddit dataset consists of a vast collection of user-generated content from the popular social media platform Reddit. It encompasses a wide range of topics, discussions, and interactions among users across numerous communities.
    \item \textbf{StackExchange}~\cite{geigle2021tweac}: The StackExchange dataset is a rich collection of structured data encompassing various online communities and knowledge-sharing platforms. It contains a vast array of questions, answers, comments, tags, and user interactions about specific technical problems.
\end{itemize}
\vspace{-1ex}

We summarize the statistics of used dataset in Table~\ref{tab:dataset}, from which we can see that the involved datasets not only have high cardinality but also come with high imbalance ratio, \ie, the ratio of the sample size of the majority class to that of the minority class, which reflects the long-tail class issue in real applications~\cite{alex2021raft}.

\begin{table}[h]
 \caption{Statistics of datasets.}
 \centering
\resizebox{0.85\linewidth}{!}{
\begin{tabular}{cccccccc}
\toprule
    \bf Dataset & \bf Domain & \bf Task & \bf \# Train & \# \bf Valid & \bf \# Test &\bf  \#  Class & \bf Imbalance Ratio  \\\midrule
    Amazon~\cite{blitzer2007biographies}     & Reviews  &  Multi-class & 15.0K & 0.2K & 1.2K & 23 & 155.6 \\
    NYT~\cite{meng2019weakly}  & News     &  Multi-class  & 9.0K & 0.2K & 1.2K & 26 & 357.0 \\
    Reddit~\cite{geigle2021tweac}  & Web     &  Multi-class & 26.6K & 0.2K & 2.3K &  45 &  447.4 \\
    StackExchange~\cite{geigle2021tweac} & Web     &  Multi-class & 27.0K & 0.3K & 2.5K &  50 &  1283.7 \\ 
    \bottomrule
\end{tabular}
}

\label{tab:dataset}
\end{table}

\subsection{Interactive Attribute Generation}
Different from the existing works \cite{supergen,ye-etal-2022-zerogen,gao2023selfguided} that directly use the simple class-conditional prompts for querying LLMs, our initial step involves identifying various types of data attributes (or metadata) that can be manipulated to generate attributed data samples. 
To facilitate this process, we employ ChatGPT to help establish both attribute dimensions and attribute values.
Specifically, we begin by engaging ChatGPT in generating essential attribute dimensions. This is achieved by posing questions such as \mquote{Which attribute dimensions do you consider vital in determining the topic of a news article?} for the NYT dataset, resulting in responses like \mquote{subtopics, length, location, reader group, style, time}.
Then, we adopt the \emph{human-ai collaboration} scheme~\cite{liu-etal-2022-wanli,wiegreffe2022reframing,zhang2022prompt} to  interactively select the attribute dimensions of the highest quality that best suit the dataset.
Similarly, we prompt ChatGPT (the prompt format is listed in Appendix~\ref{sec:prompt_format}) to suggest potential attribute values within each attribute dimension and choose high-quality candidates.

\paragraph{Attribute dimensions and values.} 

There are two types of attribute dimensions: \emph{class-independent} attributes and  \emph{class-dependent} attributes. Class-independent attributes, such as \mquote{length}, remain unchanged across different classes, while class-dependent attributes, like \mquote{subtopic}, have varying attribute values for each class.
 We list attribute dimensions and values for all datasets in Table~\ref{tbl:attribute}. 
These data attributes provide a human-manipulatable interface for generating attributed data. In this study, we explore the potential of leveraging attributes to enhance the data generation process, while leaving the search for the optimal data attributes for a specific task to future work.

\begin{table}[t]
 \caption{Attribute dimensions and values. Attributes with an asterisk* are class-dependent attributes.}
 \centering
\scalebox{0.6}{
\begin{tabular}{c|c|c|c}
\toprule 
{\textbf{Dataset}} &
  {\textbf{\# configurations / class}}  &
 {{\textbf{Attribute dimension}}} &
  {\textbf{Attribute value}} \\  \midrule

\multirow{5}{*}{NYT} & \multirow{5}{*}{600} & Subtopic*
   & Appendix~\ref{subsubsec:nyt-subtopics}
  \\ \cmidrule(lr){3-4} 
 &  & Location & Asia, North America, South America, Africa, Oceania, Europe \\ \cmidrule(lr){3-4} 
 &  & Writing Style & Investigative journalism, Op-Eds, Feature writing, News analysis, Profiles and interviews \\ \cmidrule(lr){3-4} 
 &  & Length & short (30-80 words); long (100-150 words) \\ \midrule
 \multirow{7}{*}{Amazon} & \multirow{7}{*}{1000} & Product Brands*  & Appendix~\ref{subsubsec:amazon-brand} \\ \cmidrule(lr){3-4} 
 & &   Product Names*  & Appendix~\ref{subsubsec:amazon-name} \\
\cmidrule(lr){3-4} 
&  & Usage Experience & Worst, Bad, Average, Good, Excellent \\ \cmidrule(lr){3-4} 
&  & Writing Style & Detailed Review; Comparative Review; Pros and Cons Review; Recommendation Review \\ \cmidrule(lr){3-4} 
 &  & Length & short (30-80 words); long (100-150 words)  
\\ \midrule
 \multirow{5}{*}{Reddit} & \multirow{5}{*}{500} & Resources*   & Appendix~\ref{subsubsec:reddit-resource} \\ \cmidrule(lr){3-4} 
 & &   Experience*  & Appendix~\ref{subsubsec:reddit-experience} \\
\cmidrule(lr){3-4} 
&  & Writing Style & Informative/Educational; Entertaining/Funny; Discussion; Storytelling; Help/Advice \\ \cmidrule(lr){3-4} 
 &  & Length & short (30-80 words); long (100-150 words)  
\\ \midrule
 \multirow{5}{*}{StackExchange} & \multirow{5}{*}{400} & Scenario*   & Appendix~\ref{subsubsec:stackexchange-scenarios} \\ \cmidrule(lr){3-4} 
 & &   Technical Depth  & Beginner; Intermediate; Advanced; Expert \\
\cmidrule(lr){3-4} 
&  & Writing Style & Specific; Comparative; Problem-Solution; Troubleshooting; Tutorial \\ \cmidrule(lr){3-4} 
 &  & Length & short (30-80 words); long (100-150 words)  
   \\ \bottomrule
\end{tabular}
}
\label{tbl:attribute}
\end{table}

\paragraph{Class-dependent attribute value filtering.} 
When dealing with class-dependent attributes, it is crucial to ensure that their attribute values are specifically associated with the corresponding class to avoid ambiguity and potential connections to multiple classes. For example, in the case of the \mquote{economy} class in the NYT dataset, a candidate attribute value generated by ChatGPT for the \mquote{subtopic} could be \mquote{effect of trade tariffs on manufacturing companies}, which is also relevant to the \mquote{international business} class in the NYT. This overlap may introduce ambiguity in the generated data. To address this issue, we employ a filtering process called Class-Dependent Attribute Value Filtering (CAF). First, we query ChatGPT for the top-5 similar classes and then check with ChatGPT whether each class-dependent attribute value is related to these top-5 similar classes. Then, if the answer is positive which indicates a potential ambiguity, we remove that attribute value for the specific class.

\subsection{Data generation and model training}
\vspace{-1ex}

Given the data attributes, one could prompt LLMs to generate data samples with diverse \emph{attribute configurations}. For example, an attribute configuration for the \mquote{federal budget} class of the NYT dataset could be \{\mquote{subtopic}=\mquote{defense spending}, \mquote{length}=\mquote{short:min-words=30, max-words=80}, \mquote{style}=\mquote{investigative journalism}, \mquote{location}=\mquote{North America}\}.
In Table~\ref{tbl:attribute}, we list the number of configurations per class, and one can further expand the number of configurations by adding more attribute dimensions and values.
To generate attributed data samples, we prompt ChatGPT with random configurations. In particular, each time we generate a random configuration, complete a \emph{prompt template} (see Table~\ref{tbl:prompt}) with the generated configuration, and query ChatGPT with the completed prompt to collect generated data samples.

Throughout the experiments, we compare our method (\method) against simple class-conditional prompt (\baseline,~\cite{ye-etal-2022-zerogen}) and the original training set of each dataset (Gold). 
For a fair comparison, we set the number of generated data the same as Gold for both \method and \baseline.
In principle, the generated dataset can be combined with  any classifier (Sec.~\ref{sec:model}) and training techniques (Sec.~\ref{sec:combo}); if not otherwise specified, we choose to fine-tune \texttt{BERT-base-uncased}~\cite{devlin2019bert} as the backbone and use the standard cross-entropy loss by default.
For hyperparameter selection, we adhere to the recommendation in \cite{perez2021true,supergen} for \baseline and \method, and \emph{do not use the validation set} for model selection. Detailed hyperparameter configurations can be found in Appendix~\ref{app:implement}.

\paragraph{A glimpse of the generated data.} 

Here, we present examples of data generated by \method and \baseline, and real data from the Gold set for the \mquote{federal budget} class in the NYT dataset (Table~\ref{tab:text-example}). It is evident that the data generated by ChatGPT exhibit high quality. Particularly, when comparing \method to \baseline, we observe that \method renders more diverse samples. This is because \baseline tends to generate news focused on the U.S., while \method has the capability to generate news from various locations around the world.

\begin{table}[!h]
 \caption{Data examples of different datasets: the \mquote{federal budget} class of the NYT dataset.}
 \centering
\scalebox{0.52}{
\begin{tabular}{l|l}
\toprule 
{\textbf{Method}} &
  \multicolumn{1}{c}{\textbf{Example}} \\  \midrule

\toprule
\multirow{6}{*}{\bf Gold} & 
1. The emergency manager who was sent to reverse the fortunes of this financially troubled city asked some of its creditors on Friday to accept pennies on the dollar as \\
&    he laid out his plan for tackling Detroit's staggering debt, kick starting negotiations that could determine whether the city is headed to bankruptcy court... \\
& 2. Saying that its debt could reach 45 billion by 2017 if Congress does not act, the Postal Service on Wednesday called on lawmakers to give it the flexibility to change \\
&  its business model to keep itself solvent. \\
& 3. Governmental Affairs Committee, Patrick R. Donahoe, the postmaster general, asked Congress to give the Postal Service permission to run its own health plan for \\
&  employees and retirees, modify a Congressional mandate that requires the agency to pay... \\
\midrule
\multirow{7}{*}{\bf \baseline} & 1. Washington D.C. The United States government has recently passed its federal budget for the next fiscal year, setting aside a whopping 4.8 trillion for spending. \\
&  This is the largest budget in the country's history and reflects the government's commitment to improving the country's economy and infrastructural development. \\
& 2. WASHINGTON D.C. The Trump administration released its proposed budget for the federal government on Monday, calling for cuts to programs that provide \\
& assistance to low income Americans while increasing military spending. The proposed budget includes a 10 percent increase in defense spending, bringing the total \\
& military budget to 574 billion... \\
& 3. Washington D.C. The U.S. federal government has presented its budget for the fiscal year 2022, which includes a total of 6 trillion in spending and initiatives aimed
\\
& at making the country more competitive and sustainable. The budget plan includes significant investments in infrastructure, healthcare, education, and renewable energy.\\
\midrule
\multirow{8}{*}{\bf \method}  & 1. In Germany, public opinion on federal budget issues has been influenced by The Federal Agency for Civic Education. Using in depth storytelling, this agency  \\
& communicates budget policies to citizens, emphasizing their impact on society. Its efforts have led to greater engagement and awareness of budget issues among the\\
&   public in recent years. \\ 
& 2. Op Eds The Government Spending Priorities in Mexico As the Mexican federal budget draws near, all eyes are on the government's spending priorities. One area of \\ 
& concern is the role of the Banco de Mexico in managing the country's finances. Many experts argue that the central bank should play a more active role in balancing the  \\
& budget and stimulating economic growth... \\
& 3. TOKYO The recent federal budget proposal has triggered a wave of agency budget cuts and restructuring in Japan, with the Japan Foundation for Regional Development \\
&  being one of the latest casualties. The foundation, which aims to promote regional development and revitalization,  is set to have its budget slashed by 20  next year. \\
  \bottomrule
\end{tabular}
}
\label{tab:text-example}
\end{table}

\section{Diversity Analysis of the Generated Data}

\paragraph{Quantitative study of diversity.} 
To quantify the diversity of the generated training data of \baseline and \method,  we first show the \emph{vocabulary size} of the generated dataset and the Gold dataset, which is a natural way to check the lexical diversity of datasets (Table~\ref{tab:vocabulary}). From the table, we can see that \method has higher lexical diversity than \baseline in terms of both vocabulary size of the whole dataset (All in the table) and the averaged vocabulary size across classes (Class Avg. in the table). Yet, both have much smaller vocabulary sizes than the Gold, indicating there is still room for improvement of the ChatGPT's lexical diversity.





\begin{table}[!h]
\centering
\vspace{-0.9ex}
\caption{Comparison of the vocabulary size of different datasets.
}
\scalebox{0.75}{%
\begin{tabular}{lccc cc cc cc}
\toprule
\multirow{2.5}{*}{\bf Method} & \multicolumn{2}{c}{\bf NYT}    & \multicolumn{2}{c}{\bf Amazon}    & \multicolumn{2}{c}{\bf Reddit} & \multicolumn{2}{c}{\bf StackExchange}     \\ \cmidrule(lr){2-3} \cmidrule(lr){4-5} \cmidrule(lr){6-7}  \cmidrule(lr){8-9} 
& All & Class Avg.       & All & Class Avg.          & All & Class Avg.   & All & Class Avg.      \\ \midrule

Gold & 70.8k & 11.3k &44.7k & 6.64k &50.8k & 4.62k& 52.3k & 3.60k\\

\baseline & 20.6k & 3.13k &  11.6k & 2.50k & 19.9k& 3.06k & 13.3k & 2.20k\\

\rowcolor{green!15}{\method} & 21.4k & 3.50k & 14.0k & 2.76k & 25.4k& 3.64k & 17.8k & 2.93k\\ 
 \bottomrule
\end{tabular}
}
\label{tab:vocabulary}

\end{table} 
\begin{table}[!h]
\centering
\vspace{-0.9ex}
\caption{{Comparison of two quantitative metrics on diversity: the average pairwise sample similarity (APS) and inter-sample N-gram frequency (INGF) of different datasets. For APS, the \emph{lower} stands for better diversity. For INGF, the \emph{higher} stands for better diversity.}
}
\scalebox{0.75}{%
\begin{tabular}{lcccc cccc cccc cccc}
\toprule
\multirow{2.5}{*}{\bf Method} & \multicolumn{4}{c}{\bf NYT}    & \multicolumn{4}{c}{\bf Amazon}        \\ \cmidrule(lr){2-5} \cmidrule(lr){6-9} 
& Inter-Class	APS &   Intra-Class	APS  & APS &	INGF
& Inter-Class	APS &   Intra-Class	APS  & APS &	INGF
     \\ \midrule
{Gold} & 0.098 &	0.358	 &0.122	 &7618.1 &	0.101	 &0.251	 &0.114	 &4992.1\\
{\baseline} & 0.101	& 0.568	 &0.135 &	5277.2	 &0.207	 &0.620	 &0.241 &	2266.5\\
\rowcolor{green!20} {\method} & 0.159 & 	0.474	 &0.182	 &6688.6	 &0.225	 &0.483 &	0.246 &	2605.5\\ \bottomrule
& \multicolumn{4}{c}{\bf Reddit} & \multicolumn{4}{c}{\bf StackExchange} \\
\cmidrule(lr){2-5} \cmidrule(lr){6-9} 
& Inter-Class	APS &   Intra-Class	APS  & APS &	INGF
& Inter-Class	APS &   Intra-Class	APS  & APS &	INGF \\ \midrule
{Gold} &  0.044	& 0.261 &	0.054	 &9079.6 &	0.056	 &0.196	 &0.063 &	5492.4\\
{\baseline} & 0.173 &	0.818 &	0.201 &	2697.8 &	0.282 &	0.804	 &0.302	 &2259.8 \\
\rowcolor{green!20} {\method} & 0.106	& 0.474	 &0.122 &	3994.5 &	0.105	 &0.375	 &0.114	 & 2464.3 \\
 \bottomrule
\end{tabular}
}
\label{tab:diversity}
\end{table}

\begin{figure}[!t]
	\centering
	\vspace{-2ex}
	\subfigure[NYT]{
		\includegraphics[width=0.235\linewidth]{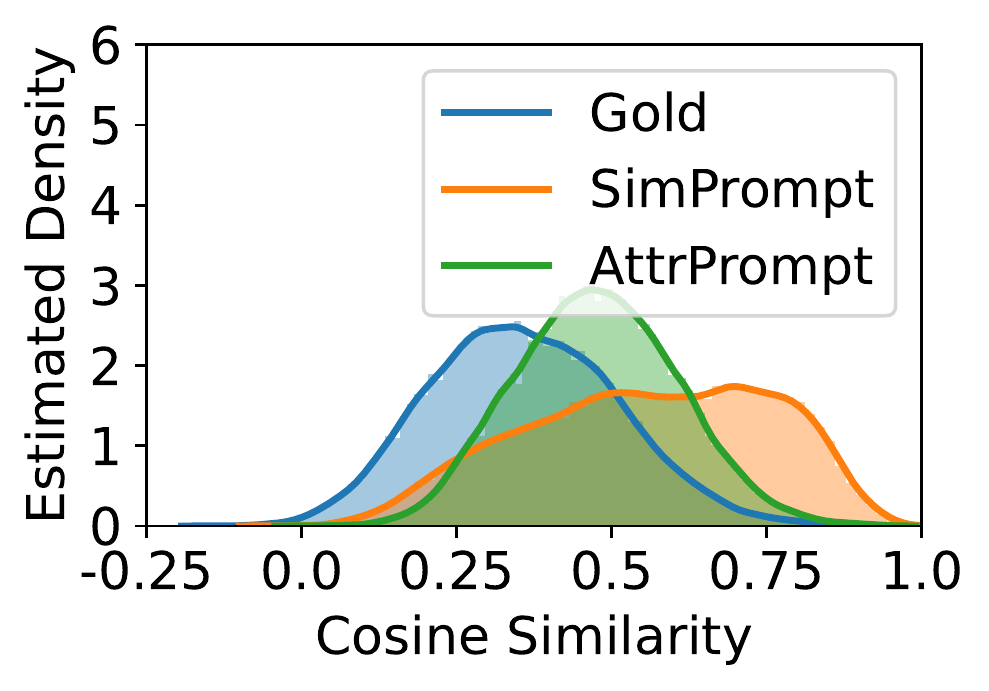}
		\label{fig:nyt-Intra}
	} \hspace{-1ex} 
	\subfigure[Amazon]{
		\includegraphics[width=0.235\linewidth]{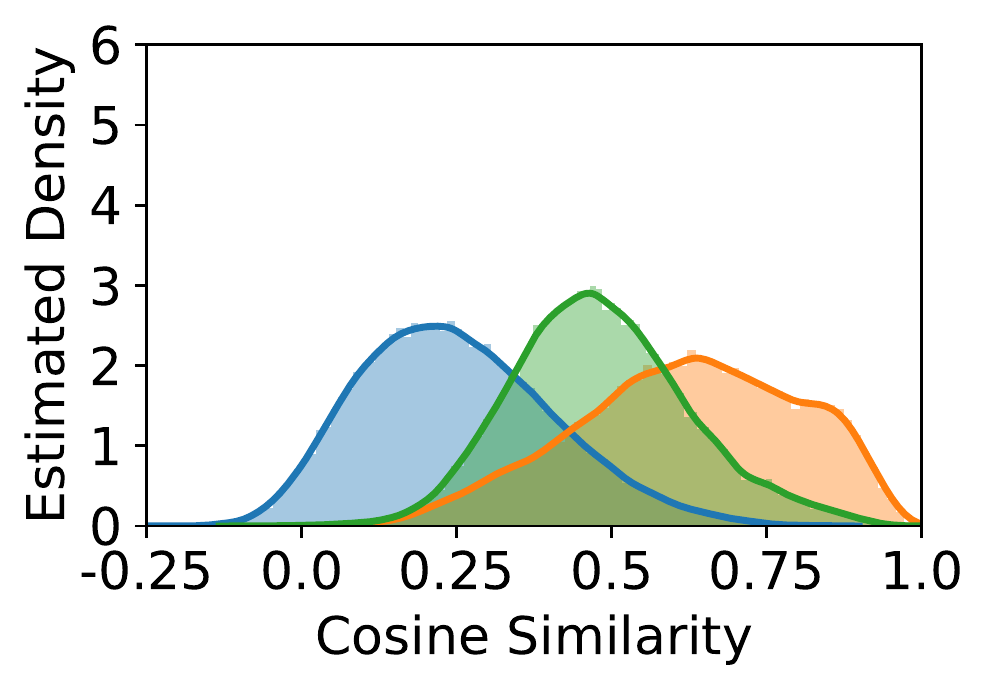}
		\label{fig:amazon-Intra}
	} \hspace{-1ex} 
	\subfigure[Reddit]{
		\includegraphics[width=0.235\linewidth]{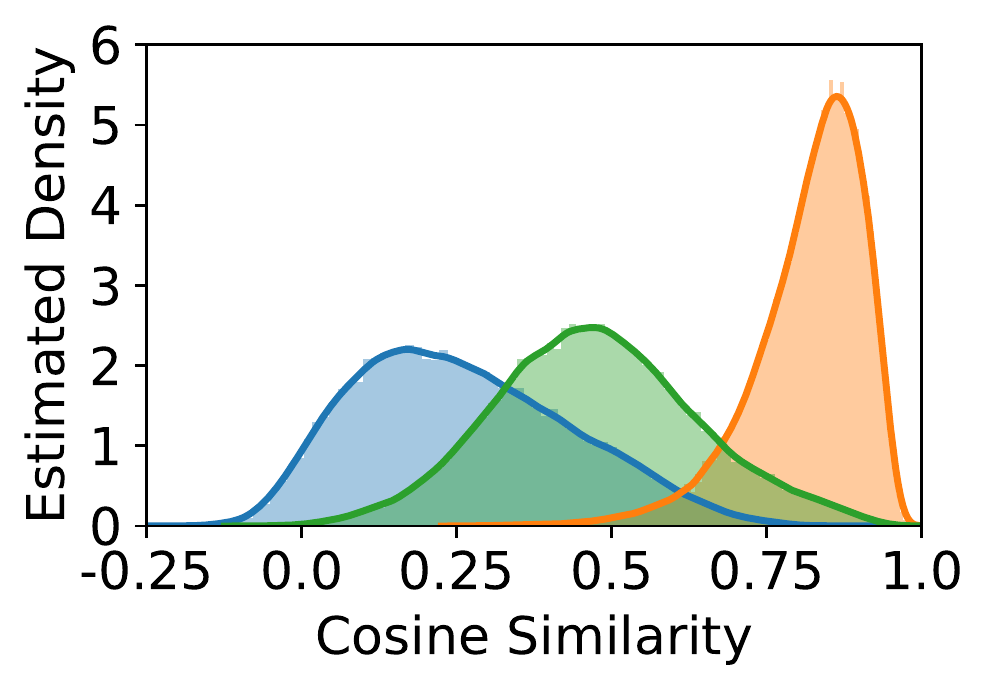}
		\label{fig:reddit-Intra}
	} \hspace{-1ex}  
     \subfigure[StackExchange]{
		\includegraphics[width=0.235\linewidth]{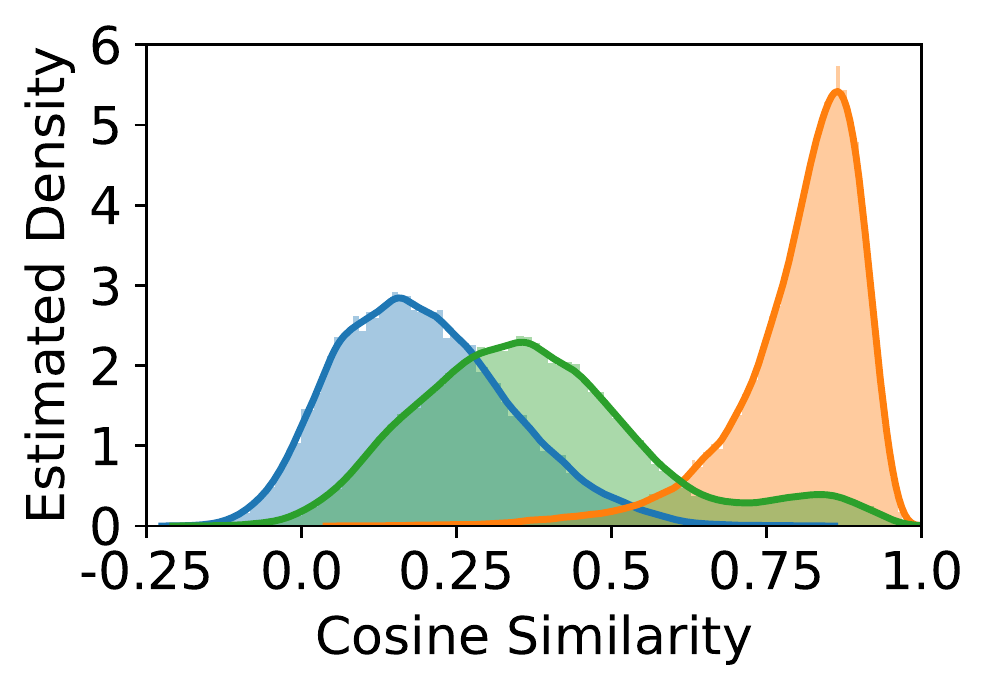}
		\label{fig:stackexchange-Intra}
	} 
    \caption{The distribution of cosine similarity of text pairs sampled from the same class.}
    \label{fig:cosine-sim}
\end{figure}

We then visualize the diversity of datasets via the distribution of cosine similarity of same-class text pairs (Figure~\ref{fig:cosine-sim}), where the cosine similarity is calculated based on the embedding of Sentence-BERT~\cite{reimers2019sentence}, {as well as including two additional metrics, namely \emph{average pairwise sample similarity (APS)} and \emph{inter-sample N-gram Frequency (INGF)}~\cite{mishra2020dqi}, as shown in table \ref{tab:diversity}.}  
We can see that the Gold dataset has the lowest cosine similarity, indicating that real data has the largest diversity. 
In contrast, the similarity between  samples generated by \baseline is  high
Compared to \baseline, dataset generated with \method exhibits lower cosine similarity and the distribution is close to that of the Gold, which shows \method could render more diverse data.   
Apart from the above automatic evaluation processes, we also conduct \emph{human study} in Appendix~\ref{sec:human} to manually evaluate the quality of the generated training data. 


\paragraph{The importance of the attribute diversity.}

\begin{figure}[!h]
	\centering
	\vspace{-2ex}
	\subfigure[Amazon]{
		\includegraphics[width=0.9\linewidth]{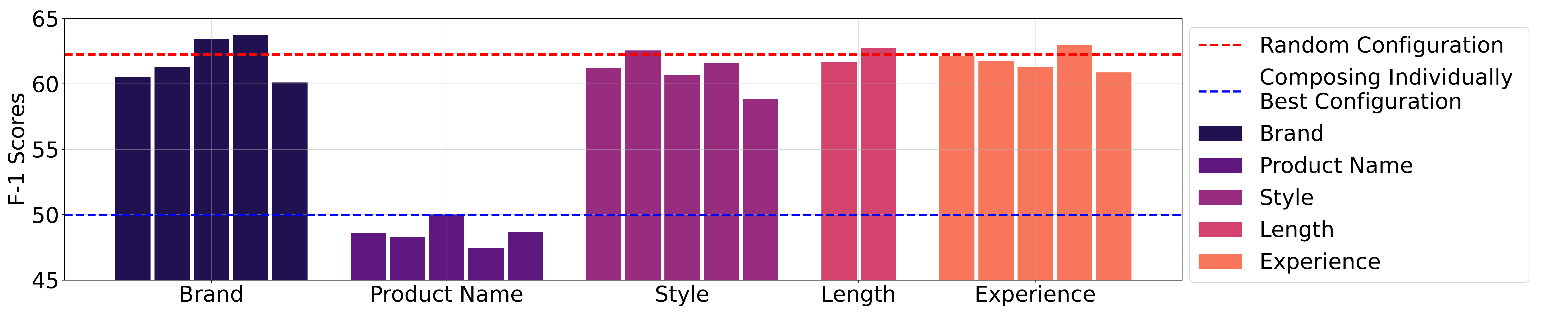}
		\label{fig:config-amazon}
	} 
        \vspace{-1ex}
	\subfigure[NYT]{
		\includegraphics[width=0.9\linewidth]{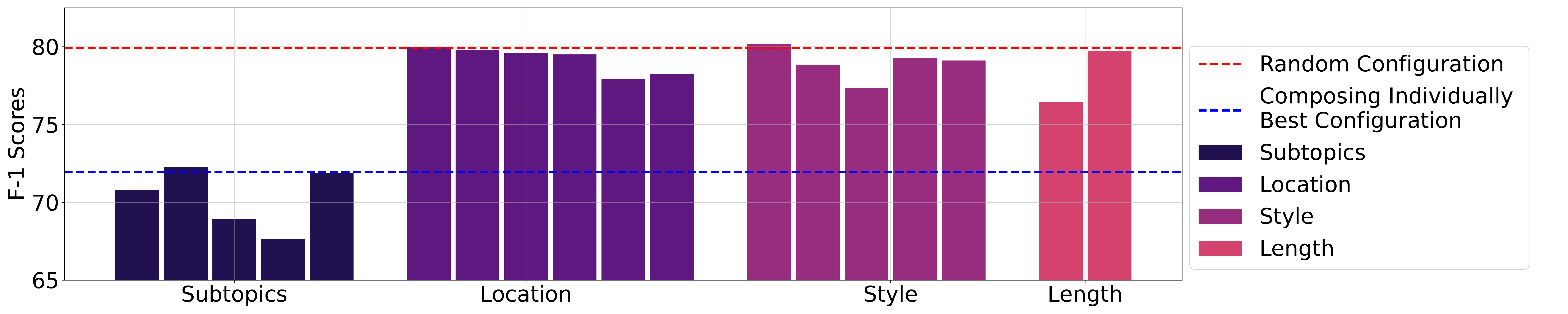}
		\label{fig:config-nyt}
	} 
	\vspace{-1ex}
	\caption{Bar charts of model performance with different attribute configurations of \method.}\label{fig:config}
	\vspace{-1.5ex}
\end{figure}

We investigate the impact of attribute diversity within \method on model performance. Specifically, we conduct experiments by fixing one attribute dimension to a candidate value while keeping other attribute values random. 
Then, we generate 50 data per class using such a one-fixed-others-random configuration to compose a dataset and evaluate the performance of the trained model.
Note that for class-dependent attributes, we sample one value for each class and repeat it 5 times, since it is \emph{computationally prohibitive} to enumerate all combinations of class-dependent attribute values. In Figure~\ref{fig:config}, each bar stands for a specific one-fixed-others-random configuration; compared to random configurations, most of one-fixed-others-random configurations result in a performance drop. 
To further reduce the attribute diversity, we pick the attribute value with the best performance for each attribute dimension (the highest bar within each attribute dimension) and compose them to a single configuration (the dashed blue line). We can see that the dashed blue line is significantly worse than the random configuration, even though it is composed of individually best attribute values. This illustrates the importance and necessity of designing prompts with diverse attributes.

\section{Bias Analysis of the Generated Data}
\label{sec:bias}

In this section, we study the attribute bias in both real and generated datasets. In particular, we pick the \mquote{location} attribute of the NYT data as a case study\footnote{Studies of attribute biases on other datasets can be found in Appendix~\ref{app:bias}.}. While existing works using LLMs as data generators usually overlook the bias embedded in the generated data, we hope that this preliminary analysis could raise the attention of the community to the attribute bias behind the generated data of LLMs such as ChatGPT.

\begin{figure}[h]
  \centering
  \includegraphics[width=0.62\linewidth]{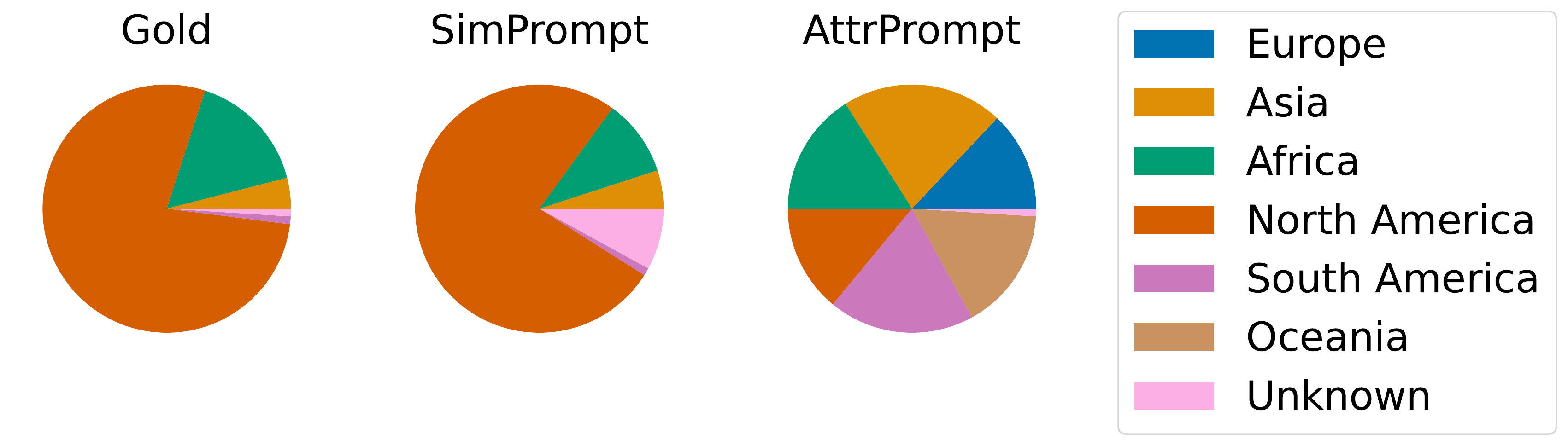}
  \caption{Pie charts of the distributions of the \mquote{location} attribute for the NYT dataset.}
  \label{fig:human-pie}
\end{figure}

We manually annotate the location for 100 samples from each of the Gold, \baseline, and \method dataset. 
Note that we include \mquote{unkown} as an option in manual annotation to absorb text without clear location specifications.
To visualize the distribution of annotated locations in datasets, we plot the pie charts in Figure~\ref{fig:human-pie}.
From the visualizations, one can see that both the Gold and \baseline dataset are largely biased towards \mquote{North America}, while the \method datasets renders a relatively balanced \mquote{location} distribution.

To scale up the study of attribute bias, we leverage the dataset generated by \method as a probe. In particular, we employ the attributes associated with each data of \method to train an \emph{attribute classifier}, which is in turn used to make attribute predictions on Gold and \baseline dataset. Note that the attribute values associated with each data of \method is not necessary the ground truth, yet since ChatGPT has shown remarkable performance in following instructions~\cite{ouyang2022training}, the generated data could decently reflect the desired attributes and therefore the attribute classifier trained with them could partially reveal the underlying attribute distribution of tested dataset, \ie, Gold and \baseline. 
In Appendix~\ref{sec:human}, we justify the use of the attribute classifier by comparing the prediction of the attribute classifier and that of manual annotations.

\begin{figure}[!h]
  \centering
  \includegraphics[width=0.85\linewidth]{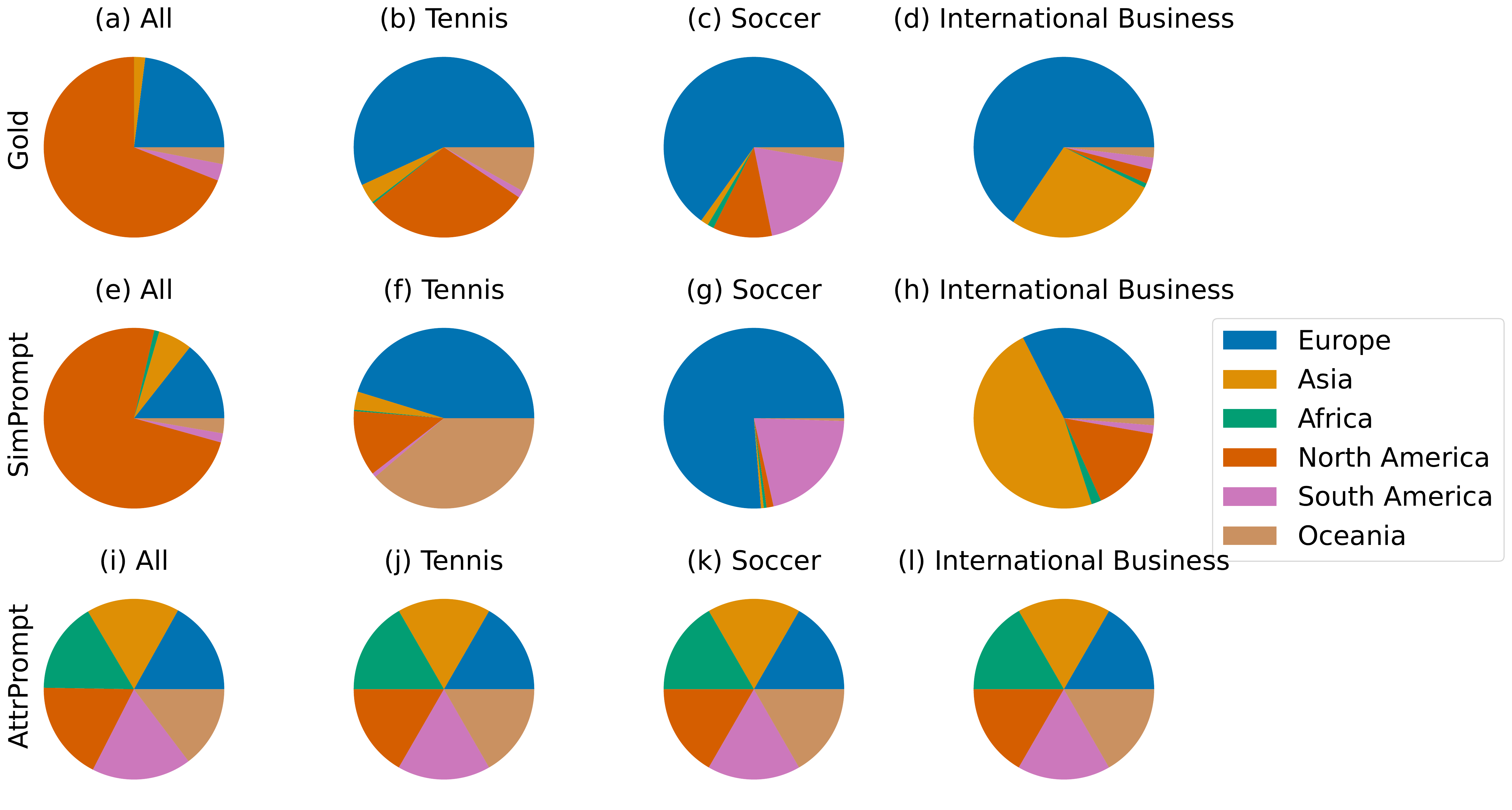}
  \caption{Pie charts of the distributions of \mquote{location} predicted by an attribute classifier for the NYT dataset. (a), (e), and (i) are \mquote{location} distributions over the whole dataset, while others are for specific classes.}
  \label{fig:attr-pie}
\end{figure}

We visualize the distributions of the predicted \mquote{location} in Figure~\ref{fig:attr-pie}.
From the results, we can see that the \mquote{location} distribution of the whole dataset (the first column of Figure~\ref{fig:attr-pie}) is similar to that of manual annotations (Figure~\ref{fig:human-pie}). Regarding the \mquote{location} distribution of specific classes, we can see that while the \method still exhibits balanced distribution, the Gold and \baseline are biased towards continents other than \mquote{North America}. In addition, for the class \mquote{tennis}, the Gold dataset contains much more \mquote{North America} than \mquote{Oceania}, while the \baseline, in contrast, demonstrates an opposite trend with a higher representation of \mquote{Oceania} than \mquote{North America}. Such a noticeable disparity highlights the unpredictable nature of biases, potentially posing risks to models trained on such biased datasets.

\section{Experiments on the Trained Models}
\subsection{Training with generated data}
We quantitatively evaluate the quality of generated datasets via the test performance of models trained with them. 
Apart from the \method and the direct baseline \baseline, we include an additional baseline MetaPrompt \cite{reynolds2021prompt} which leverage LLM to generate additional guidance information for improving upon \baseline. The details for MetaPrompt are shown in Appendix \ref{sec:metaprompt}.
In addition, we use ChatGPT as a zero-shot predictor for comparison.
The results are in Table~\ref{tab:Main}.
Besides the test performance, we include the cost of querying the ChatGPT per 1000 data in the table. 

From the results, we can draw the following conclusions.
First, the \method consistently renders better performance compared to the \baseline with a margin of 6–10 points\footnote{In Appendix~\ref{sec:temp} we show that simply increasing the temperature $t$ for \baseline does not significantly improve its performance.}. Second, the class-dependent attribute value filter (CAF) is beneficial since the \method outperforms its variant without CAF\footnote{Examples of the filtered attributes are exhibited in Appendix~\ref{sec:filter_attr}.}. 
Third, out of the four datasets, the \method outperforms the LLM zero-shot method on three datasets in terms of accuracy, while for the F1 score, the \method surpasses the LLM zero-shot on all the datasets; combined with the observation that the LLM zero-shot inference incurs much higher costs compared to data generation and the fact that the generated data is re-usable for training any model, we argue that for topic text classification generating training data could be a better practice of leveraging LLM than direct zero-shot inference.
Lastly, in most cases, the generated data underperform the original training set, indicating that there is still room for future improvement.
We conduct further studies in Appendix~\ref{sec:longtail} to  illustrate the performance over different classes.

\begin{table}[h]
\centering
\caption{Performance of the models trained with created datasets and the cost of constructing the datasets. The results are averaged over five runs. The gain of \method has passed the statistical test with $p<0.05$.
We also include the performance and cost of using LLM as a zero-shot predictor. 
}
\scalebox{0.7}{%
\begin{tabular}{lccc ccc cc cccc}
\toprule
\multirow{2.5}{*}{\bf Method} & \multicolumn{3}{c}{\bf NYT}    & \multicolumn{3}{c}{\bf Amazon}    & \multicolumn{3}{c}{\bf Reddit} & \multicolumn{3}{c}{\bf StackExchange}     \\ \cmidrule(lr){2-4} \cmidrule(lr){5-7} \cmidrule(lr){8-10}  \cmidrule(lr){11-13} 
& Acc. & F1  & Price/1k      & Acc. & F1    & Price/1k         & Acc. & F1      & Price/1k     & Acc & F1     & Price/1k     \\ \midrule
LLM Zero-Shot  & 74.16 &	69.84 & 5.44 &	59.55 &	54.56	 & 2.11 & 67.00 &	56.66 & 2.89 & 44.70	 & 43.80 & 3.12 \\ \midrule
 
 Gold &  83.80	&81.02	& --- &		82.23 &	81.12	 &--- &		84.22	 &83.38	 &---&	67.56	& 63.28  &---\\ 
\baseline  &75.47	& 76.22 &	0.76 &	57.34	 &56.96	 &0.77 &53.48 &	53.81 &	0.65 &	42.88	 &41.30	 &0.69\\
{MetaPrompt} & {79.58} & {79.83} & {0.87} & {56.35} & {55.98} & {0.84} & {54.61} & {54.30} & {0.74} & {44.81} & {44.02} & {0.83} \\
{\method}       w/o CAF & 80.40 &	80.92  &0.91	 &61.67	 &61.57 &	0.82 & 61.22	 &60.18	  &	0.72 &	45.90	 &44.84	 &0.81 \\
\rowcolor{green!20} {\method}                                       & 81.30 &	82.26   & 1.05 &	66.08	 & 65.65	 & 0.87 &	63.33 &	63.10	 & 0.84 &	48.99	 &47.42 &0.90     \\ 
 \bottomrule
\end{tabular}
}
\label{tab:Main}
\vspace{-1ex}
\end{table}

\subsection{Augmenting existing dataset with generated data}

Here, we merge the generated dataset and the original training set into a single training set, and then test the model performance when it is trained with the merged dataset to see whether the generated dataset can further improve model performance with the original training set available. We present the results in Table~\ref{tab:MainAug}. 
From the table, we can see that the generated dataset is an effective complement to the original training set, since most of the generated datasets introduce performance gain when combined with the original training set, especially our \method which leads to improvement for all the cases. This notable improvement with simple dataset merge may motivate future studies of more advanced ways of using the generated data as augmentations to boost existing dataset.

\begin{table}[!h]
\centering
\caption{Performance of the model trained with the original training set/augmented with the generated dataset. We present the performance gain/drop compared to using the original training set in green/red. 
}
\scalebox{0.7}{%
\begin{tabular}{lccc cc cc cc}
\toprule
\multirow{2.5}{*}{\bf Method} & \multicolumn{2}{c}{\bf NYT}    & \multicolumn{2}{c}{\bf Amazon}    & \multicolumn{2}{c}{\bf Reddit} & \multicolumn{2}{c}{\bf StackExchange}     \\ \cmidrule(lr){2-3} \cmidrule(lr){4-5} \cmidrule(lr){6-7}  \cmidrule(lr){8-9} 
& Acc. & F1       & Acc. & F1           & Acc. & F1         & Acc & F1      \\ \midrule

\baseline & 85.56 {\small\green{+1.76}} &	86.34 {\small\green{+5.32}}	 &	81.85	{\small\red{-0.38}} & 80.23 {\small\red{-0.89}} &	85.11 {\small\green{+0.89}} &	84.88 {\small\green{+1.50}} & 74.53 {\small\green{+6.97}} &	74.23 {\small\green{+10.95}}\\

{Metaprompt} & {87.14 {\small\green{+3.34}}} &	{87.33 {\small\green{+6.31}}}	 &	{82.12	{\small\red{-0.11}}} & {80.14 {\small\red{-0.98}}} &	{84.71 {\small\green{+0.49}}} &	{84.62 {\small\green{+1.24}}} & {76.02 {\small\green{+8.46}}} &	{75.70 {\small\green{+12.42}}}\\

{\method}       w/o CAF  & 85.71 {\small\green{+1.91}} &	87.18 {\small\green{+6.16}} &		82.24	{\small\green{+0.01}} & 80.76 {\small\red{-0.36}} &	85.86 {\small\green{+1.64}} & 85.65 {\small\green{+2.27}} & 75.16 {\small\green{+7.60}} & 74.64 {\small\green{+11.36}}\\

\rowcolor{green!20} {\method} &  87.47 {\small\green{+3.67}} &	88.06 {\small\green{+7.04}} &		83.95 {\small\green{+1.72}} & 83.93 {\small\green{+2.81}} & 86.08 {\small\green{+1.86}} &	85.98 {\small\green{+2.60}} & 76.86 {\small\green{+9.30}} &	76.53 {\small\green{+13.25}}\\ 
 \bottomrule
\end{tabular}
}
\label{tab:MainAug}
\end{table}

\subsection{The budget and sample efficiency of the generated data}
\label{sec:efficiency}
\begin{figure}[!h]
	\centering
	\vspace{-1ex}
 	\subfigure[Amazon Budget]{
		\includegraphics[width=0.24\linewidth]{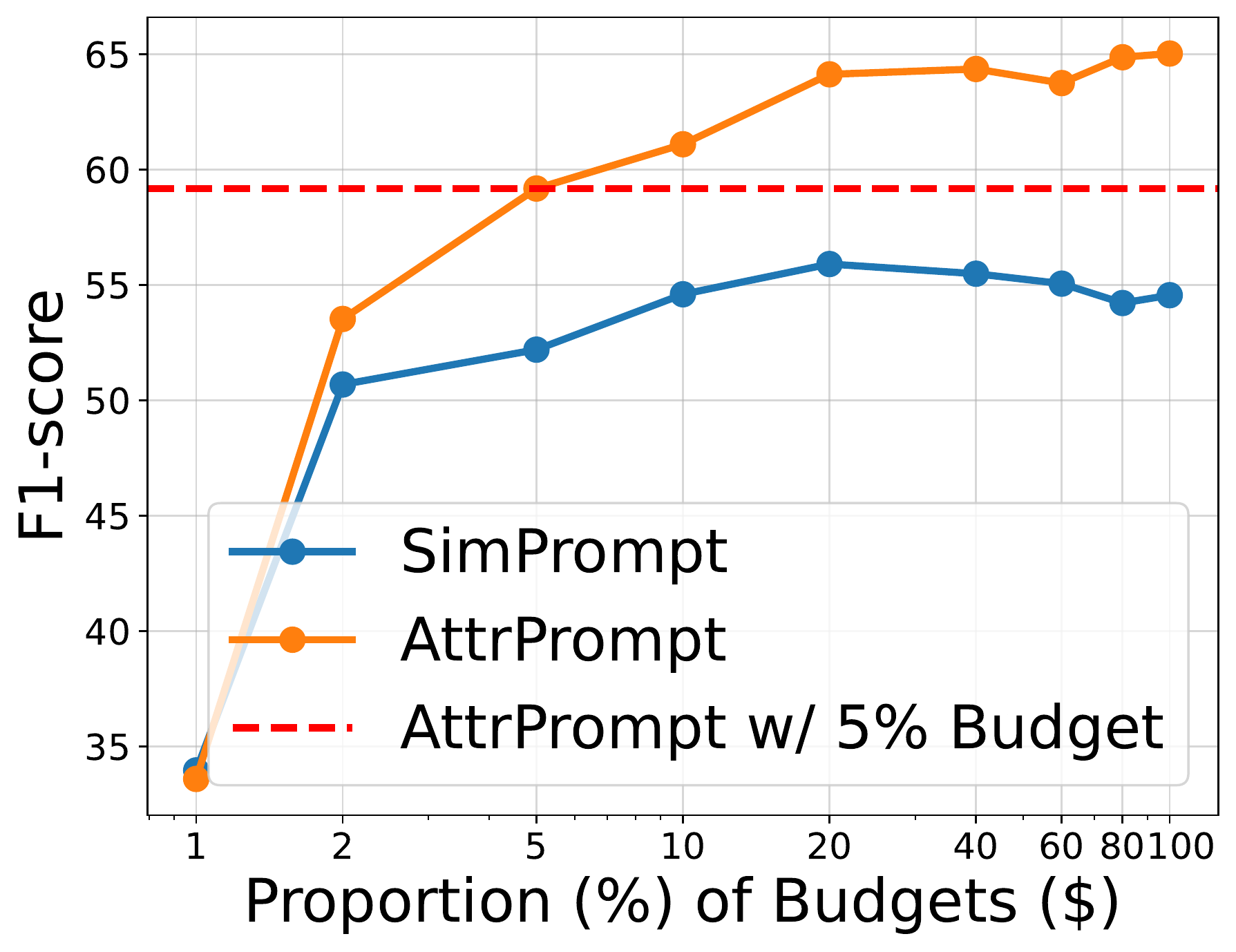}
		\label{fig:budget-amazon}
	} \hspace{-1.5ex} 
	\subfigure[NYT Budget]{
		\includegraphics[width=0.24\linewidth]{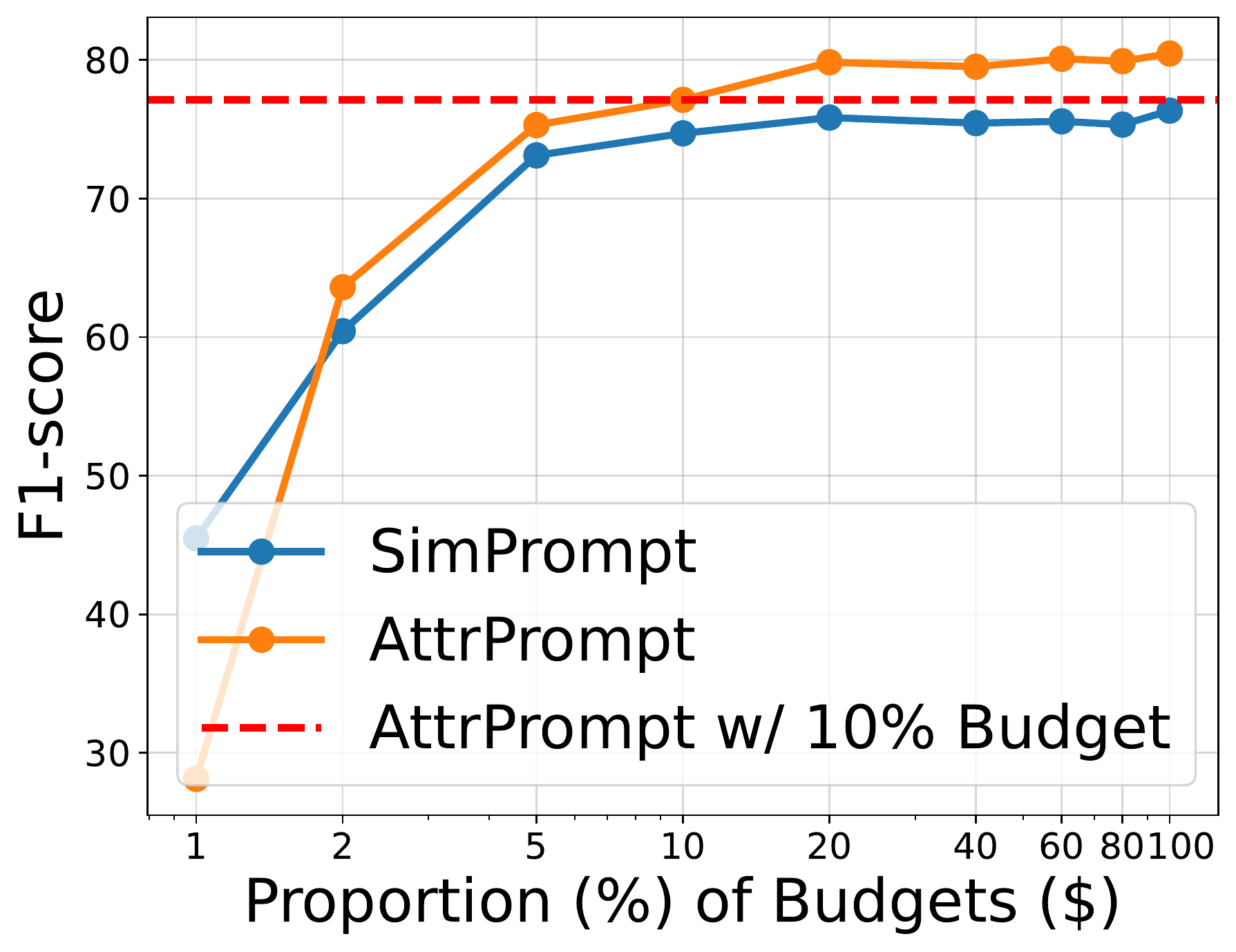}
		\label{fig:budget-NYT}
	}  \hspace{-1.5ex} 
	\subfigure[Reddit Budget]{
		\includegraphics[width=0.24\linewidth]{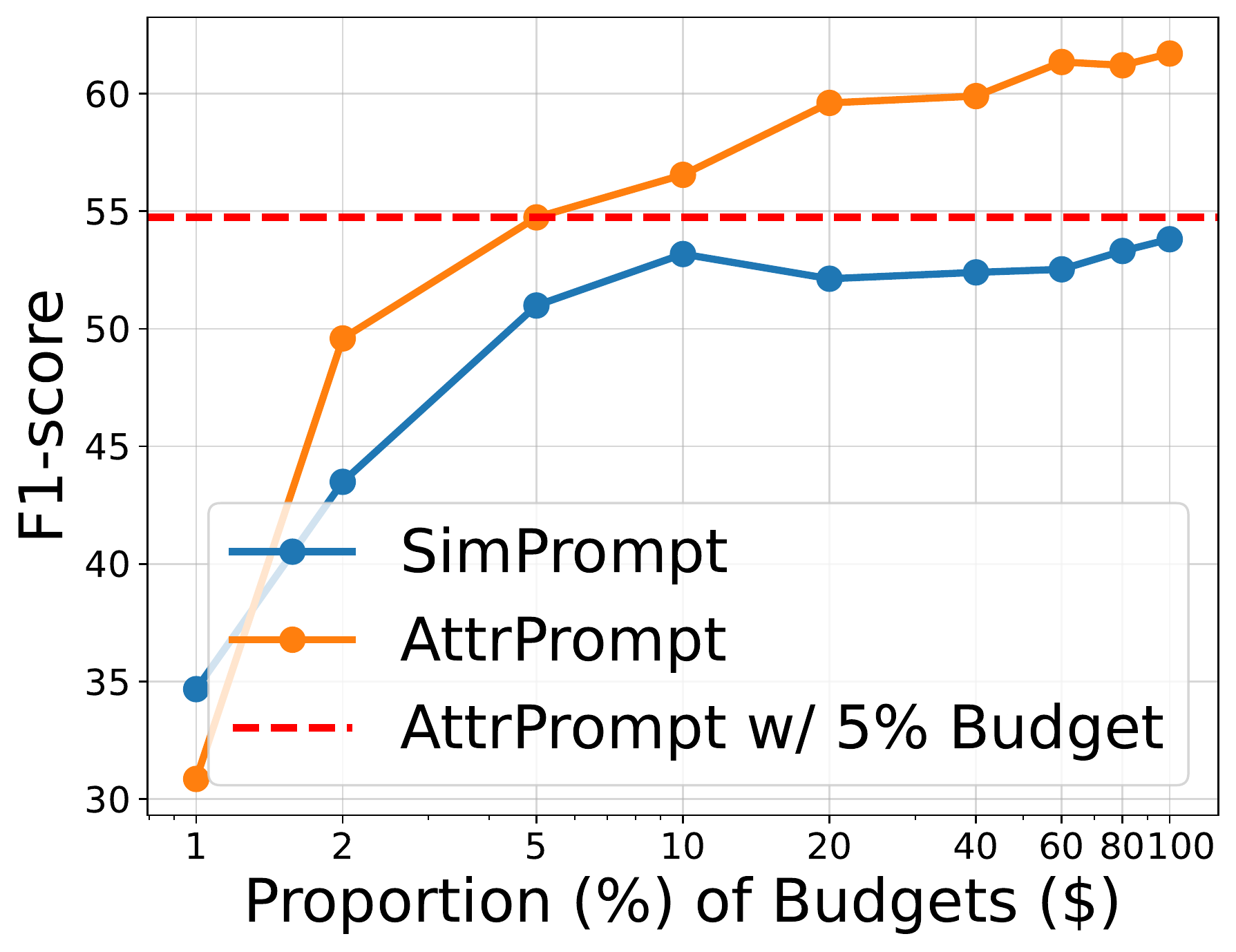}
		\label{fig:budget-reddit}
	} \hspace{-1.5ex} 
	\subfigure[StackExchange Budget]{
		\includegraphics[width=0.24\linewidth]{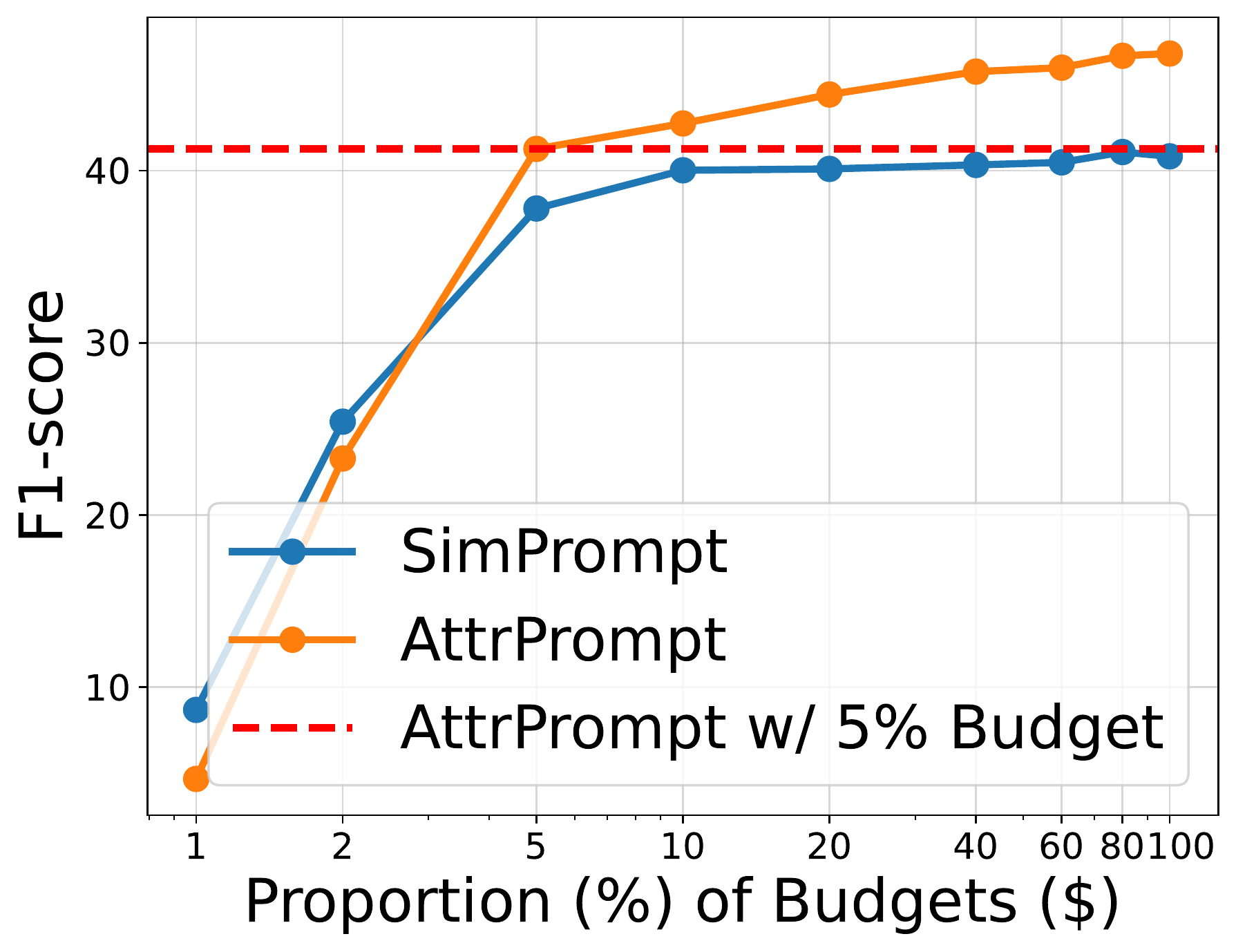}
		\label{fig:budget-stackexchange}
	}  \hspace{-1.5ex} 
	\vspace{-1ex}
	\caption{The comparisons on budget efficiency on four datasets.}\label{fig:budget-1}
\end{figure}

Here, we aim to study two types of efficiency of the generated dataset, \ie, budget efficiency and sample efficiency, on the model performance. 
First, in Figure~\ref{fig:budget-1}, we compare the budget efficiency of \method against that of \baseline. Surprisingly, \method only requires 5\% of budget to be on par with or outperform \baseline with 100\% of budget across all the datasets. This observation highlights the  significance of diverse prompts in the training data generation process.

\begin{figure}[!h]
	\centering
	\vspace{-0.1ex}
 	\subfigure[Amazon Data]{
		\includegraphics[width=0.24\linewidth]{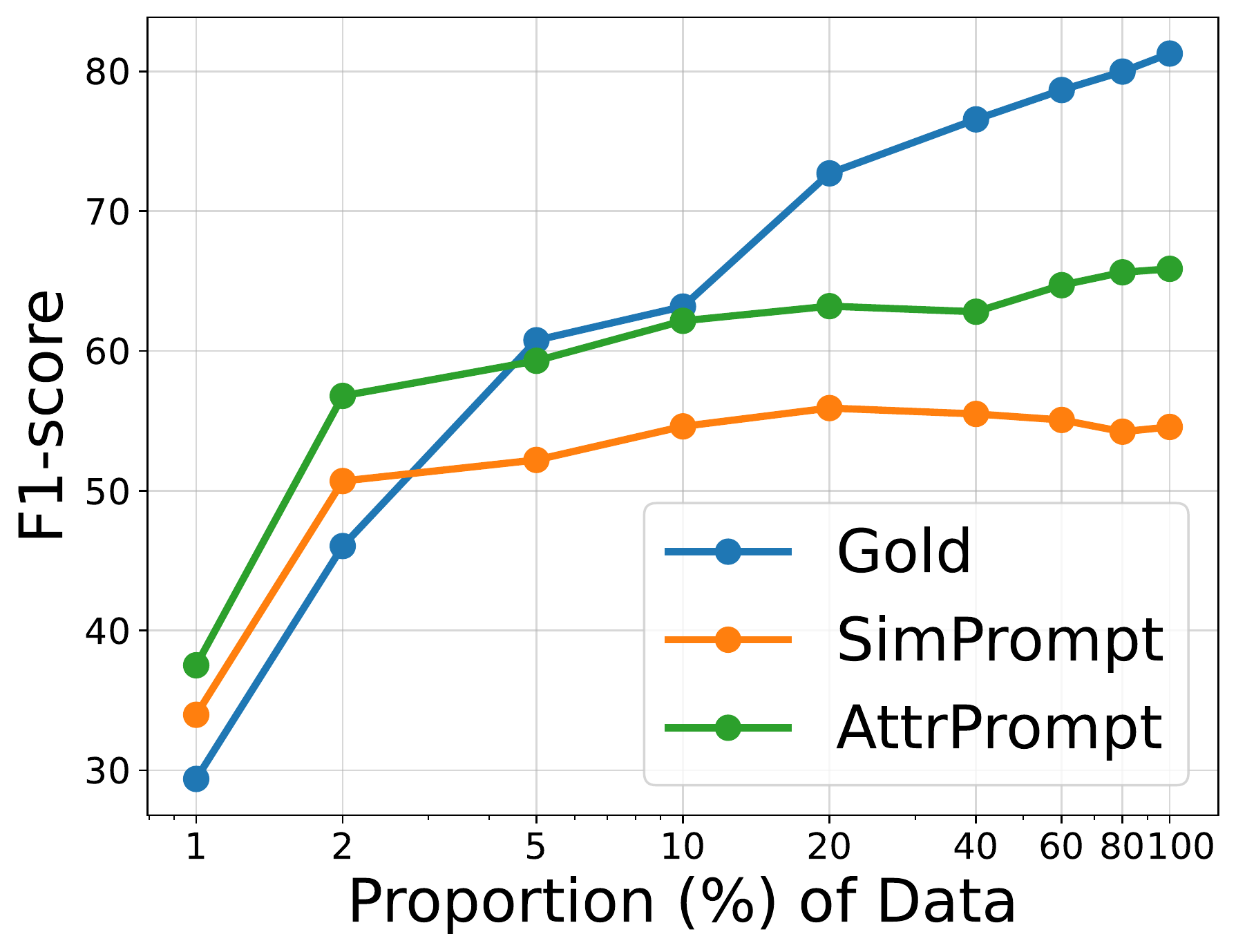}
		\label{fig:data-amazon}
	} \hspace{-1.5ex} 
	\subfigure[NYT Data]{
		\includegraphics[width=0.24\linewidth]{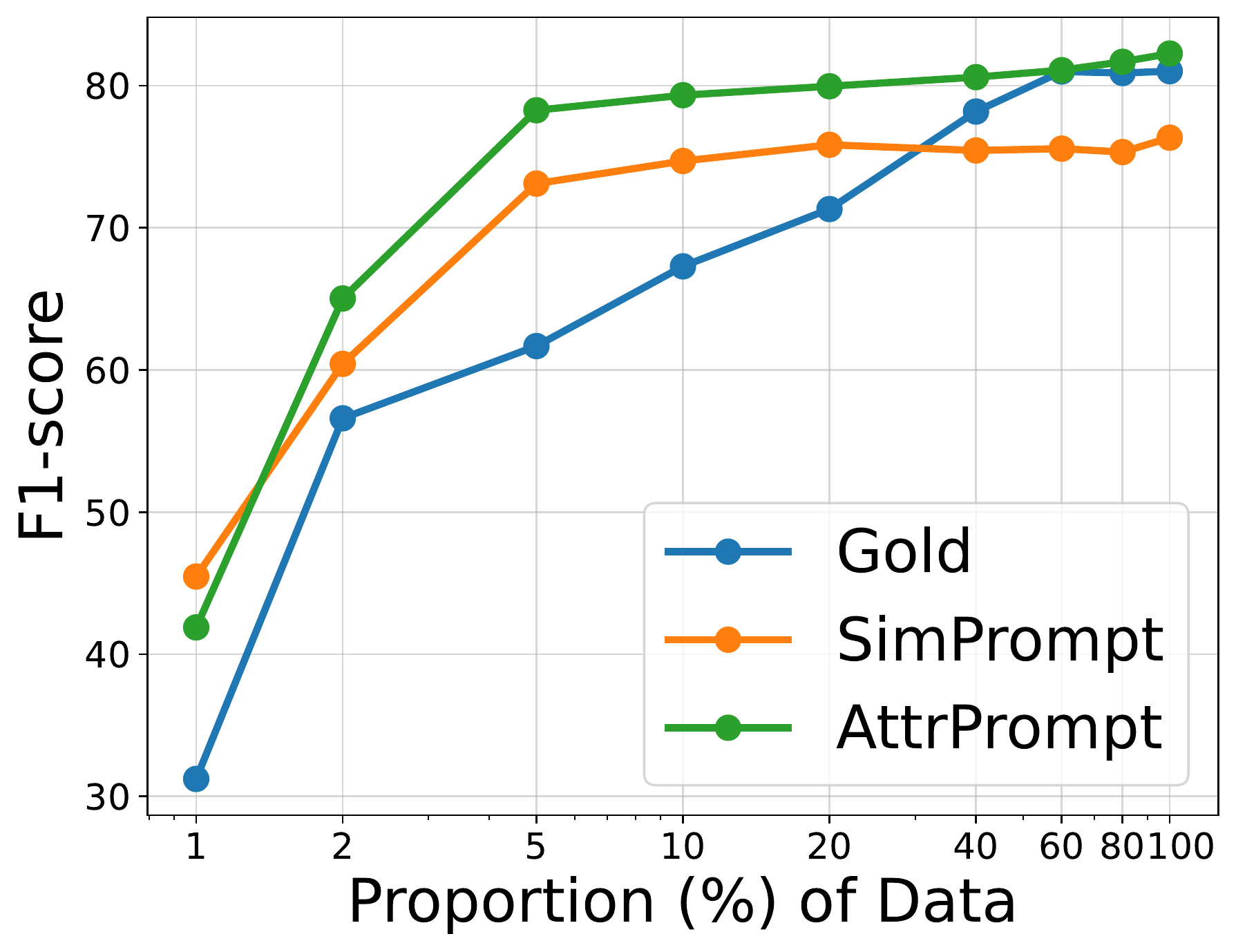}
		\label{fig:data-NYT}
	}  \hspace{-1.5ex} 
	\subfigure[Reddit Data]{
		\includegraphics[width=0.24\linewidth]{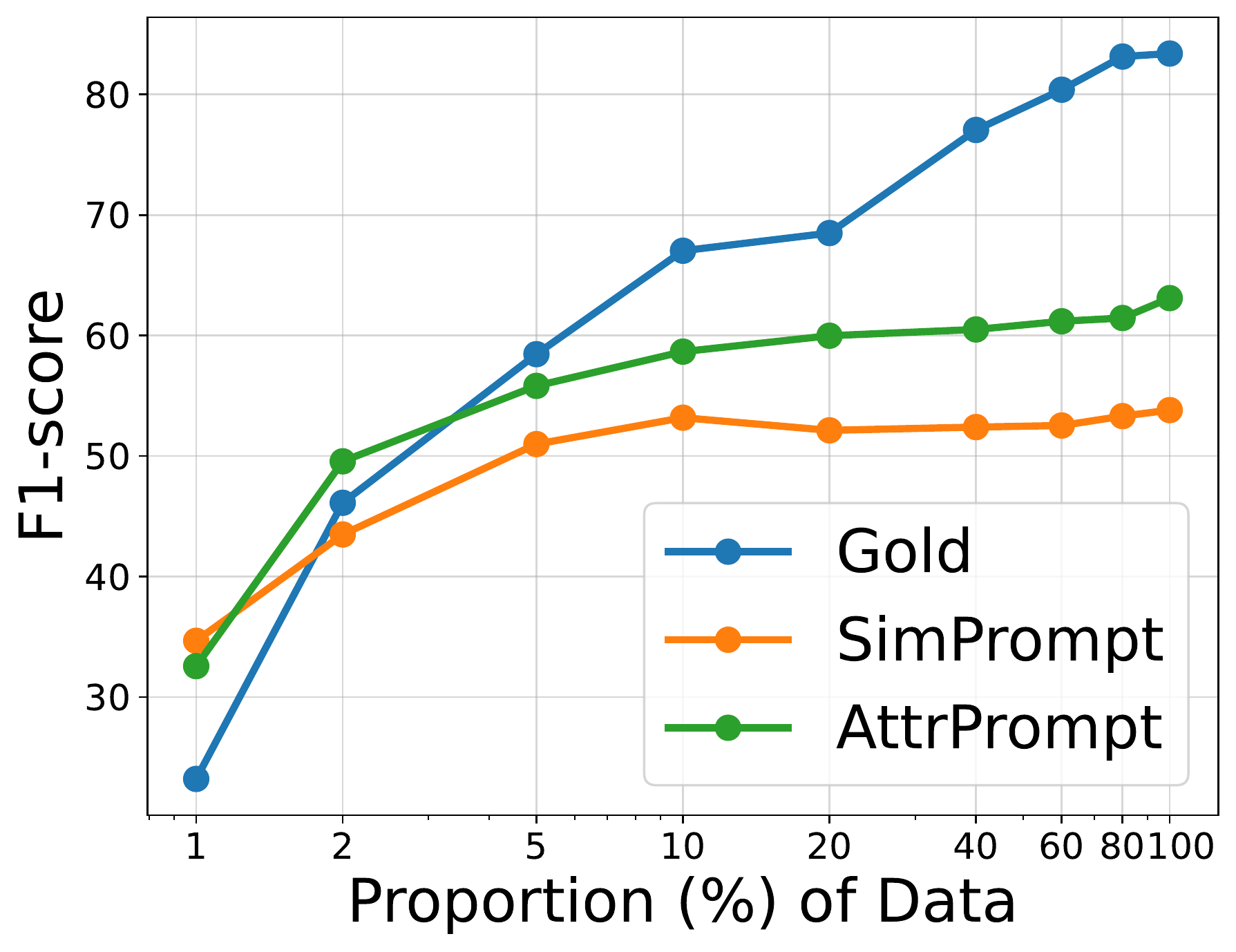}
		\label{fig:data-reddit}
	} \hspace{-1.5ex} 
	\subfigure[StackExchange Data]{
		\includegraphics[width=0.24\linewidth]{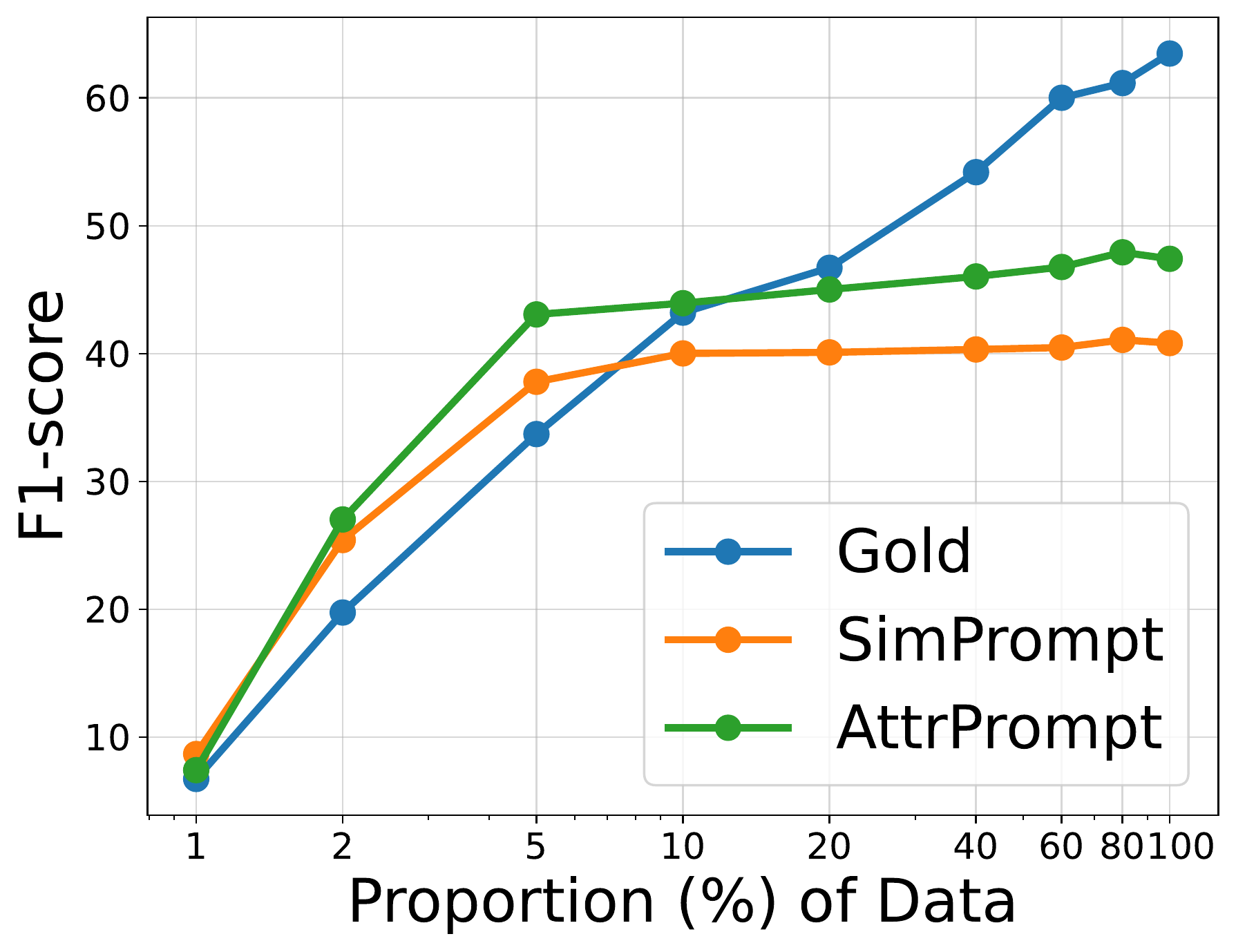}
		\label{fig:data-stackexchange}
	}  \hspace{-1.5ex} 
	\vspace{-1ex}
	\caption{The comparisons on data efficiency on four datasets.}\label{fig:data-1}
\end{figure}

Secondly, we examine the sample efficiency of Gold, \baseline, and \method in Figure~\ref{fig:data-1}. 
While both \baseline and \method exhibit better sample efficiency than Gold in the low-data regime, with superior performance when the dataset size is relatively small, Gold data shows better sample efficiency in the high-data regime.
Overall, \method renders better sample efficiency than \baseline, which suggests that increasing the diversity of the prompts could be an effective way to improve the unsatisfactory data scaling trend of using LLM as data generator~\cite{ye-etal-2022-progen}.

\subsection{The performance with respect to model parameter size}
\label{sec:model}

\paragraph{Effect of the Model Size for LLM Generators.} To study the effect of different LLMs on \method, 
we use other instruction-finetuned GPT models as the generator, namely \texttt{text-ada-001}~\cite{ouyang2022training}, \texttt{text-babbage-001}~\cite{ouyang2022training}, \texttt{text-curie-001}~\cite{ouyang2022training}, and GPT-4~\cite{gpt4} (due to budget constraints, we only generate a subset with 10\% size of the original dataset). 
Under all settings, our model outperforms the direct baseline SimPrompt by a great margin. Besides, the performance is generally better with larger models, as they often have better instruction-following capabilities. In addition, an interesting finding is that for SimPrompt (but not for AttrPrompt), the average performance of using ChatGPT is worse than \texttt{text-curie-001}. This suggests that straightforward class-dependent prompts might not exploit the capabilities of LLMs as effectively as our proposed approaches.

\begin{figure}[!t]
	\centering
	\subfigure[Amazon]{
		\includegraphics[width=0.24\linewidth]{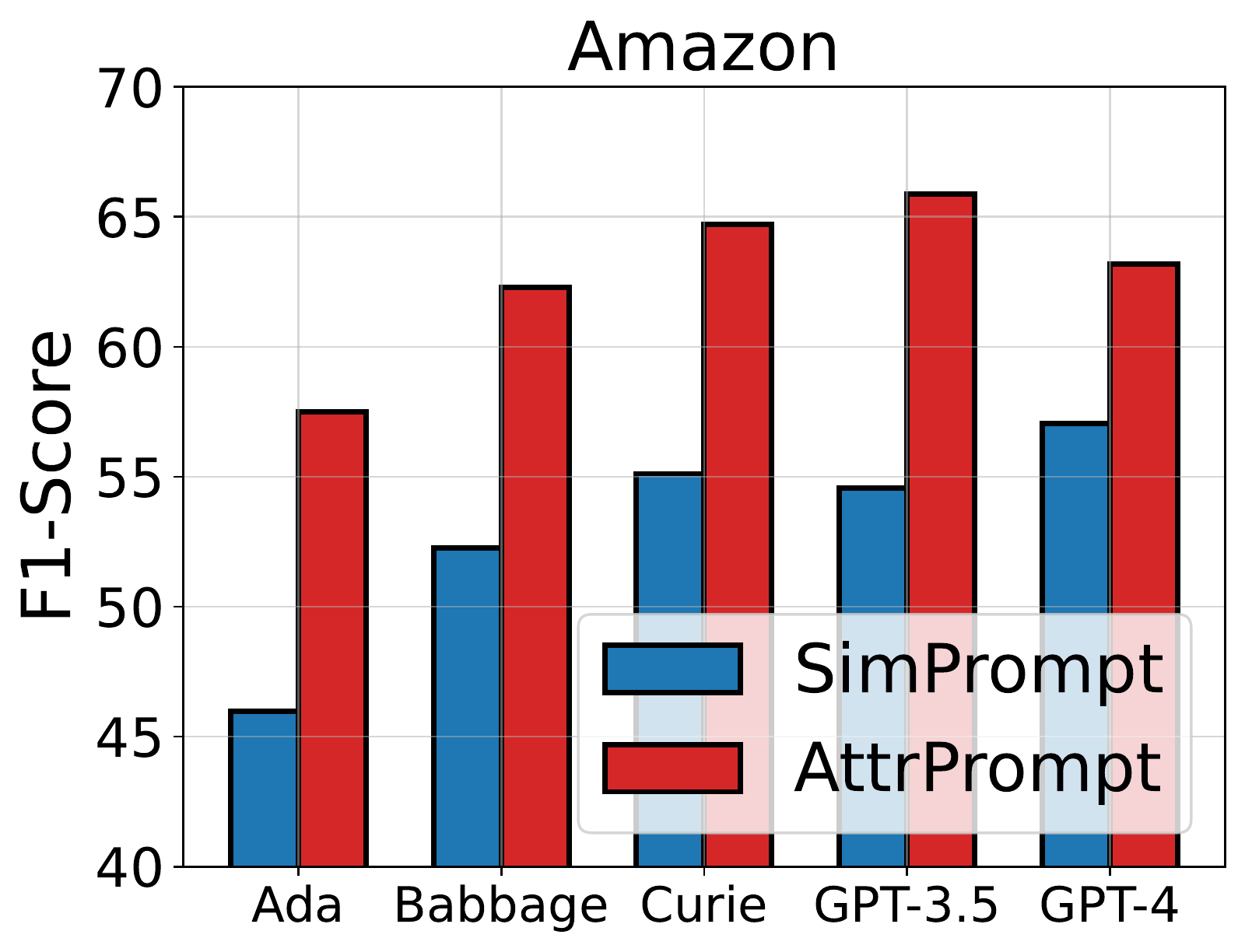}
		\label{fig:gen-params-amazon}
	} \hspace{-1.5ex} 
	\subfigure[NYT]{
		\includegraphics[width=0.24\linewidth]{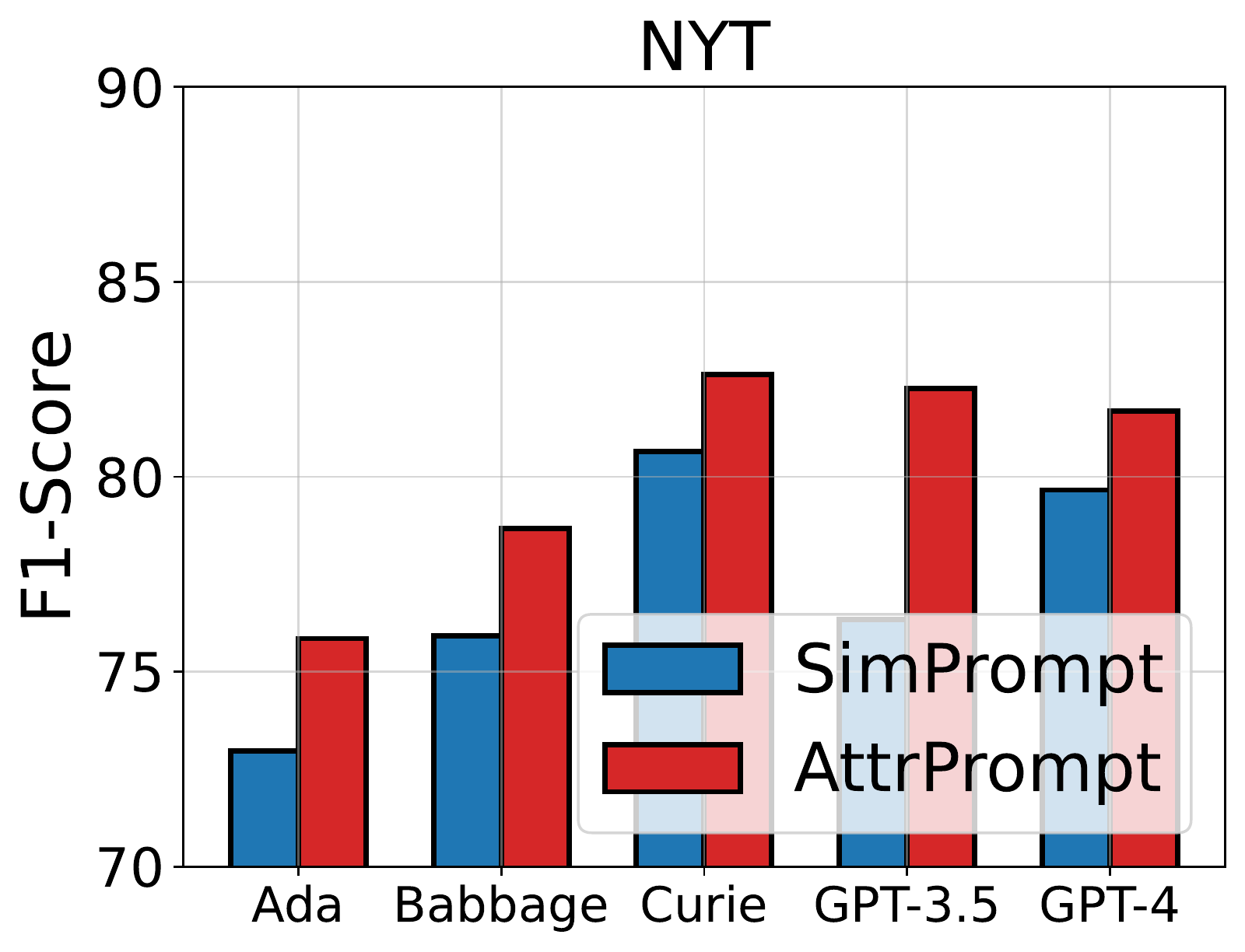}
		\label{fig:gen-params-NYT}
	}  \hspace{-1.5ex} 
	\subfigure[Reddit]{
		\includegraphics[width=0.24\linewidth]{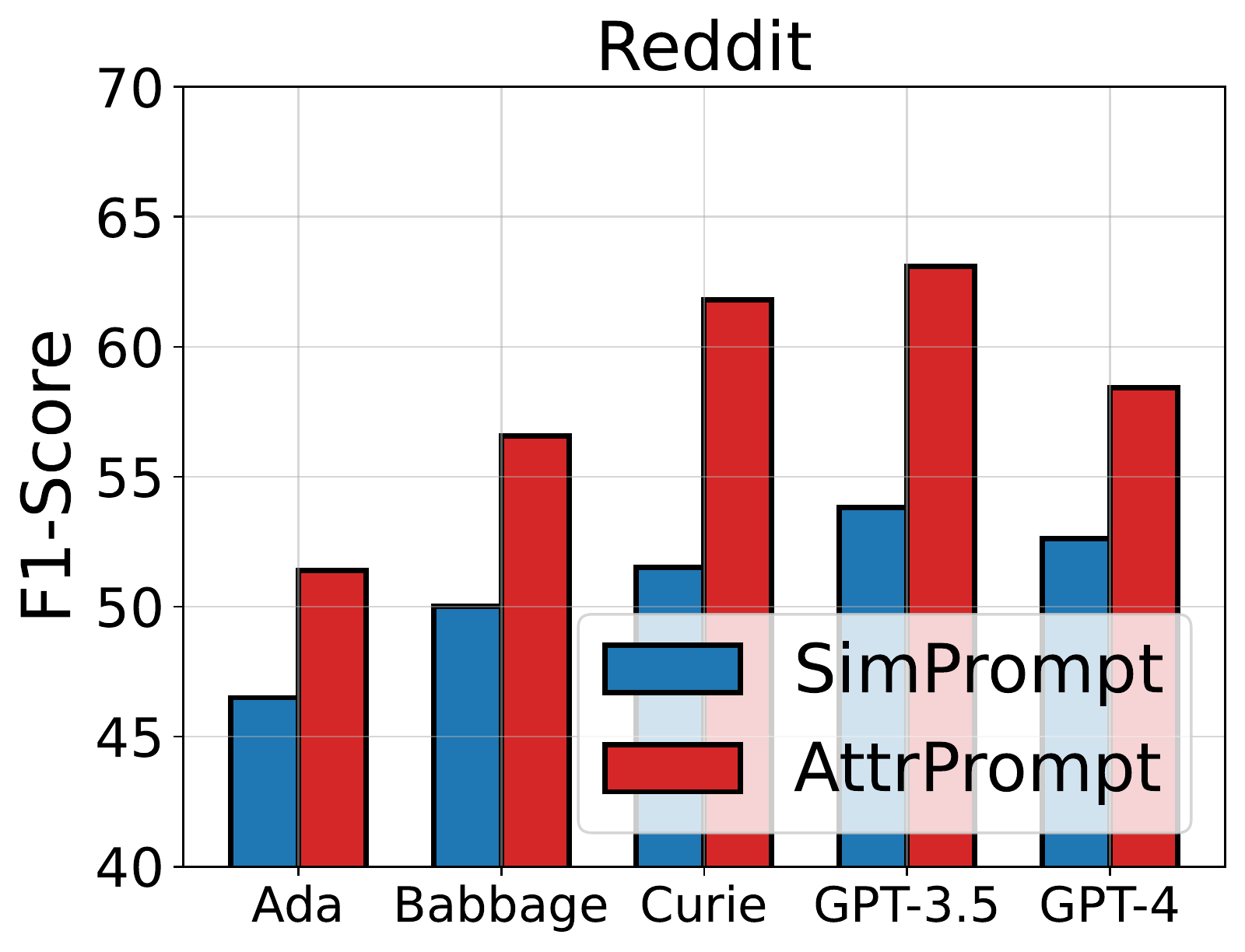}
		\label{fig:gen-params-reddit}
	}  \hspace{-1.5ex} 
	\subfigure[StackExchange]{
		\includegraphics[width=0.24\linewidth]{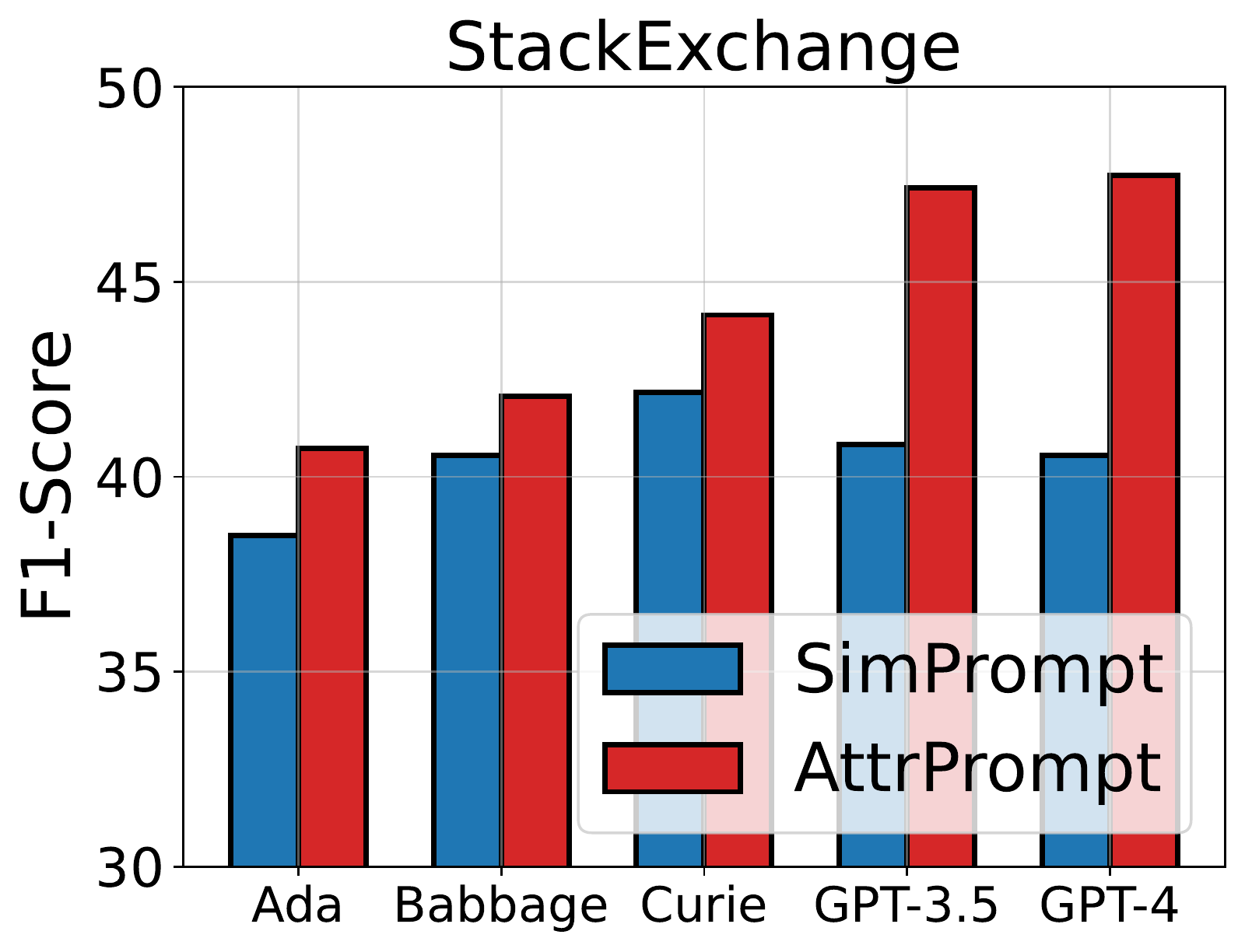}
		\label{fig:gen-params-stackexchange}
	} 
	\vspace{-0.5ex}
	\caption{{The barplot of performance with LLM Generators of different parameter sizes. Note that due to budget limit, for GPT-4 model, the size of the generated dataset is only 10\% of the full set thus the result is not directly comparable with other models.}}\label{fig:gen-params}
\end{figure}

\begin{figure}[!h]
	\centering
	\vspace{-1ex}
	\subfigure[Amazon]{
		\includegraphics[width=0.24\linewidth]{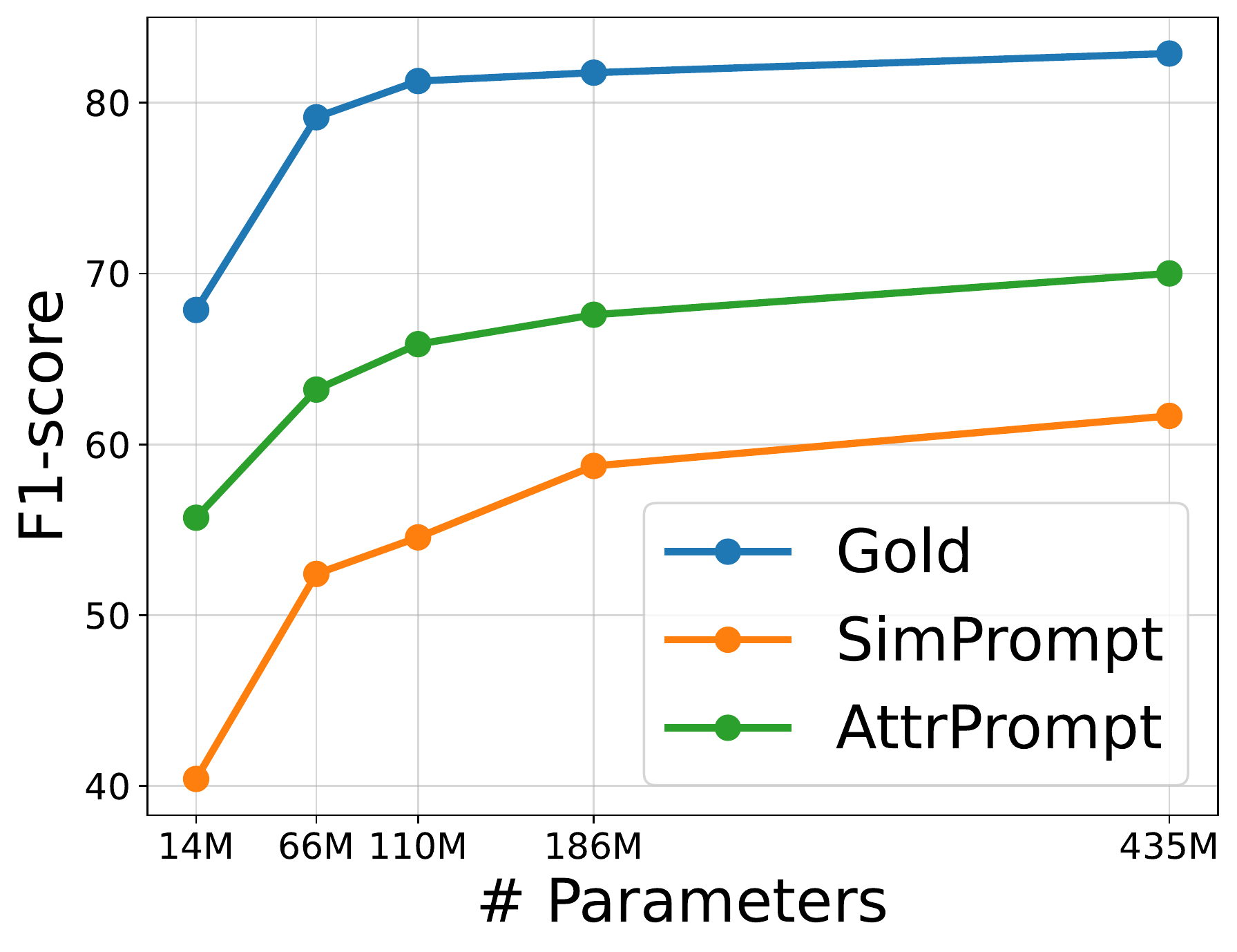}
		\label{fig:params-amazon}
	} \hspace{-1.5ex} 
	\subfigure[NYT]{
		\includegraphics[width=0.24\linewidth]{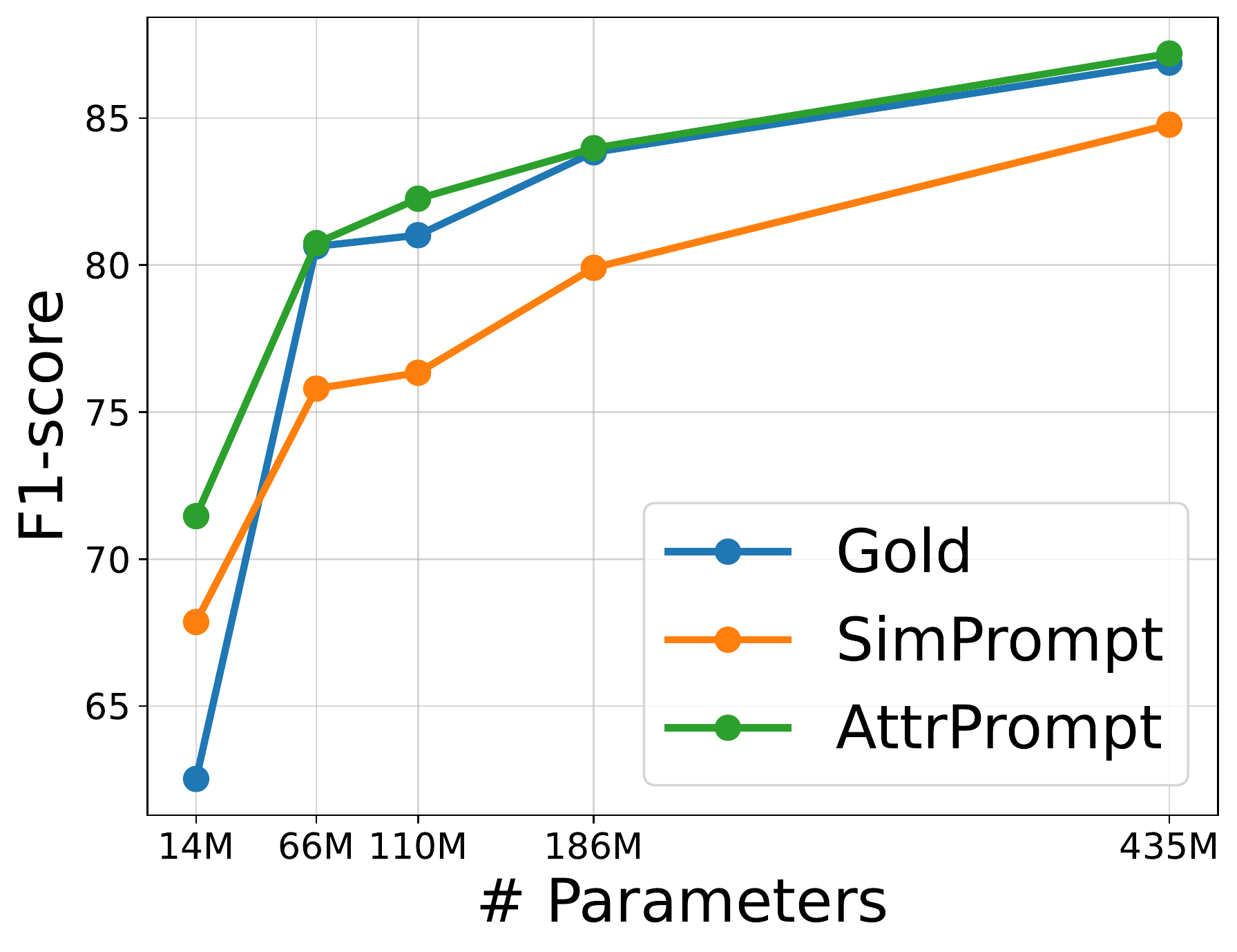}
		\label{fig:params-NYT}
	}  \hspace{-1.5ex} 
	\subfigure[Reddit]{
		\includegraphics[width=0.24\linewidth]{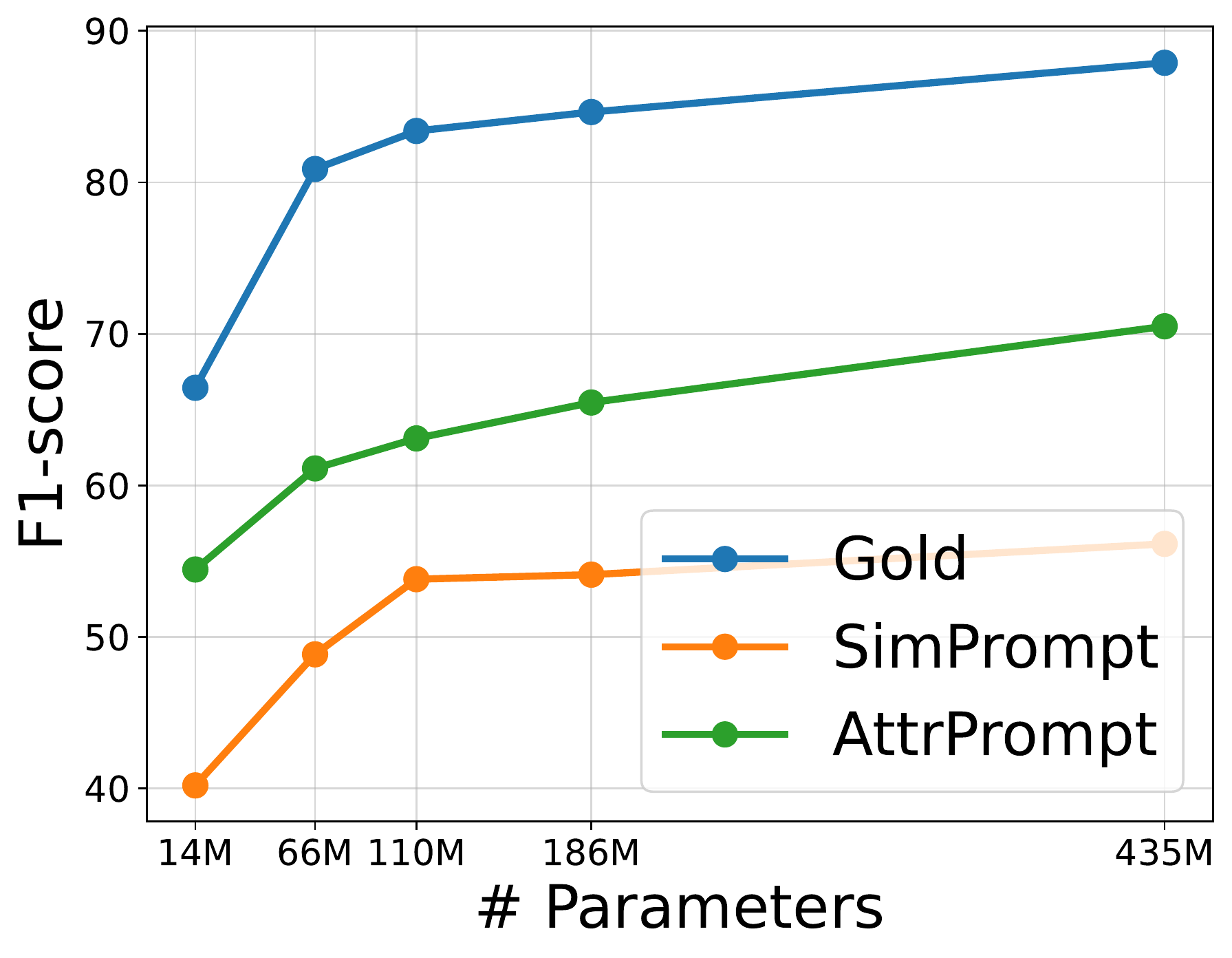}
		\label{fig:params-reddit}
	}  \hspace{-1.5ex} 
	\subfigure[StackExchange]{
		\includegraphics[width=0.24\linewidth]{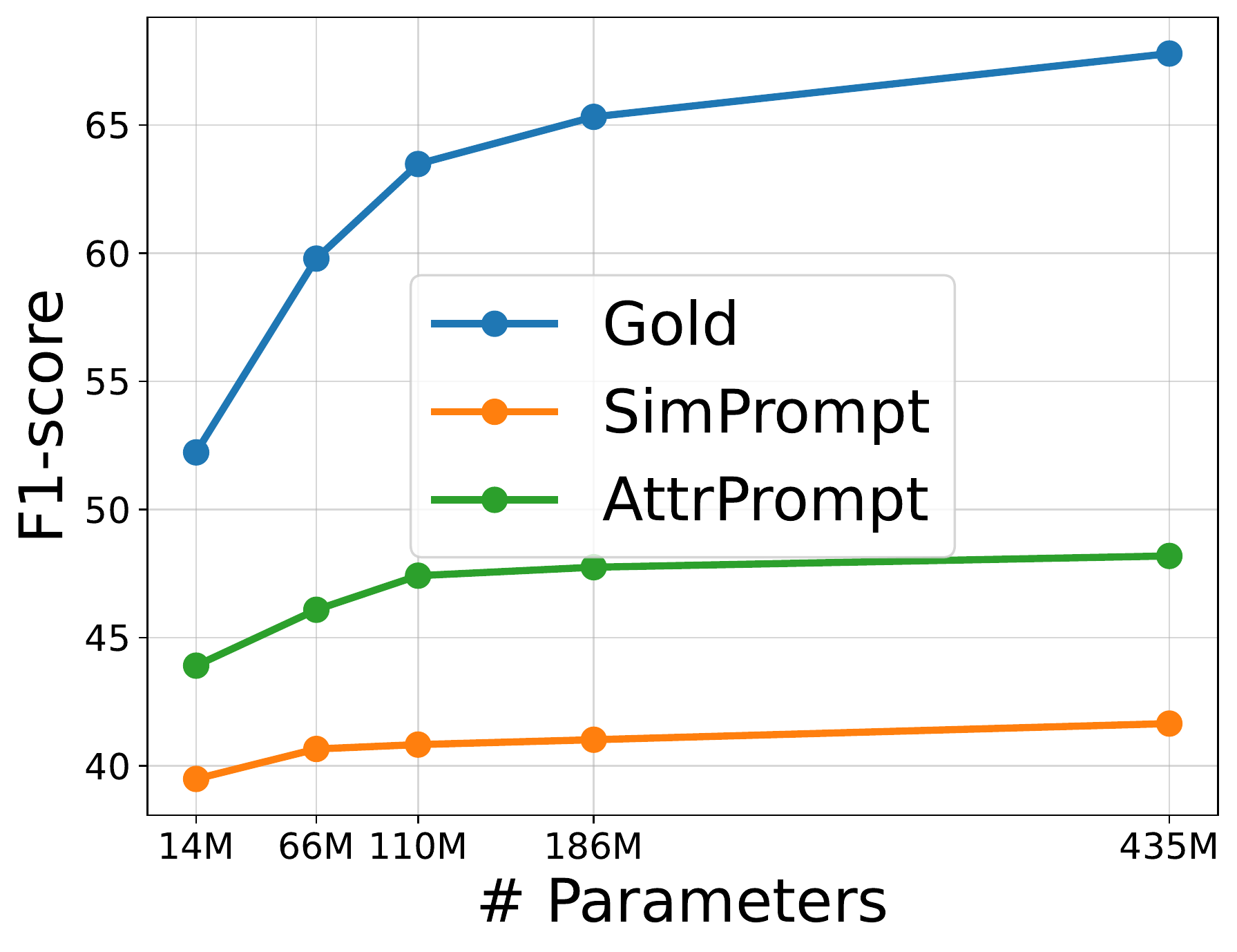}
		\label{fig:params-stackexchange}
	} 
	\caption{The performance curves with classifier models of different parameter sizes.}\label{fig:params}
	\vspace{-2ex}
\end{figure}

\paragraph{Effect of the Model Size for Classifiers.} We experiment with other model choices in addition to the \texttt{BERT-base-uncased}~\cite{devlin2019bert} used throughout the paper. They are \texttt{TinyBERT}~\cite{jiao2020tinybert}, \texttt{DistillBERT}~\cite{sanh2019distilbert}, \texttt{DeBERTa-V3-base}, and \texttt{DeBERTa-V3-Large}~\cite{he2023debertav3} with parameter size from 14M to 435M\footnote{Additionally, we also evaluate 
 under the setting of linear probing~\cite{kumar2022finetuning} which trains a linear classifier over the frozen pretrained features. The results are deferred to Appendix~\ref{sec:linear_prob}.}.
We visualize the results in Figure~\ref{fig:params}. Overall, \method outperforms \baseline by a large margin yet underperforms the Gold across different model choices. 
With a light-weighted backbone of 66M parameters, \method can often outperform the \baseline trained with the model containing 435M parameters.  
This indicates that diversely attributed prompts could help close the performance gap between the Gold and simple class-conditional prompts, and such an improvement is robust to model parameter size.

\subsection{Plugging \method in existing approaches}
\label{sec:combo}
In this section, we demonstrate that \method can be painlessly integrated with prior zero-shot training data generation techniques. 
Table \ref{tab:combo} shows the results for several recently proposed methods, which design additional techniques based on the noisy-robust loss to further reduce the effect of noisy labeled data~\cite{gao2023selfguided,supergen,ye-etal-2022-progen}, and leverage in-context examples for data generation~\cite{ye-etal-2022-progen}. 
Despite these approaches achieving notable performance gains on simple binary classification tasks, their gains become more marginal for fine-grained classification: the performance gain is less than 2\% for all methods on two datasets. 
Instead, using \method lead to consistent performance boosts (more than 5\% in all cases) for those approaches, indicating that compared with label noise, \emph{data diversity} is a more crucial bottleneck for existing dataset generation methods.

More interestingly, \method even benefits dataset generation approaches that do not use LLMs. To demonstrate this, we use the LLM-generated contents (subtopics for NYT and product name for Amazon) to enrich the label names used in ReGen~\cite{yu2023zero}, a retrieval-based approach for training data generation. With the expanded label names, \method largely improves (14\%--26\% absolute gain) the performance of ReGen on fine-grained classification tasks. These results justify the advantage of \method for serving as a generic plug-in module for existing training data generation approaches.





\begin{table}[h]
\centering 
\caption{Performance comparison when \method serves as a plug-in  for existing approaches. 
}
\renewcommand\arraystretch{0.88}
\fontsize{7.5}{9.5}\selectfont \setlength{\tabcolsep}{0.3em}
\resizebox{0.87\linewidth}{!}{%
\begin{tabular}{@{}llc
>{\columncolor{green!15}}c c
>{\columncolor{green!15}}c c
>{\columncolor{green!15}}c c
>{\columncolor{green!15}}c c
>{\columncolor{green!15}}c @{}}
\toprule
                         &      & \multicolumn{1}{c}{SuperGen} & \multicolumn{1}{c}{\cellcolor{green!15}\begin{tabular}[c]{@{}c@{}}SuperGen\\ w/ AttrPrompt\end{tabular}} & \multicolumn{1}{c}{SunGen} & \multicolumn{1}{c}{\cellcolor{green!15}\begin{tabular}[c]{@{}c@{}}SunGen\\ w/ AttrPrompt\end{tabular}} & \multicolumn{1}{c}{ProGen} & \multicolumn{1}{c}{\cellcolor{green!15}\begin{tabular}[c]{@{}c@{}}ProGen\\ w/ AttrPrompt\end{tabular}} & \multicolumn{1}{c}{ReGen} & \multicolumn{1}{c}{\cellcolor{green!15}\begin{tabular}[c]{@{}c@{}}ReGen\\ w/ AttrPrompt\end{tabular}} \\ \midrule
                         & Acc.   & 76.11 & 82.05 & 75.82 & 81.65 &  77.05 & 80.93   &  70.01 & 82.18  \\
\multirow{-2}{*}{NYT}    & F1   & 76.80 & 82.62 & 76.52 & 82.70 & 76.70 & 81.32 &    68.14       &  82.50         \\\midrule
                         & Acc.  &58.17 & 66.76 & 54.30 & 63.89  & 58.40 & 66.43 & 34.70 & 58.40      \\
\multirow{-2}{*}{Amazon} & F1   &56.06 & 66.33 & 53.50 & 63.76  & 56.95 & 66.02 & 30.93  & 56.00     \\ \bottomrule
\end{tabular}}\label{tab:combo}
\vspace{-2ex}
\end{table}



\vspace{-0.5ex}
\subsection{Extension to multi-label classification}
\vspace{-0.5ex}
In this section, we take the first attempt to extend the paradigm of using the LLM as a training data generator to the more challenging multi-label classification setting.
In particular, we adopt the arXiv dataset \cite{clement2019use} consisting of 98 fine-grained classes, on which we apply both \baseline and \method. 
Following \cite{guo2019breaking,shen2021taxoclass}, we consider different  evaluation metrics including Micro/Macro-F1, Precision@$k$, Normalized Discounted Cumulative Gain (nDCG@$k$), and Mean Reciprocal Rank (MRR). The experimental details are in Appendix~\ref{app:data}.
We present the results in Table~\ref{tbl:multilabel}.
Similar to our findings for single-label classification, \method largely outperforms \baseline across all the metrics, which not only strengthens the superiority of \method but also opens the door to using LLM as a training data generator for future research in multi-label classification.
\begin{table}[h]
\centering
\vspace{-1ex}
\caption{Multi-label classification performance (in \%) on the arXiv dataset.}
\scalebox{0.8}{
\begin{tabular}{@{}lccccccc}
\toprule
\bf Method     & Macro F1 & Micro F1 & Precision@1 & Precision@5 & NDCG@5 & MRR & Price/1k  \\ \midrule
Gold       &     27.34     &  58.22        &   73.71          &  27.00           &  79.70      &   82.16  & ---  \\
SimPrompt  & 21.03    & 26.75    & 37.00       & 15.08       & 42.49  & 49.60  & 1.41 \\
\rowcolor{green!15} AttrPrompt & 27.10    & 37.88    & 49.27       & 18.79       & 54.74  & 61.23  & 1.53\\ \bottomrule
\label{tbl:multilabel}
\vspace{-4ex}
\end{tabular}}
\end{table}

\section{Limitation, Social Impact, and Future Work}

\subsection{Limitation}
\label{sec:limitation}
One limitation of this research is that the investigation focuses primarily on text classification. While this provides valuable insights into the performance of attributed prompts compared to simple class-conditional prompts in these specific contexts, the generalizability of the findings to other data types and tasks remains to be explored. In addition, the \method assumes the label name of the target task following prior works~\cite{supergen,ye-etal-2022-zerogen} and requires some extent of human effort in creating  data attributes.

In addition, one issue with LLM-based training data generation is the phenomenon of hallucination, wherein the model generates information that may sound plausible but is, in fact, not accurate or grounded in reality~\cite{bang2023multitask}. This can lead to the propagation of misinformation, particularly in sensitive areas such as news generation where accuracy is of utmost importance. To alleviate this issue, it is possible to leverage additional fact-checking mechanisms to cross-verify the generated text with a reliable knowledge base or dataset~\cite{peng2023check}. Furthermore, incorporating an additional layer of human review can also help in mitigating hallucinations and ensuring the faithfulness of generated outputs. 
\subsection{Social impact}
\label{sec:social_impact}
The findings of this research have significant implications for the development of trustworthy machine learning models. By addressing the limitations of simple class-conditional prompts and exploring the potential of attributed prompts, the research aims to mitigate biases and enhance the diversity of generated training data. This has a positive social impact as it contributes to the development of more robust and fair machine learning models, reducing the potential for biased decision-making systems and promoting inclusivity in various domains.  
We do not involve human subjects research and do not contain any personally identifiable information.
Possible misuse may lead to negative outcomes, such as intentionally introducing biases to the generated dataset via attributes.

\subsection{Future work}

\paragraph{Broader Coverage of Task and Modalities.} In future work, it would be valuable to extend the investigation to other types of classification tasks beyond text classification. Exploring the effectiveness of attributed prompts in image classification, audio classification, or other modalities could provide insights into the broader applicability and effectiveness of \method.

\paragraph{Automatic Prompt Optimization.} In this work, we aim to demonstrate the usefulness of attributes and why they are useful through the lens of diversity and bias, and we believe this opens the door for future work to further explore the methodology for automatic attribute discovery and selection for optimal performance, such as exploring automated or semi-automated methods for identifying high-quality attribute dimensions and values. 
One promising direction is to incorporate a few labeled examples as demonstrations for generating better training data. Under this scenario, leveraging Language Models can aid in automatically detecting diverse attributes, modeling the high-order interactions among attributes as well as estimating attribute value distributions for each class, which further reduces the human efforts for selecting key attributes. 


\section{Conclusion}
We delve into the realm of training data generation using complex, attributed prompts, which possess the potential to produce a wide range of diverse and attributed generated data. Specifically, we focus on datasets characterized by diverse domains and high cardinality and class-imbalance, and our results demonstrate the superior performance of attributed prompts compared to simple class-conditional prompts. Furthermore, we present a comprehensive empirical study on training data generation that covers essential aspects such as bias, diversity, and efficiency.



\bibliographystyle{plain}
\bibliography{ref}  

\clearpage

\appendix

\section{Dataset Preprocessing}
\label{app:data}

\subsection{Multi-class Classification}
For Reddit and StackOverflow, we select the classes with more than $65$ examples from the original corpus as the target set of topics. 
For each dataset, we use $50$ examples per class for the test set and no more than $10$ examples for the validation set ($10$ for NYT/Amazon and $5$ for Reddit/StackOverflow). The remaining data is used to compose the gold training set. 
It is worth noting that, some of the class names on Reddit may contain toxic information. To eliminate their effects, we filter label names with Detoxify\footnote{{\url{https://github.com/unitaryai/detoxify}}}, a tool that leverages the multilingual XLM-RoBERTa~\cite{Conneau2019UnsupervisedCR} for toxic comment identification. 
We follow \cite{gadre2023datacomp} to use a threshold of 0.1 to filter out potentially toxic topic classes.

\subsection{Multi-label Classification}
For the arXiv multi-label classification, we use the arXiv labels from the website\footnote{\url{https://arxiv.org/}} as the label names. We select topics under the category of \emph{Physics}, \emph{Mathematics}, \emph{Computer Science}, \emph{Quantitative Biology}, and \emph{Quantitative Finance}. 
Besides, we remove the topic of \emph{General Physics}, \emph{General Mathematics}, \emph{General Literature}, \emph{Other Computer Science}, \emph{Other Quantitative Biology} as they only contain generic topics without concrete semantics. 
In total, there are 98 classes in general. 

To generate examples using the \baseline approach, we follow a specific procedure. For each document, we begin by sampling from the distribution of label numbers, utilizing prior knowledge obtained from the original training set. This enables us to determine the precise number of labels, denoted as $n$, that should be associated with the document. Once we have determined $n$, we proceed to randomly select $n$ classes from the original set of 98 labels. These chosen classes are then utilized as the input for prompting the LLM, forming a foundation for generating the examples.

To generate examples using the \method approach, we employ a distinct methodology. For each document, we consider a set of attributes, namely \texttt{subtopics}, \texttt{technique}, \texttt{writing style}, and \texttt{length}. It is important to note that, unlike multi-class classification, where an additional step is typically taken to filter out irrelevant subtopics, our multi-label classification process follows a different approach. 
In our method, we introduce an additional step called \emph{merging} to handle the different subtopics. 

To determine the merging of subtopics, we utilize the fuzzy matching score\footnote{\url{https://github.com/seatgeek/thefuzz}}, which calculates the similarity between the subtopics. Setting the threshold at $\gamma=90$, we proceed to merge subtopics whenever their similarity score exceeds this threshold. Consequently, specific subtopics may be associated with multiple classes, providing a more comprehensive representation of the document content. 
When generating text utilizing these subtopics, the corresponding labels are determined by the classes associated with the specific subtopics. This ensures that the generated examples accurately reflect the relevant classes based on the subtopic associations.

\section{Implementation Details}
\label{app:implement}

\subsection{Hardware information}
All experiments are conducted on \emph{CPU}: Intel(R) Core(TM) i7-5930K CPU @ 3.50GHz and \emph{GPU}: NVIDIA GeForce RTX A5000 GPUs using python 3.8, Huggingface 4.6.0 and Pytorch 1.10. 

\subsection{Parameter configuration}
We follow \cite{Peng2023InstructionTW} to keep the  parameter $\operatorname{top\_p}=1.0$ and temperature $t=1.0$ for calling ChatGPT APIs~\cite{chatgpt} for the training data generation part. 
For finetuning the classifier, we optimize the model using AdamW~\cite{loshchilov2018decoupled} with a linear warmup of the first 5\% steps and linear learning rate decay. The maximum number of tokens per sequence is 128. Table~\ref{tab:hyperpara} lists the hyperparameters used for {\method} and {\baseline}. 
For the generated synthetic dataset, we stick to the strict zero-shot learning setting~\cite{supergen}, train all the models for 6 epochs and use the model from the last epoch \emph{without using the validation set} for evaluation.
For the original clean dataset, we train all models for 6 epochs and use the model with the best performance on the validation set for evaluation. 

\begin{table*}[h]
\centering
\resizebox{0.95\columnwidth}{!}{
\begin{tabular}{*{10}{c}}
\toprule
\bf Backbone &  \bf Learning Rate $\operatorname{lr}$ & \bf Batch Size & \bf  Training Epochs $E$ &\bf  Weight Decay&\bf  Warmup Ratio \\
\midrule
 \texttt{BERT-base-uncased}~\cite{devlin2019bert} &  5e-5  & 32   & 6 &  1e-4&   6\% \\
  \texttt{TinyBERT}~\cite{jiao2020tinybert} &  1e-4  & 32   & 6 &  1e-4&   6\% \\
 \texttt{DistilBERT-base-uncased}~\cite{sanh2019distilbert} &  5e-5  & 32   & 6 &  1e-4&   6\% \\
 \texttt{DeBERTa-V3-base}~\cite{he2023debertav3} &  5e-5  & 32   & 6 &  1e-4&   6\%  \\
 \texttt{DeBERTa-V3-large}~\cite{he2023debertav3} &  2e-5  & 32   & 6 &  1e-4&   6\% \\
\bottomrule
\end{tabular}
}
\caption{
Hyperparameters for fine-tuning classifiers on different tasks.
}
\label{tab:hyperpara}
\end{table*}

\begin{table*}[h]
\centering
\resizebox{0.95\columnwidth}{!}{
\begin{tabular}{*{2}{c}}
\toprule
\bf Model &  \bf Link \\
\midrule
 \texttt{BERT-base-uncased}~\cite{devlin2019bert} &  \url{https://huggingface.co/bert-base-uncased}  \\
  \texttt{TinyBERT}~\cite{jiao2020tinybert} &  \url{https://huggingface.co/huawei-noah/TinyBERT_General_4L_312D}   \\
 \texttt{DistilBERT-base-uncased}~\cite{sanh2019distilbert} &  \url{https://huggingface.co/distilbert-base-uncased}  \\
 \texttt{DeBERTa-v3-base}~\cite{he2023debertav3} &  \url{https://huggingface.co/microsoft/deberta-v3-base} \\
 \texttt{DeBERTae-v3-large}~\cite{he2023debertav3} &  \url{https://huggingface.co/microsoft/deberta-v3-large}  \\ \midrule
\texttt{SentenceBERT}~\cite{reimers2019sentence} &  \url{https://huggingface.co/sentence-transformers/all-mpnet-base-v2}  \\
 \texttt{SimCSE}~\cite{gao2021simcse} &  \url{https://huggingface.co/princeton-nlp/sup-simcse-bert-base-uncased} \\
 \texttt{COCO-DR}~\cite{cocodr} &  \url{https://huggingface.co/OpenMatch/cocodr-base-msmarco}  \\ 
 
\bottomrule
\end{tabular}
}
\caption{The reference of model checkpoints used in this study.}
\label{tab:ckpt}
\end{table*}

\subsection{Model checkpoint}
All the checkpoint used in fine-tuning and linear probing experiments are listed in table \ref{tab:ckpt}.

\begin{table}[t]
\caption{
Performance of linear probing with different pre-trained encoders. 
}
\centering
\scalebox{0.8}{
\begin{tabular}{llcc cc cc cc}
\toprule
\bf {Sequence Encoder}                &  \bf    Method       & \multicolumn{2}{c}{\bf NYT} & \multicolumn{2}{c}{\bf Amazon}                             & \multicolumn{2}{c}{\bf Reddit} & \multicolumn{2}{c}{\bf StackExchange} \\
                          &            & Acc        & F1         &  Acc & F1 & Acc          & F1          & Acc             & F1              \\ \midrule
                          & Gold       & 65.39      & 70.26      & 66.73                      & 68.76                     & 55.73        & 58.00       & 17.17           & 17.68           \\
                          & SimPrompt  & 60.69      & 62.26      & 37.95                      & 38.07                     & 27.41        & 27.60       & 12.27           & 13.71           \\
                        & {MetaPrompt} & 58.43	&57.05	&27.52	&24.43	&23.93&	23.99	&11.40&	12.60\\
\multirow{-4}{*}{BERT-CLS~\cite{devlin2019bert}}    & AttrPrompt & 62.38      & 64.17      & 48.18                      & 49.46                     & 45.03        & 45.95       & 17.33           & 17.64           \\\midrule
                          & Gold       & 81.16      & 84.43      & 80.95                      & 83.30                     & 80.76        & 81.37       & 55.13           & 54.25           \\
                          & SimPrompt  & 84.32      & 85.91      & 67.55                      & 68.29                     & 69.94        & 70.62       & 45.37           & 47.79           \\
                           & {MetaPrompt} & 87.65	&87.11 &	67.16 &	65.78	 &79.26	 &69.15	 &45.71	 &47.54\\
\multirow{-4}{*}{SentenceBERT~\cite{reimers2019sentence}}   & AttrPrompt & 88.77      & 88.36      & 68.10                      & 69.11                     & 71.62        & 71.73       & 46.80           & 48.35           \\\midrule
                          & Gold       & 78.02      & 80.17      & 73.20                      & 75.13                     & 73.75        & 74.26       & 45.26           & 41.86           \\
                          & SimPrompt  & 78.93      & 78.86      & 52.40                      & 52.83                     & 54.78        & 53.86       & 30.98           & 32.92           \\
                          & {MetaPrompt} & 78.17	& 78.27	 &53.53	 &52.55 &	51.30 &	48.93	 &30.54	 &31.50 \\
\multirow{-4}{*}{SimCSE~\cite{gao2021simcse}}  & AttrPrompt & 81.41      & 81.04      & 60.66                      & 61.32                     & 59.22        & 59.15       & 30.31           & 31.43           \\\midrule
                          & Gold       & 80.66      & 82.60       & 76.13                      & 78.34                     & 79.63        & 80.04       & 61.02           & 63.77           \\
                          & SimPrompt  & 84.37      & 84.17      & 55.66                      & 55.28                     & 63.66        & 64.71       & 43.22           & 41.50           \\
                           & {MetaPrompt} & 84.51 &	84.26	 &57.96 &	56.58 &	65.36 &	65.91	 &44.30 &	42.97 \\
\multirow{-4}{*}{COCO-DR~\cite{cocodr}} & AttrPrompt & 86.74      & 86.43      & 60.47                      & 62.03                     & 66.73        & 66.71       & 45.44           & 43.76           \\ \midrule
                          & Gold       &  76.31   &  79.37   &  74.25    &  76.38   &  72.47   &  73.42   &  44.65   &  44.39   \\
                          & SimPrompt  &  77.08   &  77.80   &  53.39    &  53.62   &  53.95   &  54.20   &  32.96   &  33.98   \\
                          & {MetaPrompt} & 77.19 &	76.67 &	51.54 &	49.84 &	54.96 &	52.00	 &32.99 &	33.65\\
\multirow{-4}{*}{Average} & \cellcolor{green!20}AttrPrompt &  \cellcolor{green!20}79.83   &  \cellcolor{green!20}80.00   &  \cellcolor{green!20}59.35    &  \cellcolor{green!20}60.48   &  \cellcolor{green!20}60.65   &  \cellcolor{green!20}60.89   &  \cellcolor{green!20}34.97   &  \cellcolor{green!20}35.30  \\ \bottomrule
\end{tabular}}
\label{tbl:lp}
\end{table}



\section{Additional Experimental Results}

\subsection{Linear probing with generated datasets}
\label{sec:linear_prob}

In this section, we evaluate the generated dataset under linear probing setting. We employ four text embedding methods and present the results in Table~\ref{tbl:lp}.
We can see that \method consistently outperforms \baseline and surpasses the Gold in the NYT dataset. This further indicates the datasets generated with \method have higher quality than \baseline and this superiority is robust to the model choices.

\subsection{Different temperature parameters for ChatGPT}
\label{sec:temp}
Temperature ($t$) is one crucial hyperparameter of LLMs that controls the diversity of the generated text~\cite{hashimoto-etal-2019-unifying}, while the studied attributed prompts are also for diversifying the generated data. We are then curious about the effectiveness of the temperature and how it compares to the \method. We study different values of the temperature using the NYT dataset and present the results in Table~\ref{tab:temp}. From the results, we can see that compared with the temperature, \method brings more significant performance gain, demonstrating its superiority over temperature tuning.

\begin{table}[!h]
\centering
\caption{Study of the temperature.
}
\scalebox{0.75}{%
\begin{tabular}{lccccccc}
\toprule
\multirow{3}{*}{\bf Method} &  \multicolumn{2}{c}{$t=1.0$} &  \multicolumn{2}{c}{$t=1.5$} &  \multicolumn{2}{c}{$t=2.0$}   \\ \cmidrule(lr){2-3} \cmidrule(lr){4-5} \cmidrule(lr){6-7} 
& Acc. & F1 & Acc. & F1 & Acc. & F1  \\ \midrule
\baseline &   76.00 &	76.34 & 76.78 &	77.31 &76.55	& 77.42 \\
\rowcolor{green!15} {\method} & 81.30 & 	82.26  & 81.47 &	82.23 &	79.47	 & 79.86\\
 \bottomrule
\end{tabular}
}
\label{tab:temp}
\end{table}

\subsection{The performance with respect to long-tail classes}
\label{sec:longtail}

\begin{figure}[h]
  \centering
  \includegraphics[width=\linewidth]{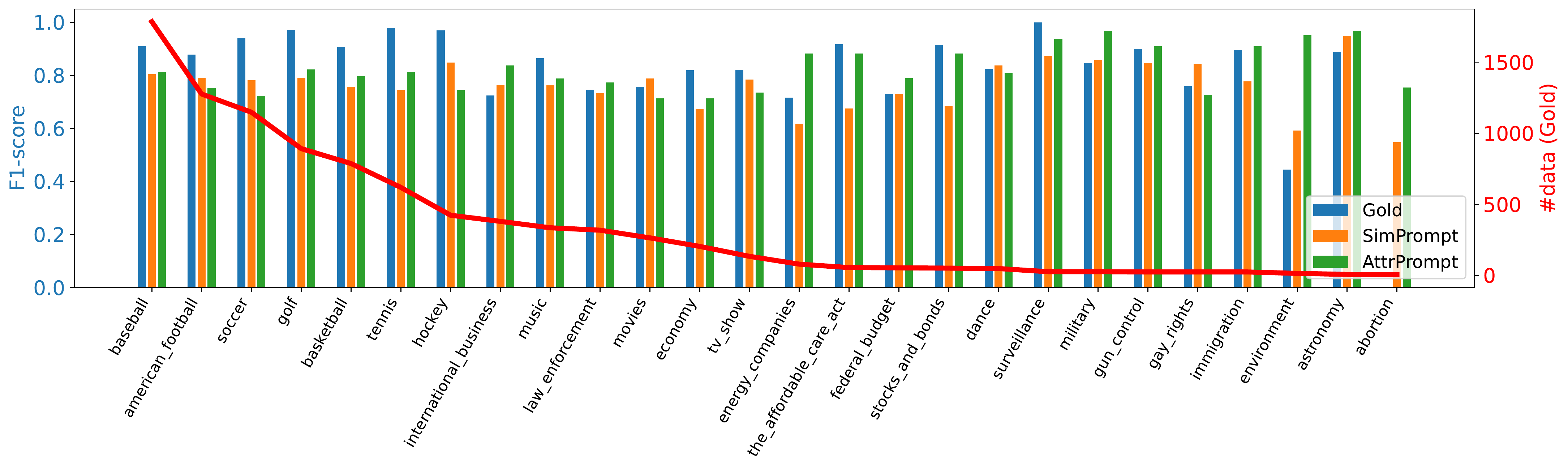}
    \vspace{-2ex}
  \caption{Per-class F1-score of the NYT dataset.}
  \vspace{-2ex}
  \label{fig:long-tail}
\end{figure}

As we have seen in Table~\ref{tab:dataset}, the original training sets of the involved datasets have severe long-tail classes issue since the imbalance ratio is high, yet the generated dataset are class-balanced, we are then curious how the class balance in the generated dataset benefits the model performance on long-tail classes.
We take the NYT dataset as an example and plot the per-class F1 score of Gold, \baseline, and \method in Figure~\ref{fig:long-tail}, where the x-axis is classes sorted by their number of data in the Gold dataset in descending order.
From the figure, we can see that out of 26 classes, \method renders the best per-class F1 score on 10 classes, which is 13 for Gold and 3 for \baseline.
Notably, for classes with few examples in the Gold set (the rightmost 4 classes in the figure), \method is better than the Gold and \baseline, especially for the class \mquote{abortion} with the fewest examples.
This suggests a data-centric way to handle the long-tail class issue in topic classification: one may use LLMs to generate class-balanced training set or augment the existing training set with the LLM-generated data such that the augmented dataset is class-balanced, the in-depth study of which is left as future work.

\subsection{Experiment results on other datasets}
\label{sec:other_datasets}
To provide a comprehensive evaluation of {\method}, we conducted additional empirical studies using datasets commonly employed in prior works, which typically exhibit \emph{low cardinality}. Consistent with \cite{ye-etal-2022-progen,ye-etal-2022-zerogen,yu2023zero}, we focused on four specific tasks: SST-2~\cite{socher2013recursive}, 
IMDB~\cite{maas2011learning}, 
Yelp~\cite{zhang2015character}, and
AG News~\cite{zhang2015character}. 
The IMDB, SST-2, and Yelp datasets involve binary sentiment classification, while the AG News dataset entails news classification with only four classes. This selection allows for diverse evaluation across different classification scenarios. 

To ensure a fair comparison, we utilized DistillBERT~\cite{sanh2019distilbert} as the classifier, in line with \cite{ye-etal-2022-progen,ye-etal-2022-zerogen,gao2023selfguided}. For the purpose of fine-tuning the classifier, we generated a total of $6000$ examples, evenly distributed among the classes.
For fine-tuning, we use a learning rate of $2\times 10^{-5}$ with 5 epochs and use the model after the final epoch for evaluation.

The results of our evaluation are presented in table~\ref{tab:zerogen}. Overall, we observe that \method demonstrates competitive performance across all four tasks. It outperforms our direct baseline (\baseline) by an impressive margin of 3.8\%. Moreover, when compared to other baselines incorporating dedicated noisy-aware learning techniques~\cite{wang2019symmetric,muller2019does}, \method still achieves promising performance, despite using the standard cross-entropy loss. This suggests that in addition to leveraging noisy-robust learning objectives, enhancing the diversity of input prompts serves as an effective alternative to improving downstream performance. It is interesting yet important to combine these different learning paradigms together to further boost the performance.

\begin{table}[t]
\caption{
Results with recent baselines using DistilBERT~\cite{sanh2019distilbert}. *: Results are copied from the previous papers~\cite{ye-etal-2022-progen,ye-etal-2022-zerogen,gao2023selfguided,yu2023zero}. 
}
\centering
\resizebox{0.72\columnwidth}{!}{
\begin{tabular}{l*{5}{c}}
\toprule
\bf Method/Dataset & \bf IMDB & \bf SST-2 &\bf Yelp & \bf AG News & \bf Avg. \\ 
\midrule \multicolumn{5}{l}{\emph{Simple fine-tuning with standard cross-entropy loss}}  \\
\midrule
\baseline &85.64	&79.38	&88.39&	74.30 &81.93 \\
\rowcolor{green!15} \method &85.92	&83.64 &	90.60 &	82.74 &85.73\\
\midrule \multicolumn{6}{l}{\emph{Other results with dedicated training techniques (for reference only)}}\\
\midrule
{Zero-shot Prompting}* & 77.31 & 82.63  & 78.10 & 73.20 & 77.81 \\
 ZeroGen* & 82.77&	80.41	&87.84	&76.48	&81.88  \\
 SuperGen* &85.05 &	84.28&	89.88	&77.40&	84.15 \\
ProGen* & 87.20	&84.12&	89.39&	80.81	&85.38 \\
SunGen* & 89.45	&89.38&	89.19&	80.49&	87.12 \\ 
ReGen* & 85.32 &	87.84	 &89.00 &	80.18 &	85.58 \\
\bottomrule
\end{tabular}
}
\vspace{-1ex}
\label{tab:zerogen}
\end{table}

\section{Additional Bias Analysis}

\subsection{Regional Bias with Manual Annotations}
\label{sec:human}

In the main body of the paper, we manually annotate location for 100 samples from each of the Gold, \baseline, and \method dataset.
Here, we compare manual annotations with the prediction of the attribute classifier in Table~\ref{tbl:human-loc} (the first three rows). 
Note that we include \mquote{unkown} as an option in manual annotation to absorb text without clear location specifications.
As per the results, there is a significant alignment between the manual annotations and the attribute classifier predictions. The accuracy is greater than 0.95 for the Gold and \method datasets, while it is 0.82 for \baseline. The lower accuracy in \baseline is attributed to its tendency to generate more data with unknown locations.

Furthermore, we use the manually annotated data to evaluate ChatGPT's adherence to the location attribute stipulated in the prompts. By comparing the location attribute in the prompt associated with each \method sample against the human annotations, we calculate an accuracy of 0.96 (as shown in the last row of Table~\ref{tbl:human-loc}). This high accuracy means that ChatGPT can effectively generate news adhering to the specified location attribute in the prompts.

\begin{table}[h]
\centering
\renewcommand\arraystretch{0.9}
\caption{Accuracies based on manual annotation of the location attribute of the NYT dataset.}
\begin{tabular}{@{}lcc@{}}
\toprule
\textbf{Comparison}                           & \textbf{Data Source}     & \textbf{Accuracy }\\ \midrule
\multirow{3}{*}{Attribute Classifier v.s. Human} & Gold       & 0.96     \\
                                     & SimPrompt  & 0.82     \\
                                     & AttrPrompt & 0.95     \\ \midrule
Attribute in Prompt v.s.   Human      & AttrPrompt & 0.96     \\ \bottomrule
\end{tabular}
\label{tbl:human-loc}
\end{table}

\subsection{Biases other than regional bias}
\label{app:bias}
\begin{figure}[!h]
	\centering
	\vspace{-2ex}
	\subfigure[  Amazon-Experience.]{
		\includegraphics[width=0.8\linewidth]{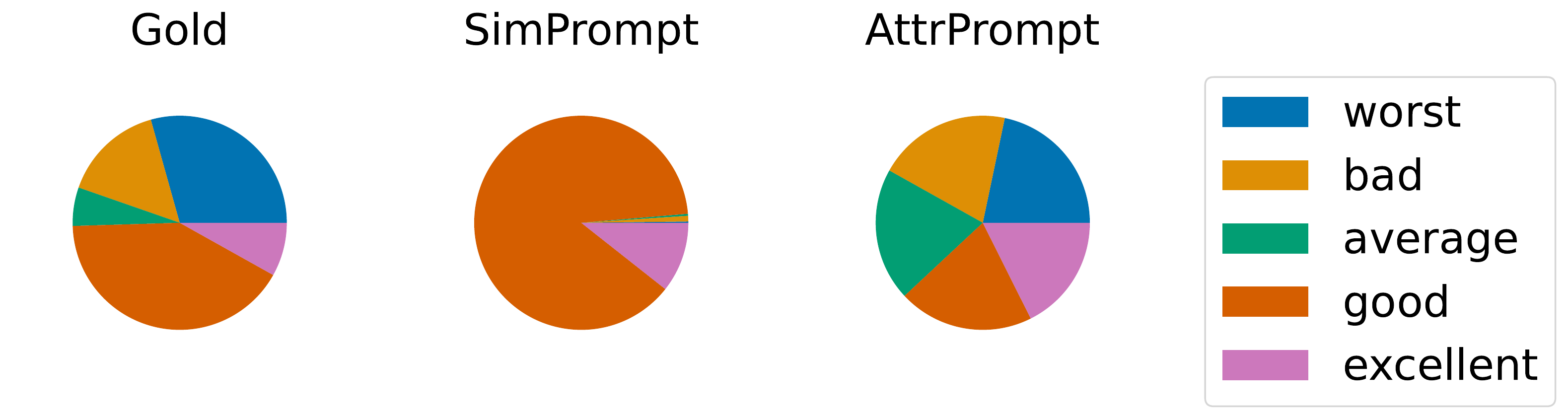}
		\label{fig:amazon-experience}
	} 

	\subfigure[ Reddit-Style.]{
		\includegraphics[width=1\linewidth]{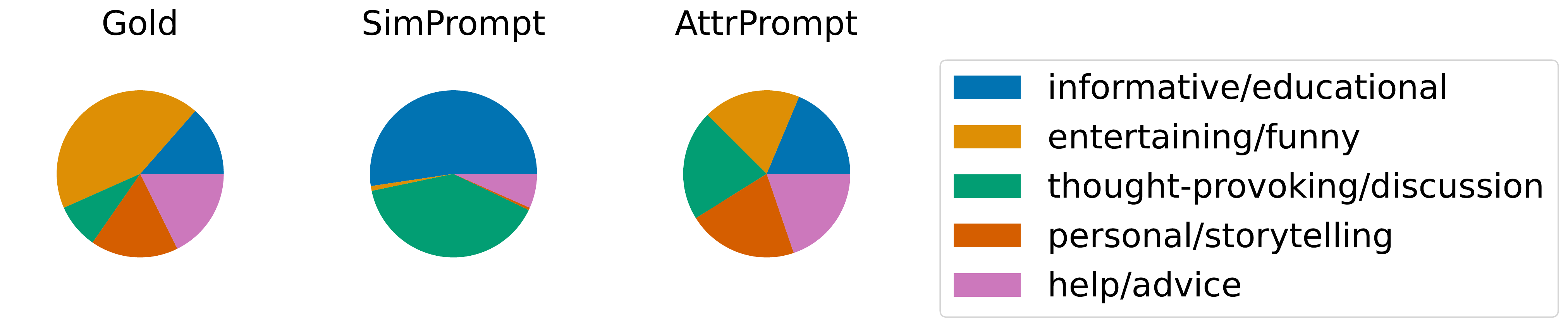}
		\label{fig:reddit-style}
	}  

	\subfigure[StackExchange-Style.]{
		\includegraphics[width=0.9\linewidth]{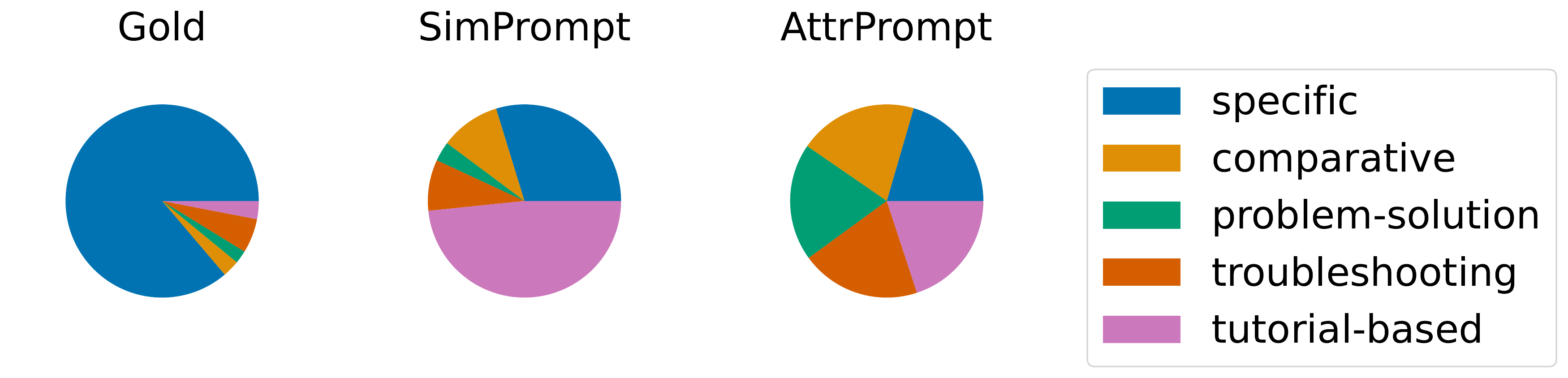}
		\label{fig:stackexchange-style}
	}  
	\caption{Pie charts of predicted attribute distributions on different datasets.}\label{fig:pie-otherattr}
\end{figure}

In addition to the regional bias we study in the main body of the paper, here we investigate bias in other datasets (Figure~\ref{fig:pie-otherattr}). From the results, we can see that the dataset generated with Gold and \baseline exhibit different level of biases, which shows that in general the biases exists in both generated and real dataset, while dataset generated with \method has relatively balanced attribute distribution.

\section{The Prompt Format for LLM-Generated Attributes}
\label{sec:prompt_format}
\subsection{Prompts for NYT Dataset}

\subsubsection{Key Attributes}
\textbf{Prompts to ChatGPT.} What do you think are important attributes to generate some diverse news from NYT under a specific topic, such as Football, Federal Budget. Examples: subtopics, writing style...

\textbf{Response from ChatGPT.} Important attributes of news from the New York Times under a specific topic could include:

Subtopics: exploring different angles and subtopics within the main topic can lead to a wider range of news stories.

Writing style: varying the writing style can help to make the news stories more interesting and engaging for readers. This can include using different tones, perspectives, and sentence structures.

Sources: using a variety of sources, such as experts, eyewitnesses, and affected individuals, can provide different perspectives and add diversity to the news stories.

Geographical locations: covering the topic from different geographical locations can help to show how it affects different communities and provide a global perspective.

Timeliness: covering current events and breaking news related to the topic can add diversity and relevance to the news stories.

Multimedia: using a range of multimedia elements, such as images, videos, and infographics, can help to add variety and appeal to the news stories.

Length: varying the length of the news stories, from short and concise to longer and more in-depth, can help to provide a wider range of information and cater to different readers' preferences.

\paragraph{Selected Attributes}
\begin{enumerate}
    \item Subtopics
    \item Writing Styles
    \item Geographical locations (determined by six continents by default)
    \item Length (determined by \hl{30-80} and \hl{100-150} by default)
\end{enumerate}

\subsubsection{Attributes Details}
\paragraph{Prompts to ChatGPT for Similar Classes.}  List 5 similar classes for \hl{class-name} news on NYT. The set of classes is listed as: \hl{[a list of class-names]}.

\paragraph{Prompts to ChatGPT for Subtopics.}  List 10 diverse subtopics for \hl{class\_name} news on NYT.
\paragraph{Response from ChatGPT.} See section \ref{sec:details} for details. 

\subsubsection{Writing Styles}
\paragraph{Prompts to ChatGPT.} List 5 different writing styles of NYT news.

\paragraph{Response from ChatGPT.}
\begin{enumerate}
\item Investigative reporting
\item Opinion/editorial
\item Human interest/profiles
\item Long-form narrative
\item Breaking news
\end{enumerate}

\subsection{Prompts for Amazon Dataset}
\subsubsection{Key Attributes}
\paragraph{Prompts to ChatGPT.} What do you think are important attributes to generate some diverse product reviews. Examples: writing style...

\paragraph{Response from ChatGPT.} When writing a review for a product, here are five important attributes to consider:

Product Type: Clearly mention the type of product you are reviewing, such as a smartphone, laptop, or fitness tracker. This helps readers understand the category and purpose of the product.

Brand: Specify the brand of the product as it often influences quality, reputation, and customer support. Discuss the brand's overall credibility and whether it aligns with your expectations.

User Experience: Evaluate the overall user experience of the product. Discuss its speed, accuracy, reliability, and efficiency in performing its intended tasks. Highlight any exceptional or lacking performance aspects.

Quality and Durability: Assess the quality of the product, including the materials used, construction, and overall durability. Discuss whether it feels well-made, solid, and likely to withstand regular use over time.

Features and Functionality: Describe the specific features and functions of the product. Highlight any unique or standout features that enhance its usability or set it apart from similar products in the market.

\subsubsection{Selected Attributes}
\begin{enumerate}
    \item Product Name
    \item Brands
    \item User Experience
    \item Writing Styles
    \item Length
\end{enumerate}

\subsubsection{Attributes Details}
\paragraph{Prompts to ChatGPT for Similar Classes.}  List 5 similar classes for \hl{product-class} products on Amazon. The set of classes is listed as: \hl{[a list of class-names]}.

\paragraph{Prompts to ChatGPT for Product Name/Brands.}  List 20 diverse product names/brands for \hl{product-class} on Amazon.
\paragraph{Response from ChatGPT.} See section \ref{app:amazon-attr} for details.

\subsubsection{Usage Experience}
\paragraph{Prompts to ChatGPT.} List 5 different usage experiences of Amazon reviews.

\paragraph{Response from ChatGPT.}
\begin{enumerate}
\item Worst: A highly unsatisfactory or problematic experience. Users may encounter multiple issues, such as usability problems, errors, and frustrations.
\item Bad: A below-average experience. Users may encounter some difficulties or frustrations while interacting with the product or service, but the overall experience falls short of meeting their expectations.
\item Average: A neutral or middling experience. Users may not encounter major issues, but the experience lacks any exceptional aspects or standout features. 
\item Good: A positive and satisfactory experience. Users have a smooth and enjoyable experience with minimal issues or obstacles.
\item Excellent: The highest level of satisfaction. Users have a seamless, delightful, and memorable experience.
\end{enumerate}

\subsubsection{Writing Styles}
\paragraph{Prompts to ChatGPT.} List 5 different writing styles of Amazon reviews.

\paragraph{Response from ChatGPT.}
\begin{enumerate}
\item Detailed Review: A detailed review provides a comprehensive analysis of the product, covering various aspects such as features, performance, durability, and user experience. 
\item Comparative Review: A comparative review compares the product with similar options available on Amazon.
\item Pros and Cons Review: A pros and cons review presents a balanced assessment of the product by listing its positive and negative aspects. 
\item Personal Experience Review: A personal experience review shares the reviewer's firsthand experience with the product. 
\item Recommendation Review: A recommendation review offers a clear recommendation or endorsement of the product.
\end{enumerate}

\subsubsection{Prompt Template}
See table \ref{tbl:amazon_prompt} for details.
\begin{table}[!h]
 \caption{Prompt template for the Amazon Product Review dataset.}
 \centering
\scalebox{0.7}{
\begin{tabular}{c|l}
\toprule 
{\textbf{Method}} &
  \multicolumn{1}{c}{\textbf{Prompt}} \\  \midrule
\baseline & {Suppose you are a review writer. Please write a review for \hl{product-class} in Amazon. } \\\midrule
\multirow{5}{*}{\method} & {Suppose you are a review writer. Please write a review for \hl{product-class} product in Amazon following the requirements below:} \\
&1. The review should be about the product of \hl{subtopic};\\ 
&2. The brand for the  product should be  \hl{brand};\\ 
&3. Should be in length between \hl{length:min-words} and \hl{length:max-words} words;\\
&4. Should describe the usage experience \hl{usage-experience} \\
& 5. The writing style of the review should be \hl{style};\\
& 6. the review must be relevant to  \hl{product-class} and irrelevant to: \hl{similar-class}.
  \\\bottomrule
\end{tabular}
}
\vspace{-2ex}
\label{tbl:amazon_prompt}
\end{table}
\subsection{Prompts for Reddit Dataset}

\subsubsection{Key Attributes}
\paragraph{Prompts to ChatGPT.} List 5 important attributes to generate some diverse reddit posts under a specific topic, like build\_a\_pc, premed, etc. Examples: writing style...

\paragraph{Response from ChatGPT.} 

Informative Guides and Tutorials: Create posts that provide step-by-step guides, tutorials, or informative content. These can include detailed instructions, tips, or recommendations to help readers in their journey. For example, in "build\_a\_pc," you could create a guide on assembling a gaming PC or optimizing system performance.

Writing Style: Utilize different writing styles to create diverse posts. For instance, you can write informative guides or tutorials, opinion pieces, comparison reviews, troubleshooting tips, or personal stories related to the topic.

Discussion Starters: Pose thought-provoking questions or statements to initiate discussions and engage the Reddit community. These can be open-ended queries, asking for opinions or experiences, or presenting a controversial viewpoint to encourage debate and dialogue.

Personal Experiences: Share personal stories or experiences related to the topic. This allows you to connect with the community on a more personal level and provide insights based on your firsthand experiences. It can include successes, challenges, lessons learned, or interesting anecdotes.

Resources and Recommendations: Provide valuable resources such as articles, websites, tools, or recommended products/services related to the topic. These resources can help the community access additional information, useful tools, or trusted recommendations, enhancing their understanding and engagement with the subject.

\subsubsection{Selected Attributes}

\begin{enumerate}
    \item Personal Experiences
    \item Resources
    \item Writing Styles
    \item Length
\end{enumerate}

\subsubsection{Attributes Details}
\paragraph{Prompts to ChatGPT for Similar Classes.}  List 5 similar classes for \hl{post-class} products on Reddit. The set of classes is listed as: \hl{[a list of class-names]}.

\paragraph{Prompts to ChatGPT for Personal Experiences/Resources.}  List 20 diverse Personal Experiences/Resources for posts about \hl{post-class} in Reddit.
\paragraph{Response from ChatGPT.} See section \ref{app:reddit-attr} for details.

\subsubsection{Writing Styles}
\paragraph{Prompts to ChatGPT.} List 5 different writing styles of Reddit Posts.

\paragraph{Response from ChatGPT.}
\begin{enumerate}
\item Informative/Educational: Create posts that aim to provide valuable information or educate the community on a particular topic. Share informative articles, research findings, guides, tutorials, or explain complex concepts in a simplified manner.
\item Entertaining/Funny: Craft posts that are meant to entertain or bring humor to the community. Share funny stories, jokes, memes, amusing videos, or any form of content that can elicit laughter and enjoyment.
\item Thought-Provoking/Discussion: Initiate posts that spark thoughtful discussions and invite users to share their opinions or experiences. Pose open-ended questions, present philosophical dilemmas, discuss current events, or explore societal issues to foster engaging conversations.
\item Personal/Storytelling: Share personal experiences, anecdotes, or narratives that others can relate to. Tell compelling stories, share achievements, challenges, or life lessons that can inspire or resonate with the community.
\item Help/Advice: Create posts where you seek help or offer advice to the community. Ask for guidance on a specific problem, seek recommendations, provide support, or share expertise in a particular field.
\end{enumerate}

\subsubsection{Prompt Template}
See table \ref{tbl:reddit_prompt} for details.
\begin{table}[!t]
 \caption{Prompt template for the Reddit dataset.}
 \centering
\scalebox{0.7}{
\begin{tabular}{c|l}
\toprule 
{\textbf{Method}} &
  \multicolumn{1}{c}{\textbf{Prompt}} \\  \midrule
\baseline & {Suppose you are a writer for posts in Reddit. Please give a post for \hl{post-class} in Reddit. } \\\midrule
\multirow{5}{*}{\method} & {Suppose you are a review writer. Please give a post in Reddit on \hl{post-class} community following the requirements below:} \\
&1. The post should focus on \hl{experience};\\ 
&2. Should be in length between \hl{length:min-words} and \hl{length:max-words} words;\\
& 3. The writing style of the post should be \hl{style};\\
& 4. Should mention the resource of \hl{resource};\\ 
& 5. the post must be relevant to  \hl{post-class} and irrelevant to: \hl{similar-class}.
  \\\bottomrule
\end{tabular}
}
\vspace{-2ex}
\label{tbl:reddit_prompt}
\end{table}

\section{Prompts for StackExchange Dataset}
\subsection{Selected Attributes}
\begin{enumerate}
    \item Depth of Knowledge
    \item Writing Styles
    \item Problem Scenarios
    \item Length
\end{enumerate}

\subsection{Key Attributes}
\textbf{Prompts to ChatGPT.} What do you think are important attributes to generate diverse posts under a specific topic on the stackexchange forum, like 3D meshes, server, etc.  For example, one important attribute could be  writing style...

\textbf{Response from ChatGPT.} When generating diverse posts under a specific topic on the Stack Exchange forum, several important attributes can contribute to the variety and richness of the discussions:

Depth of Knowledge: Participants with different levels of expertise can offer insights and perspectives from beginner to advanced, allowing for a broader range of questions and answers.

Diverse Perspectives: Encouraging participants from various backgrounds, industries, and experiences can bring fresh viewpoints and approaches to the topic.

Objectives: Encouraging participants to state their objectives or goals in their posts can help generate diverse content. Some posts may aim to seek advice, share experiences, provide step-by-step guides, analyze specific problems, or discuss broader concepts.

Problem Scenarios: Encourage users to discuss real-world problem scenarios related to the topic. This helps generate diverse posts by exploring different use cases, challenges, and approaches to problem-solving.

Problem Solving: Promoting problem-solving skills and encouraging users to share their challenges, roadblocks, and solutions can foster a collaborative and dynamic environment.

Clear Communication: Emphasizing clear and concise communication helps to ensure that questions are well-articulated and that answers are easily understood by a wide range of users.

\subsection{Prompts for arXiv Dataset}

\subsubsection{Key Attributes}
\textbf{Prompts to ChatGPT.} What do you think are important attributes to generate diverse papers under a specific topic on arXiv.  For example, one important attribute could be  writing style...

\textbf{Response from ChatGPT.} Generating diverse papers under a specific topic on platforms like arXiv involves considering various important attributes. Here are a few attributes that can contribute to generating diverse papers:

Writing Style: Encouraging papers with different writing styles, such as technical, expository, theoretical, or empirical, can bring diversity to the presentation and appeal to a wider range of readers.

Subtopics: Promoting papers that explore different subtopics within the broader topic can provide comprehensive coverage and delve into specific areas of interest.

Techniques: Encouraging papers that employ different research methodologies, such as experimental, computational, or analytical, can bring diverse approaches to studying the topic.

Data Sources: Promoting papers that utilize diverse data sources, such as surveys, simulations, real-world datasets, or case studies, can offer different perspectives and insights into the topic.

Interdisciplinary Perspectives: Encouraging papers that incorporate interdisciplinary perspectives, drawing insights from multiple fields or combining methodologies from different disciplines, can contribute to a richer understanding of the topic.

\subsubsection{Selected Attributes}
\begin{enumerate}
    \item Subtopics
    \item Techniques
    \item Writing Styles
    \item Length
\end{enumerate}

\paragraph{Prompts to ChatGPT for Subtopics/Techniques.}  List 20 diverse subtopics/techniques for paper about \hl{paper-class} in arXiv.
\paragraph{Response from ChatGPT.} See section \ref{app:arxiv-attr} for details.

\subsubsection{Writing Styles}
\paragraph{Prompts to ChatGPT.} List 5 different writing styles of arXiv papers.

\paragraph{Response from ChatGPT.}
\begin{enumerate}
\item Experimental: This style focuses on presenting empirical research with a strong emphasis on experimental design, data collection, and statistical analysis.
\item Theoretical: This style emphasizes the development of new theoretical frameworks or conceptual models. 
\item Review: This style involves synthesizing and analyzing existing literature on a specific topic.
\item Case Study: This style presents an in-depth examination of a specific case or situation, aiming to provide insights into complex phenomena or phenomena that are context-specific. 
\end{enumerate}

\subsubsection{Prompt Template}
See table \ref{tbl:stackexchange} for details.
\begin{table}[!t]
 \caption{Prompt template for the arXiv multi-label dataset.}
 \centering
\scalebox{0.75}{
\begin{tabular}{c|l}
\toprule 
{\textbf{Method}} &
  \multicolumn{1}{c}{\textbf{Prompt}} \\  \midrule
\baseline & {Please give a synthetic paper abstract about \hl{paper-classes}. } \\\midrule
\multirow{5}{*}{\method} & { Please write an abstract of a \hl{paper-classes} paper in arXiv following the requirements below:} \\
&1. Should focus on the  \hl{subtopic};\\ 
&2. Should be in length between \hl{length:min-words} and \hl{length:max-words} words;\\
& 3. The paper should use the technique relevant to \hl{technique};\\
& 4. The style of the paper should be in \hl{style};\\ \bottomrule
\end{tabular}
}
\vspace{-2ex}
\label{tbl:stackexchange}
\end{table}

\subsection{Prompts for AG News Dataset}
The prompts are the same as the NYT dataset. The only difference is the subtopics for different labels. The subtopic information is listed in \ref{app:agnews-attr} for details.
\subsection{Prompts for Yelp/SST-2/IMDB Dataset}
\subsubsection{Key Attributes}
\paragraph{Prompts to ChatGPT.}
What do you think are important attributes to generate some diverse reviews with sentiments? E.g.  writing styles.

\paragraph{Response from ChatGPT.}
Writing Styles: Utilizing different writing styles such as descriptive, narrative, persuasive, or humorous can create a diverse range of tones and emotions in the reviews.

Language and Tone: Varying the language and tone used in the reviews can evoke different sentiments. This can include formal or casual language, positive or negative tone, or even incorporating elements of storytelling.

Personal Experience: Sharing personal experiences and anecdotes in the reviews can add authenticity and emotional depth, allowing readers to connect with the sentiments expressed.

Descriptive Details: Including specific and vivid descriptions of the ambiance, decor, flavors, smells, or textures of the food can create a sensory experience that evokes various sentiments.

\subsubsection{Selected Attributes}

\begin{enumerate}
    \item Personal Experience
    \item Descriptive Details
    \item Writing Styles
    \item Length
\end{enumerate}

\subsubsection{Personal Experiences/Descriptive Details}
\paragraph{Prompts to ChatGPT.} Details are listed in Appendix \ref{app:agnews-attr} and \ref{app:yelp-attr}.

\subsubsection{Writing Styles}
\paragraph{Prompts to ChatGPT.} List 5 different writing styles of Restaurant Reviews (for Yelp).

\paragraph{Response from ChatGPT.}
\begin{enumerate}
\item Descriptive Style: Focuses on providing detailed descriptions of the restaurant's ambiance, decor, and overall atmosphere. It aims to paint a vivid picture of the dining experience.
\item Personal Narrative Style: Incorporates personal anecdotes and experiences from the visit to the restaurant. It adds a personal touch to the review and allows the reader to connect with the reviewer's perspective.
\item Analysis Style: Takes a more analytical approach by evaluating various aspects of the restaurant, such as food quality, service, cleanliness, and value for money.
\item Comparative Style: Compares the restaurant to similar establishments in terms of cuisine, price range, service quality, or other relevant factors. 
\item Gastronomic Style: Focuses primarily on the food and beverage offerings of the restaurant. 
\end{enumerate}

\paragraph{Prompts to ChatGPT.} List 5 different writing styles of Movie Reviews (for IMDB/SST-2).

\paragraph{Response from ChatGPT.}
\begin{enumerate}
\item Descriptive: Focuses on providing a detailed and vivid description of the movie and its elements.
\item Analytical: Analyzes and evaluates various aspects of the movie, such as the plot, acting, cinematography, and direction.
\item Persuasive: Aims to convince the reader to watch or avoid the movie by presenting strong arguments and opinions.
\item Comparative: Compares the movie to other films within the same genre or by the same director. 
\end{enumerate}

\subsubsection{Prompt Template}
See table \ref{tbl:yelp}  and \ref{tbl:sst2} for details.
\begin{table}[!t]
 \caption{Prompt template for the Yelp dataset.}
 \centering
\scalebox{0.75}{
\begin{tabular}{c|l}
\toprule 
{\textbf{Method}} &
  \multicolumn{1}{c}{\textbf{Prompt}} \\  \midrule
\baseline & {Suppose you are a review writer. Please give an example of a \hl{sentiment} review in Yelp. } \\\midrule
\multirow{5}{*}{\method} & { Please write a \hl{sentiment} review for a restaurant, following the requirements below:} \\
& 1. The overall review should be \hl{sentiment};\\ 
& 2. It should be a \hl{descriptive-details} restaurant ;\\
& 3. Should focus on \hl{personal-experience}; \\
& 4. Should be in length between \hl{length:min-words} and \hl{length:max-words} words;\\
& 5. The writing style of the review should be \hl{style};
  \\\bottomrule
\end{tabular}
}
\label{tbl:yelp}
\end{table}

\begin{table}[!t]
 \caption{Prompt template for the SST-2/IMDB dataset.}
 \centering
\scalebox{0.75}{
\begin{tabular}{c|l}
\toprule 
{\textbf{Method}} &
  \multicolumn{1}{c}{\textbf{Prompt}} \\  \midrule
\baseline & {Suppose you are a review writer. Please give an example of a \hl{sentiment} review for a movie. } \\\midrule
\multirow{5}{*}{\method} & { Please write a \hl{sentiment} review for a movie, following the requirements below:} \\
& 1. The overall review should be \hl{sentiment};\\ 
& 2. It should be a \hl{descriptive-details} movie ;\\
& 3. Should focus on \hl{personal-experience}; \\
& 4. Should be in length between \hl{length:min-words} and \hl{length:max-words} words;\\
& 5. The writing style of the review should be \hl{style};
  \\\bottomrule
\end{tabular}
}
\vspace{-2ex}
\label{tbl:sst2}
\end{table}

\section{Details for LLM-Generated Attributes}
\label{sec:details}
In this section, we present a brief description of LLM-Generated Attributes. For a comprehensive list of attributes, please refer to the GitHub repository:  \url{https://github.com/yueyu1030/attrprompt}.

\subsection{NYT Attributes}\label{app:nyt-attr}
\subsubsection{Subtopics}\label{subsubsec:nyt-subtopics}

We randomly select 3 categories in NYT dataset and display the corresponding subtopic attributes for each category:
\begin{itemize}
    \item astronomy: \begin{itemize}
        \item Discoveries of exoplanets
\item Black holes and their role in shaping galaxies
\item The search for extraterrestrial life
\item Gravitational waves and the study of the universe's origins
\item The use of telescopes to explore the universe
\item The mysteries of dark matter and dark energy
\item Solar flares and their impact on Earth
\item The history of the universe and its evolution over time 
\item Exploring the possibility of space tourism
\item The exploration of our neighboring planets, such as Mars and Venus.
    \end{itemize}
    \item baseball: \begin{itemize}
        \item Recent controversy surrounding sign-stealing scandal in MLB
\item Breakdown of top prospects in minor league baseball 
\item Analysis of new rule changes for upcoming baseball season
\item Coverage of recent World Series champions and their success 
\item In-depth profile of influential baseball figures, such as managers or players 
\item Updates on retired players and their post-baseball careers 
\item Highlighting standout performances by individual players or teams in recent games
\item Coverage of international baseball leagues and their top players
    \end{itemize}
    \item economy: \begin{itemize}
        \item Job market and employment rates
\item  Interest rates and monetary policy
\item  Inflation and deflation
\item  Economic growth and GDP
\item Consumer spending and retail sales
\item Income inequality and poverty
\item GDP growth and contraction
\item Labor market trends
\item Economic impacts of natural disasters and pandemics
\item  Housing market and real estate
    \end{itemize}
\end{itemize}

\subsection{Amazon Attributes}\label{app:amazon-attr}

\subsubsection{Product Brands}\label{subsubsec:amazon-brand}
We randomly select 3 categories in Amazon dataset and display the corresponding product brand attributes for each category:
\begin{itemize}
\item camera\_photo.: \begin{itemize}
\item Canon
\item Nikon
\item Leica
\item Hasselblad
\item Fujifilm
\item Lumix (Panasonic)
\item GoPro
\item Polaroid
\item Yashica
\item Mamiya
\end{itemize}
\item health\_and\_personal\_care.: \begin{itemize}
\item Johnson \& Johnson
\item Dove 
\item Colgate 
\item Head \& Shoulders 
\item Merck 
\item Pfizer
\item Merck
\item Abbott Laboratories
\item GlaxoSmithKline
\end{itemize}
\item toys\_games.: \begin{itemize}
\item Mattel
\item Fisher-Price
\item Hasbro
\item Lego
\item Nerf
\item Barbie
\item Hot Wheels
\item Playmobil
\item MGA Entertainment
\item Paw Patrol
\end{itemize}
\end{itemize}

\subsubsection{Product Names}\label{subsubsec:amazon-name}
We randomly select 3 categories in Amazon dataset and display the corresponding product name attributes for each category:
\begin{itemize}
\item camera\_photo.: \begin{itemize}
\item SnapShooter
\item FocusReady
\item ClickCapture
\item MemoriesMatter
\item FlashFinder
\item PicturePerfect
\item ShotSnap
\item VisionVibe
\item PixelPal
\item FreezeFrame
\item AngleAttack
\item SnapStash
\item FrameFlair
\item SmileSnaps
\item ImageImpact
\item ZoomZest
\item ClickCraze
\item PixelPassion
\item ShootSmart
\item CaptionChamp.
\end{itemize}
\item baby.: \begin{itemize}
\item Baby Swing
\item Diaper Genie
\item Milk Warmer
\item Baby Carrier
\item Car Seat
\item Baby Monitor
\item Baby Food Maker
\item Nursery Glider
\item Changing Table
\item Baby Bouncer
\item Playpen
\item Teething Rings
\item Baby Wipes Dispenser
\item Baby Bath Tub
\item Bibs
\item Baby Blankets
\item Pacifier Clip
\item Baby Sling
\item Baby Napper
\item Moses Basket
\end{itemize}
\item outdoor\_living.: \begin{itemize}
\item Sunbrella
\item Weber
\item Keter
\item Trex
\item Solaira
\item Tropitone
\item Bimini Solutions
\item La-Z-Boy Outdoor
\item Suncast
\item Beltwood
\item Quikrete
\item Cosco Outdoor Living
\item Anova Furnishings
\item Duramax
\item US Polymers
\item Ostrich Chairs
\item Carefree of Colorado
\item Tuff Coat
\item Fire Sense
\item Heritage Patios.
\end{itemize}
\end{itemize}

\subsection{Reddit Attributes}\label{app:reddit-attr}

\subsubsection{Resources}\label{subsubsec:reddit-resource}
We randomly select 3 categories in Reddit dataset and display the corresponding resource attributes for each category:
\begin{itemize}
\item buddy\_crossing.: \begin{itemize}
\item Meetup.com - a website that allows you to find and join groups of individuals with similar interests in your local area, including hiking, book clubs, and social events.
\item The Buddy System: Understanding Mental Illness and Addiction - a book that explores the biology of addiction and provides a guide for friends and family members of individuals struggling with these issues.
\item Lynda.com - a subscription-based online learning platform that provides courses on a variety of subjects including computer programming, business, web design, and more.
\item Codecademy.com - an interactive online platform that teaches coding skills for free or through a subscription.
\item Khan Academy - a nonprofit organization that provides free online courses in a wide range of subjects including math, science, and humanities to learners of all ages.
\item Duolingo - a language-learning app that is available for free on the App Store and Google Play, offering courses in a variety of languages including Spanish, French, and German.
\item MindBody App - a mobile app that helps users find and book local fitness, wellness, and beauty activities such as yoga classes, massages, and haircuts.
\item Headspace - a meditation app that offers guided meditation courses to help users reduce stress and improve focus.
\item The Knot - a website that provides tools and resources for wedding planning, including a Wedding Website Builder, guest list tracker, and registry management.
\item Khan Academy - a nonprofit organization that provides free online courses in a wide range of subjects including math, science, and humanities to learners of all ages.
\item Others resource for buddy\_crossing.
\end{itemize}
\item whats\_that\_book.: \begin{itemize}
\item Goodreads - A social platform for book lovers where users can search for books, create bookshelves, and write reviews.
\item LibraryThing - A community-driven cataloging website where users can create and share their personal book collections.
\item AbeBooks - An online marketplace for rare and out-of-print books, as well as other antique or collectible items.
\item Shelfari - An online book club where users can share book recommendations and read reviews from others.
\item Project Gutenberg - A digital library of freely available public domain books.
\item Paperback Swap - A book trading community where users can exchange books with others across the US.
\item Goodreads Librarians Group - A community of Goodreads users who help with book cataloging, including identifying books from incomplete information.
\item Book Riot - A website featuring book reviews and book-related news, with an emphasis on diverse and underrepresented voices.
\item The New York Times Book Review - A renowned weekly publication featuring book reviews, author interviews, and literary criticism.
\item Others resource for whats\_that\_book.
\end{itemize}
\item pokemongo\_friends.: \begin{itemize}
\item Pokemon GO Hub: A comprehensive website dedicated to news, guides, and analysis on Pokemon GO.
\item The Silph Road Subreddit: A community-run subreddit dedicated to research and analysis of Pokemon GO mechanics.
\item Poke Assistant: A website that offers a range of tools to help you optimize your Pokemon GO experience, including IV calculators and gym battle simulations.
\item The Trainer Club: A YouTube channel that provides daily updates, news, and tips for Pokemon GO trainers.
\item Gotta Catch 'Em All: A Facebook group where you can connect with other Pokemon GO players and coordinate raid battles and other activities.
\item Reddit's r/PokemonGOFriends Subreddit: A community of players looking for friends to exchange gifts and share invites for raids.
\item The PokeMap: A website that allows you to find nearby Pokemon on a map in real-time.
\item Poke Genie: An app that automatically calculates IVs and other stats for your Pokemon, saving you time and headaches.
\item Pokemon GO Gamepress: A website that offers detailed breakdowns and analysis of Pokemon, movesets, and other game mechanics.
\item The Go Ranger App: An app that helps you plan your raids and battles, with intuitive mapper tools and filters to help you find the Pokemon you're looking for.
\item Others resource for pokemongo\_friends.
\end{itemize}
\end{itemize}

\subsubsection{Experience}\label{subsubsec:reddit-experience}
We randomly select 3 categories in Reddit dataset and display the corresponding experience attributes for each category:
\begin{itemize}
\item build\_a\_pc.: \begin{itemize}
\item DIY PC Builds: Sharing personal experiences and success stories of building custom PCs, discussing component choices, troubleshooting, and performance optimizations.
\item Budget-Friendly Builds: Discussing experiences with building PCs on a tight budget, sharing cost-saving tips, and recommendations for budget-friendly components.
\item Cable Management: Sharing personal experiences and tips for effective cable management in PC builds, discussing cable routing techniques and showcasing clean build aesthetics.
\item RGB Lighting: Discussing experiences with RGB lighting setups in PC builds, sharing recommendations for RGB components, software customization, and lighting effects.
\item Troubleshooting Builds: Sharing experiences and tips for troubleshooting common issues in PC builds, helping fellow builders diagnose and solve hardware or software problems.
\item Silent and Quiet PC Builds: Discussing experiences and recommendations for building silent or quiet PCs, focusing on noise reduction techniques and quiet component choices.
\item Workstation Builds: Sharing experiences and insights into building PCs for professional workloads, such as video editing, 3D rendering, programming, and graphic design.
\item Water-Cooling Adventures: Sharing experiences and insights into custom water-cooling loops, discussing the challenges, benefits, and performance improvements achieved.
\item Unique and Custom Builds: Showcasing and discussing unique and custom PC builds, including themed builds, custom cases, or exotic cooling solutions.
\item Build Planning and Component Selection: Discussing experiences with planning PC builds, researching and selecting components, considering compatibility, and balancing performance and budget.
\item Modding and Case Customization: Sharing experiences with PC case modding and customization, discussing techniques, materials, and showcasing personal projects.
\item Compact and Small Form Factor Builds: Discussing experiences with building compact or small form factor PCs, sharing recommendations for mini-ITX cases, cooling solutions, and component choices.
\item Home Server and NAS Builds: Sharing experiences and insights into building home servers and network-attached storage (NAS) systems, discussing storage options, software, and data management.
\item Multimonitor Setups: Discussing experiences with multimonitor setups, sharing tips for optimizing productivity and gaming experiences across multiple displays.
\item PC Gaming Peripherals: Sharing experiences and recommendations for gaming peripherals, such as keyboards, mice, monitors, and headsets, discussing features and personal preferences.
\end{itemize}
\item xbox.: \begin{itemize}
\item Xbox One exclusive games such as Halo 5, Forza Horizon 4, and Gears of War 4
\item Xbox One media and entertainment apps such as Netflix and Hulu
\item memorable gaming moments or achievements on the Xbox console.
\item Purchase Xbox One online.
\item Xbox Kinect motion sensor accessory 
\item Xbox Play Anywhere program 
\item Other Experience of Xbox
\end{itemize}
\item metal\_gear\_solid.: \begin{itemize}
\item Tactical Weapon Customization: Experimenting with various weapons, attachments, and equipment to tailor loadouts to different mission objectives and playstyles.
\item Character Development: Witnessing the growth and development of iconic characters such as Solid Snake, Big Boss, or Raiden throughout their respective story arcs.
\item Stealthy Takedowns: Executing silent and non-lethal takedowns, utilizing tranquilizer darts, chokeholds, or sneaking up on enemies from behind.
\item Tactical Planning: Strategizing mission approaches, analyzing enemy patrols, setting traps, and utilizing distractions to gain the upper hand.
\item Memorable Characters: Developing connections with unique and memorable characters, such as Otacon, Meryl, Revolver Ocelot, or The Boss.
\item Stealthy Gadgets: Utilizing gadgets and tools, such as the cardboard box, thermal goggles, or the iconic Solid Eye, to gain advantages during missions.
\item Emotional Storytelling: Experiencing emotionally impactful moments within the narrative, exploring themes of loss, betrayal, loyalty, and the human cost of warfare.
\item Espionage Tactics: Participating in undercover missions, gathering intelligence, infiltrating enemy bases, and sabotaging their operations.
\item Lore and Mythology: Delving into the intricate lore, conspiracies, and historical events within the Metal Gear Solid universe, including topics like The Patriots or the Philosopher's Legacy.
\item Groundbreaking Game Design: Appreciating the innovative gameplay mechanics, cinematic presentation, and attention to detail that have made the Metal Gear Solid series a beloved and influential franchise in the gaming industry.
\item Others experience for metal\_gear\_solid.
\end{itemize}
\end{itemize}

\subsection{StackExchange Attributes}\label{app:stackexchange-attr}

\subsubsection{Scenarios}\label{subsubsec:stackexchange-scenarios}
We randomly select 3 categories in StackExchange dataset and display the corresponding scenario attributes for each category:
\begin{itemize}
\item multiplayer.: \begin{itemize}
\item Cheating/hacking in online games
\item Inappropriate player behavior
\item Unbalanced game mechanics
\item Difficulty connecting to multiplayer servers
\item Matchmaking errors
\item Unresponsive or laggy gameplay
\item Glitches in gameplay affecting online matches
\item Difficulty finding players to match with
\item Balancing player skills in matchmaking
\item Disconnects and dropped connections mid-game
\item Cross-platform compatibility issues
\item In-game communication problems
\item Difficulty managing and moderating game servers
\item Addressing griefing and trolling in multiplayer games
\item Managing player accounts and login systems
\item Implementing or improving anti-cheat measures
\item Community feedback and feature requests
\item Addressing game-breaking exploits
\item Ensuring fair and accurate reporting of player statistics
\item Addressing server crashes and downtime.
\end{itemize}
\item procedural-generation.: \begin{itemize}
\item Improving the efficiency of procedural generation algorithms in Python.
\item Troubleshooting issues with memory usage in large-scale procedural generation projects.
\item Debugging issues with randomized content generation in procedural levels.
\item Implementing procedural generation techniques in C++ for game development.
\item Exploring the potential of machine learning in procedural content generation.
\item Optimizing the generation of complex 3D models using procedural algorithms.
\item Managing complexity and maintaining consistency in procedurally generated game worlds.
\item Addressing issues with procedural generation of text-based content, such as dialogue or item descriptions.
\item Developing tools to aid in the creation and testing of procedural generation algorithms.
\item Balancing the need for randomness with player expectations for fairness and balance.
\item Addressing issues with the procedural generation of music and sound effects.
\item Improving the visual quality of procedurally generated game assets.
\item Exploring ethical concerns around the use of AI in procedural content generation.
\item Developing procedural generation techniques for non-linear narratives or branching storylines.
\item Improving the procedural generation of inhabited environments, such as procedurally generated NPCs.
\item Addressing issues with the procedural generation of terrain features such as rivers and mountains.
\item Implementing procedural generation techniques for user-generated content.
\item Supporting multithreaded execution in procedural generation algorithms.
\item Ensuring procedural generation techniques are compatible with various game engines and frameworks.
\item Improving the scalability of procedural generation algorithms for use in multiplayer games.
\end{itemize}
\item networking.: \begin{itemize}
\item Difficulty in troubleshooting network connection issues on a Linux system
\item Configuring a wireless access point for a large office space
\item Implementing load balancing across multiple servers in a network
\item Optimizing network throughput to reduce latency in a gaming environment
\item Implementing firewall rules to block unauthorized access to a network
\item Troubleshooting DNS resolution issues on a Windows server
\item Designing and implementing a secure VPN connection
\item Setting up a network file server for shared access among multiple clients
\item Configuring SNMP to monitor network traffic and utilization
\item Designing a network topology for a large enterprise with multiple locations
\item Troubleshooting issues with Ethernet switches in a data center environment
\item Implementing QoS to prioritize network traffic for critical applications
\item Configuring NAT and PAT to enable internet access for multiple devices
\item Setting up and configuring VLANs to segment a network
\item Troubleshooting issues with network printers in an office environment
\item Configuring routing protocols in a large network
\item Securing wireless access points to prevent unauthorized access
\item Troubleshooting issues with VPN connection stability and speed
\item Implementing network virtualization with virtual LANs and virtual switches
\item Designing and implementing an effective network security strategy to prevent data breaches.
\end{itemize}
\end{itemize}

\subsection{ArXiv Attributes}\label{app:arxiv-attr}
\subsubsection{Subtopics}\label{subsubsec:arxiv-sub}
We randomly select 3 categories in arXiv dataset and display the corresponding product brand attributes for each category:
\begin{itemize}
\item machine\_learning.: \begin{itemize}
\item  Text generation
\item   Natural language understanding for chatbots
\item   Sentiment analysis and opinion mining
\item   Text summarization and keyword extraction
\item   Machine translation
\item   Named entity recognition and entity linking
\item   Dialogue systems and conversational agents
\item   Cross-lingual and Multilingual NLP
\item  Text-to-speech systems
\item   Phonetics and phonology in computational linguistics
\item   Grammatical error detection and correction
\item   Speech recognition and acoustic modeling
\item   Semantic role labeling
\item   Discourse analysis and coherence modeling
\item   Lexical semantics and word sense disambiguation
\item   Computational lexicography and machine-readable dictionaries
\item   Language Modeling
\item   question answering
\item   Language resources and corpora
\item   Computational sociolinguistics and dialectology.

\end{itemize}
\item number\_theory.: \begin{itemize}
\item  Prime numbers 
\item  Diophantine equations 
\item  Modular arithmetic 
\item  Cryptography 
\item  Continued Fractions 
\item  Pell’s Equation 
\item  Fermat’s Last Theorem 
\item  Algebraic Number Theory 
\item  Riemann Hypothesis 
\item  Arithmetic Geometry 
\item  Quadratic Forms 
\item  L-Functions 
\item  Automorphic Forms 
\item  Galois Theory 
\item  Ramsey Theory 
\item  Distribution of Prime Numbers 
\item  Number Theory in Cryptography 
\item  Summation Formulas 
\item  Gaussian Integers 
\item  The Goldbach Conjecture

\end{itemize}
\item geophysics.: \begin{itemize}
\item  Seismic imaging
\item  Earthquake prediction
\item Geothermal energy
\item  Volcanic eruptions
\item  Plate tectonics
\item  Geomagnetism
\item  Paleomagnetism
\item  Geophysical surveying
\item  Geophysical fluid dynamics
\item  Gravity measurements
\item  Rock physics
\item  Crustal deformation
\item Geomorphology
\item  Mineral exploration
\item  Earth structure modeling
\item  Geodetic techniques
\item  Hydrogeophysics
\item  Earth modeling
\item  Electrical geophysics
\item  Remote sensing geophysics
\end{itemize}
\end{itemize}

\subsubsection{Techniques}\label{subsubsec:arxiv-tech}
We randomly select 3 categories in the arXiv dataset and display the corresponding attributes for each category:
\begin{itemize}
\item genomics.: \begin{itemize}
\item Genome assembly and annotation using hybrid approaches.
\item  Comparative genomics for analyzing evolutionary relationships between genomes.
\item  Differential gene expression analysis using RNA sequencing data.
\item  Metagenomics for studying the microbial communities in different environments.
\item  Epigenetic analysis for understanding gene regulation.
\item  Network analysis for identifying gene interactions and pathways.
\item  Structural variation analysis for detecting genomic rearrangements.
\item  Functional genomics for studying gene function and pathway regulation.
\item  Genome-wide association studies for identifying genetic variants associated with complex traits.
\item  High-throughput screening methods for identifying genes involved in specific biological processes.

\end{itemize}
\item number\_theory: \begin{itemize}
\item  Primality testing using elliptic curves
\item  Continued fraction factorization method
\item Algorithm for solving Diophantine equations
\item  Quadratic sieve algorithm for integer factorization
\item  Pollard rho algorithm for integer factorization
\item  Digital sum subtraction method for computing discrete logarithm
\item  Fermat's method for factorization of primes
\item  Chinese remainder algorithm for solving modular equations 
\item  Exponential-sum algorithm for computing in algebraic number fields 
\item  Generalized Ramanujan-Selberg formula for counting integer points on algebraic varieties.

\end{itemize}
\item geophysics.: \begin{itemize}
\item  Seismic attribute interpretation
\item Full waveform inversion
\item Gravity inversion
\item Spherical geometries
\item Ground penetrating radar imaging
\item Time-lapse reservoir monitoring
\item Electrical resistivity tomography
\item Joint inversion of geophysical data
\item Radiometric dating
\item Geomagnetic field modeling
\end{itemize}
\end{itemize}

\subsection{AG News Attributes}\label{app:agnews-attr}
\subsubsection{Subtopics}\label{subsubsec:agnews-subtopics}

The corresponding subtopic attributes for each category are shown as follows:
\begin{itemize}
    \item business: \begin{itemize}
\item Corporate earnings and financial reports
\item Stock market updates and analysis
\item Mergers and acquisitions
\item Business regulations and policies
\item Startups and entrepreneurship
\item Industry trends and forecasts
\item Economic indicators and market trends
\item Business strategies and management practices
\item Corporate governance and ethics
\item Consumer behavior and market research
\item Business leadership and executive profiles
\item Banking and finance industry updates
\item Energy and sustainability in business
\item Retail and e-commerce trends
\item Real estate and property market updates
\item Business disruptions and crisis management
\item Corporate social responsibility and sustainability initiatives
    \end{itemize}
    \item sci\_tech: \begin{itemize}
\item Artificial intelligence
\item Robotics
\item Quantum computing
\item Biotechnology
\item Nanotechnology
\item Internet of Things
\item Renewable energy
\item Virtual reality
\item Augmented reality
\item Cybersecurity
\item Genetic engineering
\item Big data
\item Autonomous vehicles
\item 3D printing
\item Blockchain technology
\item Bioinformatics
\item Machine learning
\item Biomedical engineering
\item Clean technology
    \end{itemize}
    \item sports: \begin{itemize}
\item Soccer
\item Basketball
\item Baseball
\item Tennis
\item Golf
\item Cricket
\item Rugby
\item Athletics
\item Formula 1
\item Olympics
\item Boxing
\item Swimming
\item Volleyball
\item Ice hockey
\item American football
\item Cycling
\item Motorsports
\item Martial arts
\item Horse racing
\item Surfing
    \end{itemize}
    \item world: \begin{itemize}
\item International politics and diplomacy
\item Global conflicts and war
\item Terrorism and security threats
\item Human rights issues and social justice movements
\item Migration and refugee crises
\item Climate change and environmental policies
\item Global health crises and pandemics
\item Natural disasters and emergencies
\item Cross-border crime and corruption
\item Cultural and social developments worldwide
\item Geopolitical tensions and territorial disputes
\item International aid and development efforts
\item Humanitarian crises and relief efforts
\item Cultural heritage preservation and promotion
\item International collaborations and partnerships
    \end{itemize}
\end{itemize}
\subsection{SST-2 Attributes}\label{app:sst2-attr}
\subsubsection{Subtopics}\label{subsubsec:agnews-subtopics}

We display the corresponding subtopic attributes for each category as follows:
\begin{itemize}
    \item positive: \begin{itemize}
\item Compelling Storyline: A strong and engaging narrative that captures the audience's attention from beginning to end.
\item Well-Developed Characters: Memorable and relatable characters that evoke emotions and drive the story forward.
\item Skillful Direction: Effective direction that showcases the filmmaker's vision, ensuring cohesive storytelling and engaging visual elements.
\item Excellent Acting: Convincing performances from the cast that bring the characters to life and immerse the audience in the story.
\item Cinematography: Expertly captured visuals, including the use of framing, lighting, and camera movements, to enhance the storytelling and create a visually appealing experience.
\item Engaging Dialogue: Well-written dialogue that is natural, meaningful, and contributes to character development and plot progression.
\item Sound Design and Music: Thoughtful and immersive sound design, including sound effects and a well-curated soundtrack or original score, that enhances the overall cinematic experience.
\item Production Design: Attention to detail in creating visually appealing and authentic sets, costumes, and overall aesthetics that contribute to the film's atmosphere and world-building.
\item Editing: Skillful editing that maintains a good pace, effectively transitions between scenes, and enhances the overall flow and impact of the story.
\item Emotional Impact: A movie that evokes emotions, whether it be through humor, drama, suspense, or other means, leaving a lasting impression on the audience.
    \end{itemize}
    \item negative: \begin{itemize}
\item Weak Plot: A poorly developed or uninteresting storyline that fails to engage the audience.
\item Lackluster Performances: Unconvincing or uninspired performances by the actors that fail to bring the characters to life.
\item Poor Production Quality: Subpar production values, including low-quality visuals, amateurish cinematography, and weak special effects.
\item Incoherent Storytelling: Confusing or disjointed narrative structure that makes it difficult to follow or understand the plot.
\item Unmemorable Characters: Underdeveloped or forgettable characters that fail to resonate with the audience.
\item Weak Soundtrack: A forgettable or poorly composed soundtrack that fails to enhance the mood or add depth to the movie.
\item Poor Dialogue: Uninteresting or poorly written dialogues that fail to engage or resonate with the audience.
\item Disjointed Atmosphere: A lack of coherence or consistency in creating an immersive and believable world for the viewers.
\item Unresolved Plotlines: Loose ends or unresolved plotlines that leave the audience feeling unsatisfied or confused.
\item Lack of Entertainment Value: A movie that fails to deliver an enjoyable or engaging experience for the audience, leaving them feeling bored or uninterested.
    \end{itemize}
\end{itemize}

\subsubsection{Descriptive Details}
We use movie genres as the characteristics of movies, and the attributes are listed as follows:
\begin{itemize}
    \item Action
    \item Drama
    \item Comedy
    \item Thriller
    \item Romance
    \item Horror
    \item Adventure
    \item Science Fiction
    \item Fantasy
    \item Animation
\end{itemize}

\subsection{Yelp Attributes}\label{app:yelp-attr}
\subsubsection{Subtopics}\label{subsubsec:agnews-subtopics}

We display the corresponding subtopic attributes for each category as follows:
\begin{itemize}
    \item positive: \begin{itemize}
\item Quality of Food: The taste, flavor, and presentation of the dishes.
\item Fresh Ingredients: The use of fresh and high-quality ingredients in the preparation of the food.
\item Menu Variety: A diverse range of options catering to different dietary preferences and restrictions.
\item Presentation: The visually appealing presentation of the dishes.
\item Service: Attentive, friendly, and prompt service from the restaurant staff.
\item Value for Money: Offering good quality and portion sizes at reasonable prices.
\item Cleanliness: A clean and well-maintained dining area, including tables, utensils, and restrooms.
\item Special Dietary Accommodations: Catering to specific dietary needs such as vegetarian, vegan, gluten-free, etc.
\item Unique and Creative Dishes: Offering innovative and creative dishes that stand out.
\item Efficient Operations: Smooth and well-coordinated operations to minimize waiting times and delays.
    \end{itemize}
    \item negative: \begin{itemize}
\item Poor Service: Slow or inattentive service from the restaurant staff.
Unfriendly Staff: Rude or unhelpful behavior from the restaurant staff.
\item Long Waiting Times: Excessive waiting times for a table or food.
\item Incorrect Orders: Receiving incorrect or poorly prepared food orders.
\item Unappetizing Presentation: Dishes that are poorly presented or lack visual appeal.
\item Unpleasant Ambience: Uncomfortable or uninviting atmosphere in the restaurant.
\item Dirty or Unhygienic Conditions: Lack of cleanliness in the dining area, restrooms, or utensils.
\item Limited Menu Options: A limited selection of dishes or lack of variety.
\item Poor Food Quality: Dishes that are poorly cooked, tasteless, or of low quality.
\item Overpriced: Excessive prices for the quality and portion sizes of the food.
    \end{itemize}
\end{itemize}

\subsubsection{Descriptive Details}
We use cuisine types as the characteristics of restaurants, and the attributes are listed as follows:
\begin{itemize}
    \item Turkish
    \item Spanish
    \item Greek
    \item Italian
    \item French
    \item American
    \item Mexican
    \item Canadian
    \item Cajun
    \item Tex-Mex
    \item Brazilian
    \item Peruvian
    \item Argentinean
    \item Colombian
    \item Venezuelan
    \item Ethiopian
    \item Moroccan
    \item South African
    \item Nigerian
    \item Egyptian
    \item Chinese
    \item Japanese
    \item Indian
    \item Thai
    \item Korean
    \item Australian
    \item New Zealand
    \item Polynesian
    \item Hawaiian
    \item Singaporean
\end{itemize}


\section{Examples for Filtered Attribute Values}
\label{sec:filter_attr}
Here we give some examples of the filtered attributes.

For the Amazon product review dataset, some filtered attributes are listed as follows.
\begin{itemize}
    \item beauty:\begin{itemize}
    \item Hair Dryer (close to health and personal care)
    \item Hair Straightener (close to health and personal care)
\end{itemize}
\item electronics:\begin{itemize}
\item Car dashcam (close to automotive)
\item Wireless earbuds (close to cell\_phones\_service)
\end{itemize}
\item office\_products:\begin{itemize}
\item Mouse pad (close to electronics)
\end{itemize}
\end{itemize}

For NYT dataset, some filtered attributes are listed as follows:
\begin{itemize}
    \item american\_football:\begin{itemize}
    \item   The economic impact of football on local communities and businesses.
    \item The role of nutrition and hydration in optimal football performance 
    \item The history of football’s integration with America’s armed forces
\end{itemize}
\item stocks\_and\_bonds:\begin{itemize}
\item The role of artificial intelligence in stock market prediction
\item The effects of climate change on bond performance
\end{itemize}
\item movies:\begin{itemize}
\item The future of cinema in a post-COVID world
\item The role of international markets in Hollywood's global success
\end{itemize}
\end{itemize}

For Reddit dataset, some filtered attributes are listed as follows:
\begin{itemize}
    \item amc\_stock:\begin{itemize}
    \item  Working in the stock market industry.
    \item Impact of COVID-19 on amc\_stock.
    \item Having a background in finance or business.
\end{itemize}
\item what\_car\_should\_i\_buy:\begin{itemize}
\item Towing capacity and trailer options
\item Financing and leasing options
\item Road trip and travel experiences
\end{itemize}
\item pittsburgh:\begin{itemize}
\item Attending a game at PNC Park
\item Watch a Steelers football game at Heinz Field
\end{itemize}
\end{itemize}

For StackExchange dataset, some filtered attributes are listed as follows:
\begin{itemize}
    \item algorithm:\begin{itemize}
    \item Debugging an optimization algorithm for a complex logistics problem.
    \item Improving the robustness of an encryption algorithm against attacks.
    \item inding a way to better handle overflow errors in a cryptographic algorithm.
\end{itemize}
\item iPhone:\begin{itemize}
\item Unable to use certain features in certain regions/countries
\item Trouble with GPS not working properly
\end{itemize}
\item opengl:\begin{itemize}
\item Difficulty setting up a custom camera in OpenGL
\item Inconsistent rendering of 3D models in OpenGL
\item  Difficulty integrating OpenGL with a cross-platform game engine
\end{itemize}
\end{itemize}

\section{Examples for Merged Attribute on the  arXiv dataset}
\label{sec:merge_attr}
Here we give some examples of the filtered attributes.

\begin{itemize}
    \item optimization techniques for computer vision:\begin{itemize}
    \item computer vision and pattern recognition
    \item machine learning
\end{itemize}
\item black holes:\begin{itemize}
\item classical physics
\item popular physics
\end{itemize}
\item evolving networks for financial trading:\begin{itemize}
\item neural and evolutionary computing
\item computational finance
\end{itemize}
\item  multi-lingual information retrieval: \begin{itemize}
\item information retrieval
\item computation and language
\end{itemize}
\item iot operator systems and their security challenges: \begin{itemize}
\item operating systems
\item cryptography and security
\end{itemize}
\item complexity of social networks: \begin{itemize}
\item computational complexity
\item social and information networks
\end{itemize}
\end{itemize}

\section{The Prompt Format for MetaPrompt}
\label{sec:metaprompt}
{In this section, we provide more details about the implementation of MetaPrompt~\cite{reynolds2021prompt}. It first leverages LLMs to provide additional task-specific information, and then augments the class-dependent prompt with the task-specific knowledge.}

{\textbf{Prompts to ChatGPT}: \texttt{[Original Prompt]}. What does this task ask us to do?}

{\textbf{Response from ChatGPT for NYT dataset}: The task of generating an example of a NYT news asks the model to create high-quality, informative, and engaging news articles that mimic the writing style and structure commonly found in The New York Times (NYT).}

{\textbf{Response from ChatGPT for Amazon dataset}: The task of generating an example of a product review asks the model to create a detailed and informative review for a particular product, highlighting its features, performance, pros and cons, and overall user experience. This review should be helpful for potential buyers, providing them with valuable insights to make an informed purchasing decision.}

{\textbf{Response from ChatGPT for Reddit dataset}: The task of generating an example of a post similar to Reddit asks the model to craft a well-written and engaging post that adheres to the guidelines and community standards of the specific subreddit it's intended for. The post should spark discussions, share valuable information, or showcase interesting content that resonates with the subreddit's audience.} 

{\textbf{Response from ChatGPT for StackExchange dataset}: The task of  generating an example of a post similar to StackExchange asks the model to compose a clear and concise question or answer related to a specific topic within the given StackExchange community. The post should be well-researched and supported by relevant sources or personal expertise, aiming to provide valuable insights or solutions to other users' queries.}

\section{Dataset Information}
\subsection{Dataset Documentations}

The dataset is provided in \emph{json} format; there are three json files corresponding to the original  \texttt{train},  \texttt{validation} and  \texttt{test} split.
We also include two additional file, \texttt{attrprompt} and  \texttt{simprompt}, which is generated by \method and \baseline as the synthetic training data.

Each data point contains the following fields:
\begin{itemize}
    \item \texttt{label}: the label for the example. For multi-class classification, the \texttt{label} field is an integer, while for multi-label classification, the \texttt{label} field is a \texttt{list[int]} containing one or multiple integers as each example may refer to multiple classes;
    \item \texttt{text}: a content of each example.
\end{itemize}

\subsection{Intended Uses}
 \method and \baseline are intended for researchers in machine learning, natural language processing, and related fields to innovate novel methods for training data generation problems.

\subsection{Hosting and Maintenance Plan}
The codebase is hosted and version-tracked via GitHub.
It will be  available under the link \url{https://github.com/yueyu1030/attrprompt}.
The download link of all the datasets can be found in the Github repository.

Note that it is a community-driven and open-source initiative. 
We are committed and have the resources to maintain and actively develop it for at minimum the next five years. 
We plan to grow the GitHub repo by including new  tasks and datasets and warmly welcome external contributors.

\subsection{Licensing}
We license our work using Apache 2.0\footnote{https://www.apache.org/licenses/LICENSE-2.0}.
All of the \texttt{train/validation/test} data are publicly released by previous work \cite{blitzer2007biographies,meng2019weakly,geigle2021tweac}.

\subsection{Author Statement}
 We the authors will bear all responsibility in case of violation of rights.

\end{document}